\documentclass[11pt]{article}

\usepackage[utf8]{inputenc} 
\usepackage[T1]{fontenc}    
\usepackage{hyperref}       
\usepackage{url}            
\usepackage{booktabs}       
\usepackage{amsfonts}       
\usepackage{nicefrac}       
\usepackage{microtype}      
\usepackage{lmodern}

\usepackage{footmisc}

\usepackage{times}
\usepackage{graphicx}
\usepackage{subcaption}
\usepackage{amsfonts}
\usepackage{amsmath,amssymb}
\usepackage{bbm}
\usepackage{xcolor}
\usepackage{fullpage}
\usepackage{tikz}
\usepackage{wrapfig}
\usetikzlibrary{decorations.markings}

\usepackage[normalem]{ulem}

\usepackage{enumitem}

\newcommand{\E}{\mathbf{E}}

\usepackage{xspace}

\usepackage{algorithm,algpseudocode}

\DeclareMathOperator{\argmin}{argmin}

\newtheorem{theorem}{Theorem}

\newtheorem{lemma}{Lemma}

\newtheorem{remark}{Remark}
\newtheorem{proposition}{Proposition}

\newtheorem{proof}{Proof}
\newtheorem{assumption}{Assumption}

\DeclareMathOperator{\trace}{Tr}

\makeatletter
\let\@fnsymbol\@arabic
\makeatother

\begin{document}

\title{Learning with Gradient Descent and Weakly Convex Losses}

\author{Dominic Richards\footnote{Department of Statistics,   University of Oxford, 24-29 St Giles', Oxford, OX1 3LB \texttt{Dominic.richards@spc.ox.ac.uk}}\ 
\footnote{Work done while an intern at Facebook AI Research Montreal.}
  \qquad
  Mike Rabbat\footnote{Facebook AI Research Montreal, QC, Canada \texttt{Mikerabbat@fb.com}}
}
\date{\today}

\maketitle

\begin{abstract}
We study the learning performance of gradient descent when the empirical risk is weakly convex, namely, the smallest negative eigenvalue of the empirical risk's Hessian is bounded in magnitude. By showing that this eigenvalue can control the stability of gradient descent, generalisation error bounds are proven that hold under a wider range of step sizes compared to previous work. Out of sample guarantees are then achieved by decomposing the test error into generalisation, optimisation and approximation errors, each of which can be bounded and traded off with respect to algorithmic parameters, sample size and magnitude of this eigenvalue. In the case of a two layer neural network, we demonstrate that the empirical risk can satisfy a notion of local weak convexity, specifically, the Hessian's smallest eigenvalue during training can be controlled by the normalisation of the layers, i.e., network scaling. This allows test error guarantees to then be achieved when the population risk minimiser satisfies a complexity assumption. By trading off the network complexity and scaling, insights are gained into the implicit bias of neural network scaling, which are further supported by experimental findings.  
\end{abstract}

\section{Introduction}
A standard task in machine learning is to fit a model on a collection of training data in order to predict some future observations. With a loss function evaluating the performance of a model on a single data point, a popular technique is to choose a model that minimises the empirical risk, i.e., the loss with respect to all of the training data, with a form of regularisation to control model complexity and avoid over fitting. Due to the complexity of modern models (e.g., neural networks), many empirical risk problems encountered in practice are non-convex, resulting in an optimisation problem where finding a global minimizer is generally computationally infeasible. Naturally, this has motivated the adoption of tractable methods that incrementally improve the model utilising first order gradients of the empirical risk. In the case of gradient descent, model parameters are iteratively updated by taking a step in the direction of the negative gradient, and there is a source of implicit regularization controlled by algorithmic parameters such as the step size and number of iterations. 

A model's predictive performance is often measured by the population risk (i.e., expected loss on a new data point), which can be decomposed into optimisation and generalisation errors~\cite{bousquet2008tradeoffs}. The optimisation error accounts for how well the model minimises the empirical risk, and thus, decreases with iterations of gradient descent. Meanwhile, the generalisation error accounts for the discrepancy between the empirical and population risk, and thus, intuitively increases with the iterations of gradient descent as the model fits to noise within the training data. Guarantees for the population risk are then achieved by considering these two errors simultaneously and, in our case, trading them off each other by choosing the number of iterations and step size appropriately, i.e., early stopping.

The optimisation and generalisation errors for gradient descent have been investigated under a variety of different structural assumptions on the loss function. For the optimisation error a rich literature on convex optimisation can be leveraged, see for instance \cite{nesterov2013introductory}, while a broader range of non-convex settings can be considered that include weak convexity \cite{nurminskii1973quasigradient} as well as the Polyak-Łojasiewicz and quadratic growth conditions \cite{karimi2016linear}. Meanwhile for the generalisation error, the technique of stability \cite{devroye1979distribution,Bousquet:2002} has been applied to variants of gradient descent in both the convex and non-convex settings \cite{Hardt:2016:TFG:3045390.3045520,lin2016generalization,kuzborskij2017data,chen2018stability,madden2020high} with differing degrees of success. Specifically, although near-optimal rates are achieved in the convex case, for general non-convex losses restrictive step size conditions are currently required as bounds grow exponentially with the step size, number of iterations and the loss's smoothness \cite{Hardt:2016:TFG:3045390.3045520,kuzborskij2017data,yuan2019stagewise,madden2020high}. This is in contrast to applications with non-convex losses where gradient descent is routinely applied with a variety of step sizes. One possible explanation for this difference is that problems encountered in typical applications are not arbitrarily non-convex; rather, the curvature of losses involving neural networks, as measured empirically by the Hessian spectrum, can often be more benign \cite{lecun2012efficient,sagun2016eigenvalues,sagun2017empirical,yao2018hessian,ghorbani2019investigation,yuan2019stagewise}. This leads to the question of whether there are natural curvature assumptions that yield improved generalisation error bounds for gradient descent in the non-convex setting.

In this work we study the generalisation performance of gradient descent on loss functions that satisfy a notion of weak convexity \cite{nurminskii1973quasigradient}. The assumption relates to the loss's curvature as encoded by the Hessian spectrum, specifically, the magnitude of the most negative Eigenvalue. Our first main result (Theorem \ref{thm:GenErrorBound}) shows that the magnitude of this Eigenvalue can control the stability of gradient descent, and thus, the generalisation error. Precisely, provided the magnitude of this Eigenvalue is sufficiently small with respect to the step size and number of iterations, and the sample size is sufficiently large, then generalisation error bounds on the same order as the convex setting hold with a wider range of step sizes than previously known~\cite{Hardt:2016:TFG:3045390.3045520,kuzborskij2017data}.

Building upon our first result, guarantees on the population risk for gradient descent with weakly convex losses are then achieved by combining our generalisation error bounds with optimisation error bounds. In short, we note that the objective is convex when adding $\ell_2$ regularisation scaled by the magnitude of the smallest negative Eigenvalue. Utilising this as a proof device, the optimisation error becomes tractable at the cost of an additional statistical bias, resulting from the regularisation, which can be interpreted as an approximation error.  Our population risk bounds then hold provided the approximation error is below the statistical precision, which then requires an assumption on the complexity (as encoded by the $\ell_2$ norm) of a population risk minimiser.

Utilising our population risk bounds, insights are then gained into the influence of neural network scaling on learning. Specifically, we consider a loss that is the composition of a convex function and a two layer neural network. The final layer of the network then being scaled by a coefficient that is decreasing with the network width to a polynomial power, allowing us to interpolate between what are referred to as the kernel (power 1/2 scaling) and mean field (power 1 scaling) regimes \cite{chizat2019lazy}. By showing that the magnitude of the Hessian's smallest negative Eigenvalue can be controlled by the network scaling, we show gradient descent can generalise provided the neural network is sufficiently wide. In particular, this suggests that \emph{wider} networks with \emph{larger scalings} are \emph{more stable}, and thus, can be trained for longer. Moreover, by trading off the complexity of the population risk minimiser with the network scaling, we argue  (theoretically and empirically) that networks with larger scaling (higher power) are biased to have weights with larger norms. We now summarise our primary contributions.  
\begin{itemize}[leftmargin = *]
    \item \textbf{Generalisation Error Bound under Pointwise Weak Convexity.} Prove generalisation error bounds for gradient descent with point wise weak convexity (Assumption~\ref{ass:LocalWeakConvexity}). When the weak convexity parameter is small enough with respect to the step size and number of iterations, and sample size is sufficiently large, bounds hold under milder step sizes conditions than in previous work \cite{Hardt:2016:TFG:3045390.3045520,kuzborskij2017data} (Theorem \ref{thm:GenErrorBound}).
    \item \textbf{Test Error Bounds under Weak Convexity.} Prove test error bounds for gradient descent with weak convexity (Assumption \ref{ass:WeakConvexity:Global}). When the weak convexity parameter is small enough and the minimiser of the population risk has small norm, bounds on the same order of the convex setting are achieved. (Theorem \ref{thm:TestError:Simple})
    \item \textbf{Influence of Neural Network Scaling.} When the loss is a composition of a convex function and a neural network, we prove that the weak convexity parameter is on the order of the network scaling. Utilising this in conjunction with our theoretical results and empirical evidence, we argue that networks with smaller scaling are biased towards weights with larger norm (Section \ref{sec:TwoLayerNN}). 
\end{itemize}

The remainder of this work proceeds as follows. Section \ref{sec:RelatedWork} summarises related works. 
Section \ref{sec:GeneralResults} introduces the setting we consider as well as the generalisation error bounds. Section \ref{sec:GeneralTestError} presents bounds on the population risk for gradient descent with weakly convex losses. Section \ref{sec:TwoLayerNN} considers the particular case of two layer neural networks. 

\subsection{Related Work}
\label{sec:RelatedWork}
In this section we discuss a number of related work. For clarity, we adopt standard big $O(\cdot)$ notation so $a = O(b)$ if there is a constant $c > 0$ independent of dimension, sample size, iterations and stepsize such that $a \leq c b$.

\textbf{Learning with First Order Gradient Methods.} Guarantees on the population risk for gradient descent typically fall into one of two categories: single-pass or multi-pass. In the single-pass setting, each sample is used once to gain an unbiased estimate of the population risk gradient, and thus, guarantees following from studying the optimisation performance of stochastic gradient descent. Weak convexity has then be previously investigated within the optimisation community \cite{poliquin1992amenable,poliquin1996prox,rockafellar1981favorable,davis2018stochastic}, with the most relevant work to ours being \cite{davis2018stochastic} which showed, for non-smooth objectives, a first order stochastic proximal algorithm converges to a stationary point at the rate $O(t^{-1/4})$ within $t$ iterations.  In contrast, we focus on smooth losses, standard gradient descent and guarantees for the function value. 

In the multi-pass setting of this work, optimisation and generalisation errors are considered as each data point is used multiple times. To bound the generalisation error, stability \cite{devroye1979distribution,Bousquet:2002} was applied within \cite{Hardt:2016:TFG:3045390.3045520,lin2016generalization,kuzborskij2017data,yuan2019stagewise,madden2020high} for stochastic gradient descent. In the general non-convex setting these works then require an $O(1/t)$ step size after $t$ iterations. In contrast, with \emph{pointwise} weak convexity and a sufficiently large sample size $N$, our bounds hold with an $O(1/(\epsilon t^{\alpha}))$ step size  where $0 \leq \alpha \leq 1$ and $\epsilon$ is an upper bound on the magnitude of the Hessian's (evaluated at the points of gradient descent only) smallest negative Eigenvalue.

A number of other works have investigated the generalisation of first-order gradient methods, which we now briefly discuss. The work \cite{lin2016iterative} considers the multi-pass setting for gradient descent on convex losses. The works \cite{charles2017stability,yuan2019stagewise,madden2020high} considers the stability on non-convex loss functions which satisfy Polyak-Łojasiewicz and/or quadratic growth conditions. Their setting is different  to ours as their curvature conditions require that the gradient norm grows unbounded on unbounded domains  (hence, e.g., excluding the logistic loss). The work \cite{mou2017generalization} gives stability bounds for Stochastic Gradient Langevin Dynamics (SGLD), while \cite{chen2018stability} demonstrate a trade off between the stability and the  optimisation error. Finally, we note works studying multi-pass gradient methods for non-parametric regression \cite{bauer2007regularization,rosasco2015learning,pillaud2018statistical,pagliana2019implicit}.

\textbf{Scaling and Spectral Properties of Neural Networks.} 
The scaling of neural networks has been shown to influence the inductive bias in a number of settings \cite{chizat2019lazy,woodworth2020kernel}, with two popular choices of scaling being ``Neural Tangent Kernel'' (NTK) \cite{jacot2018neural,du2018gradientb,du2018gradienta,lee2019wide,allen2019convergence,zou2020gradient,arora2019fine,arora2019exact,zou2019improved,cao2019generalization,ji2019polylogarithmic,jacot2019asymptotic} 
and mean field \cite{chizat2018global,mei2018mean,mei2019mean,chizat2019lazy}. Specifically, it was found that a larger network scaling (i.e., mean field),  can yield a richer implicit bias \cite{chizat2019lazy,woodworth2020kernel}. This aligns with our findings (Section \ref{sec:TwoLayerNN:Approx}) where, in short, a larger network scaling allows gradient descent to learn parameters with larger norms. A number of works have also investigated the loss's Hessian when using neural networks, which we now discuss. By decomposing the loss's Hessian into two matrices, the work \cite{jacot2019asymptotic} studies the 
asymptotic moments of the Hessian with each of the aforementioned scalings. Similarly, \cite{pennington2017geometry} utilise random matrix theory to study the spectra of this decomposition by modelling each matrix individually. In each case, focus is placed upon the spectral distribution of the sum, which, as suggested by \cite{jacot2019asymptotic}, can be dominated by the first matrix. 
In contrast, our work is purely focused on the negative Eigenvalues, for which the second matrix primarily contributes since the first matrix is positive semi-definite in our case. \footnote{The work \cite{jacot2019asymptotic} also demonstrates each matrix within the decomposition becomes orthogonal within the limit, in which case, it suffices to study the spectra of each matrix individually.} We also note the recent works \cite{liu2020linearity,liu2020loss} which demonstrate (in a similar manner to us) that the second matrix's operator norm is a different order of magnitude versus the first. 
More generally, studies investigating the loss's Hessian spectrum can be traced back to \cite{bourrely1989parallelization,bottou1991stochastic} with a number of follow-up works \cite{lecun2012efficient,sagun2016eigenvalues,sagun2017empirical,yao2018hessian,ghorbani2019investigation,yuan2019stagewise}. Within \cite{sagun2017empirical,yuan2019stagewise} in particular, it was demonstrated empirically that the negative Eigenvalues of the Hessian decreased in magnitude during training. This observation is then explicitly leveraged within our work through a pointwise weak convexity assumption (Assumption~\ref{ass:LocalWeakConvexity} below).

\textbf{Notation}: For vectors $\omega, x \in \mathbb{R}^{p}$, denote the $i$th co-ordinate as $\omega_i$, as well as standard Euclidean inner product $\langle \omega ,x \rangle = \omega^{\top} x = \sum_{i=1}^{p} \omega_i x_i$ and $\ell_2$ norm $\|\omega \|_2 = \langle \omega,\omega \rangle^{1/2}$. For matrices $A \in \mathbb{R}^{p \times q}$ denote the $i,j$th entry by $A_{ij}$ and the Euclidean operator norm, equivalently spectral norm, as $\|A\|_2$. The Frobenius norm (i.e., entrywise $\ell_2$ norm) of a matrix $A$ is denoted as $\|A\|_F$. For square matrices $A \in \mathbb{R}^{p \times p}$ let $A \succeq 0$ denote that $A$ is positive semi-definite; i.e., for any $u \in \mathbb{R}^{p}$ we have $u^{\top} A u \geq 0$. For $B \in \mathbb{R}^{p \times p}$ denote $A \succeq B$ if $A -B$ is positive semi-definite. For a function $f:\mathbb{R}^{p} \rightarrow \mathbb{R}$ denote the gradient with respect to the $i$th co-ordinate as $\partial f(\omega)/\partial \omega_i $, and the vector of gradients $\nabla f(\omega) \in \mathbb{R}^{p}$ so that $(\nabla f(\omega) )_i= \partial f(\omega)/\partial \omega_i$.  Denote the second derivative of a function with respect to the $i,j$th co-ordinates as $\partial f(\omega)/\partial \omega_i \partial \omega_j$ as well as the Hessian of the function $\nabla^2 f(\omega) \in \mathbb{R}^{p \times p}$ such that $(\nabla^2 f(\omega))_{ij} = \partial f(\omega)/\partial \omega_i \partial \omega_j$.

\section{Setup and Generalisation Error Bounds}
\label{sec:GeneralResults}
In this section we formally introduce the learning setting as well as the generalisation error bounds. Section \ref{sec:Setup} formally introduces the setting.  Section \ref{sec:GenErrorBound} presents the assumptions and generalisation error bound. Section \ref{sec:ProofSketch} presents a sketch proof of the generalisation error bound. 

\subsection{Generalisation, Stability and Gradient Descent}
\label{sec:Setup}
We consider a standard learning setting \cite{vapnik2013nature} which we now introduce.  
Let models be parameterised by Euclidean vectors $\omega \in \mathbb{R}^{p}$ and data points be denoted by $Z \in \mathcal{Z}$. A loss function $\ell: \mathbb{R}^{p} \times \mathcal{Z} \rightarrow \mathbb{R}$ then maps a model $\omega$ and data point $Z$ to a real number $\ell(\omega,Z) \in \mathbb{R}$. Observations are random variables following an unknown population distribution $Z \sim \mathbb{P}$, and the objective is to produce a model $\omega$ that minimises the expected loss with respect to the observations i.e. \emph{population risk} $r(w) := \E_{Z}[\ell(w,Z)]$.  To produce a model,  a collection of independently and identically distributed samples $Z_{i} \sim \mathbb{P}$ for $i=1,\dots,N$ are observed and an approximation to the population risk is considered; i.e., the \emph{empirical risk}  
$
    R(w) :=  \sum_{i=1}^{N} \ell(w,Z_{i})/N.
$
In our case, an algorithm maps the observations to a model $\mathcal{A}:\mathcal{Z}^{N} \rightarrow \mathbb{R}^{p}$ so that $\widehat{\omega} = \mathcal{A}(Z_1,\dots,Z_N)$. We then begin by considering the \emph{generalisation error} $\E[R(\widehat{w}) - r(\widehat{w})]$. 

To study the generalisation error we consider its stability \cite{Bousquet:2002}. Specifically, for $i=1,\dots,N$ consider the estimator with the $i$th data point resampled independently from the population, that is, $\widehat{\omega}^{(i)} = \mathcal{A}(Z_1,\dots,Z_{i-1},Z_i^{\prime},Z_{i+1},\dots,Z_N)$ where $Z_i^{\prime} \sim \mathbb{P}$. The expected generalisation error can then be written as \cite{Bousquet:2002}
\begin{align}
\label{equ:GenErrorStability}
\underbrace{ \E[R(\widehat{w}) \! - \! r(\widehat{w})]}_{\text{Generalisation Error}}
\! = \!
\frac{1}{N}\sum_{i=1}^{N} \E[ 
\underbrace{  \ell(\widehat{w}^{(i)},Z_{i}^{\prime}) \! - \! \ell(\widehat{w},Z_{i}^{\prime})}_{\text{Stability}}  ].
\end{align}
The stability of an estimator  aligns with its sensitivity to individual data points in the training set. Intuitively, a stable estimator does not change much if a data point is resampled, and thus, generalises better owing to not depending in a strong way on any individual datapoint. This is captured within the above equality, where the difference on the right hand side involves the change in loss when resampling each training data point.

The algorithm considered will be gradient descent applied to the empirical risk. This produces a sequence of estimates indexed by $s \! \geq \! 1$ and denoted $\{\widehat{\omega}_s\}_{s \geq 0}$. For a sequence of non-negative step sizes $\{\eta_s\}_{s \geq 0}$ and an initialisation $\widehat{\omega}_0 \in \mathbb{R}^{p}$, the iterates of gradient descent are defined recursively for $s \! \geq \! 0$ as  
\begin{align*}
    \widehat{\omega}_{s+1} = \widehat{\omega}_{s} - \eta_s \nabla R(\widehat{\omega}_{s}).
\end{align*}
For $i=1,\dots,N$ let us denote the sequence of estimators produced with the $i$th observation resampled by $\{\widehat{\omega}^{(i)}_{s} \}_{s \geq 0}$. These estimators are initialised at the same point $\widehat{\omega}_{0}^{(i)} = \widehat{\omega}_0$ and are updated using the gradient of the empirical risk with the resampled data point $\nabla R^{(i)}(\omega) := \nabla R(\omega) +  \big( \nabla \ell(\omega,Z_i^{\prime}) -  \nabla \ell(\omega,Z_i) \big)/N$.

\subsection{Generalisation Error Bound for Gradient Descent with Weakly Convex Losses}
\label{sec:GenErrorBound}
In this section we present a bounded on the generalisation error when the empirical risk satisfies a notation of weak convexity. We begin with a collection of Lipschitz (Lip.) assumptions on the loss function and its gradients.
\begin{assumption}[Loss Regularity]
\label{ass:LossReg}
There exists $L,\beta,\rho \geq 0$ such that  for any $\omega,\omega^{\prime} \in \mathbb{R}^{p}$ and $Z \in \mathcal{Z}$:\\
$L$-Lip. loss:
$
|\ell(\omega,Z) - \ell(\omega^{\prime},Z)|
 \leq 
L \|w - w^{\prime}\|_2.
$ \\
$\beta$-Lip. gradient:
$\|\nabla \ell(\omega,Z) - \nabla \ell(\omega^{\prime},Z) \|_2  \leq \beta \|\omega - \omega^{\prime}\|_2.$\\
$\rho$-Lip. Hessian:
$\|\nabla^2 \ell(\omega,Z) \! - \! \nabla^2 \ell(\omega^{\prime},Z) \|_2 \leq \rho \|\omega - \omega^{\prime}\|_2.$
\end{assumption}
The first two conditions in Assumption \ref{ass:LossReg} respectively state that the loss is $L$-Lipschitz and $\beta$-smooth, which are common assumptions when considering both the optimisation and generalisation of gradient descent algorithms \cite{nesterov2013introductory,Hardt:2016:TFG:3045390.3045520}. The third condition has been considered previously when studying the stability of gradient descent \cite{kuzborskij2017data} and is satisfied if the third-order derivative exists and is bounded \cite{ge2015escaping}. 

The next assumption is related to the Hessian of the empirical risk, and encodes a notation of weak convexity.
\begin{assumption}[\textbf{Pointwise Weak Convexity}]
\label{ass:LocalWeakConvexity}
There exists a non-negative deterministic sequence $\{\epsilon_s \}_{s \geq 0} $ such that \emph{almost surely} for any $s \geq 0$ and  $i = 1,\dots,N$ 
\begin{align*}
    \nabla^2 R(\widehat{w}_s) \succeq - \epsilon_s I 
    \quad
    \text{ and }
    \quad 
    \nabla^2 R^{(i)}(\widehat{w}^{(i)}_s) \succeq - \epsilon_s I \;.
\end{align*}
\end{assumption}
Assumption \ref{ass:LocalWeakConvexity} states that the smallest negative Eigenvalue of the Hessian is \emph{almost surely} lower bounded at the points evaluated by gradient descent. This is milder than standard weak convexity, which requires a lower bound everywhere, i.e., for there to exist $\epsilon > 0$ such that $\nabla^2 R(\omega) \succeq -\epsilon I $ for any $\omega \in \mathbb{R}^{p}$. The lower bound also depends upon the time step $s \geq 1$ of gradient descent. This allows us to encode that the magnitude of the most negative Eigenvalue can decrease during training \cite{sagun2017empirical,yuan2019stagewise} (precisely Figure 2 in \cite{yuan2019stagewise}), which is itself linked to the residuals contributing to the Hessian's negative Eigenvalues \cite{pennington2017geometry,jacot2019asymptotic}. In Section \ref{sec:TwoLayerNN} we demonstrate for the composition of a smooth convex function and a two layer neural network, that Assumption \ref{ass:LocalWeakConvexity} can be satisfied provided the network is sufficiently wide. With these assumptions we now bound the generalisation error of gradient descent. 

\begin{theorem}[Generalisation Error Bound]
\label{thm:GenErrorBound}
Consider Assumptions \ref{ass:LossReg} and \ref{ass:LocalWeakConvexity},  $1 \geq \alpha \geq 0$ and $t \geq 1$. If $\max_{t \geq k \geq 0} \eta_k \beta \leq 3/2$, $\max_{t \geq k \geq 0} \eta_{k} ( \epsilon_{k} + \frac{2 \beta}{N}) + \eta^{\frac{1}{\alpha}}_k < 1/2$ and
\begin{align*}
     N & \! \geq \!
     24\rho L \!
    \exp\Big( 
    2  \! \sum_{s=1}^{t} \eta_s\big( \epsilon_{s} \! + \! \frac{4  \beta}{N} \big)
    \! + \! \eta^{\frac{1}{\alpha}}_{s}
    \Big)
    \sum_{j=1}^{t} \eta_{j}^{1-\frac{1}{2\alpha}} \sum_{\ell=0}^{j-1} \eta_\ell, 
\end{align*}
then the generalisation error of gradient descent satisfies
\begin{align*}
     \E[R(\widehat{w}_t) - r(\widehat{w}_t)]
     \leq
    \frac{4L^2}{N}\sum_{j=0}^{t-1} 
    \exp\Big( 
    2  \sum_{s=j+1}^{t-1} \eta_s \big( \epsilon_{s} + \frac{4 \beta}{N}\big)
    +  \eta^{\frac{1}{\alpha}}_{s}
    \Big)\eta_j.
\end{align*}
\end{theorem}
We now provide some discussion of Theorem \ref{thm:GenErrorBound}. The condition on the sample size ensures that a sufficiently small step size $\{\eta_s\}_{s \geq 0}$ and number of iterations $t$ is taken with respect to the total number of data points $N$. In particular, if we choose the step size $\eta_{s} = (s+1)^{-\alpha}/\log(t)$ for $s \geq 0$ and the minimum Eigenvalue is lower bounded so that $\epsilon_{s} \leq 1/t^{1-\alpha}$ for $s \geq 1$,  then the term within the exponential is $O(t^{1-\alpha}/N)$. We would then expect this quantity to be small as it aligns with the Generalistion Error bound in the smooth and convex setting \cite{Hardt:2016:TFG:3045390.3045520}.  The condition on the sample size $N$ is then dominated by the summation over step sizes, which are $\sum_{j=1}^{t} (j+1)^{-\alpha + 1/2}\sum_{\ell=0}^{j-1} (\ell+1)^{-\alpha} = O( t^{2(1-\alpha) + 1/2})$, and thus, we require the number of iterations to be upper bounded by $t = O(N^{1/(2(1-\alpha)+1/2)})$. Picking $\alpha = 3/4$, the iterations can grow linearly with the total number of samples, while $\alpha >3/4$ allows the iterations to grow as $O(N^{q})$ for some $1 \leq q \leq 2$.  Given the condition on the sample size is satisfied, the resulting generalisation error bound is on the  order of $O(t^{1-\alpha}/N)$, aligning with the smooth and convex setting \cite{Hardt:2016:TFG:3045390.3045520}. 

In comparison to previous generalisation error bounds for gradient descent with non-convex objectives, see for instance \cite{Hardt:2016:TFG:3045390.3045520,kuzborskij2017data}, we highlight that the bound in Theorem \ref{thm:GenErrorBound} allows for a larger step size to be taken. Previous bounds held in the non-convex setting with $\eta_{s} = O(1/s)$, while here under the additional assumption of point wise weak convexity, we can consider $\eta_s = O(1/s^{\alpha})$ for $\alpha \in [0,1]$. Such a step size will allow us to achieve guarantees on the optimisation error, and thus, the test error (see Section \ref{sec:GeneralTestError}).

\subsection{Proof Sketch of Theorem \ref{thm:GenErrorBound}}
\label{sec:ProofSketch}
Here a proof sketch of Theorem \ref{thm:GenErrorBound} is provided, with the full proof given in Appendix \ref{sec:ProofGen:Thm}. Recalling that the loss is $L$-Lipschitz and recalling \eqref{equ:GenErrorStability}, we see to bound the generalisation error it is sufficient to bound the deviation between the iterates of gradient descent with and without the resampled datapoint, i.e., $\widehat{\omega}_{t} - \widehat{\omega}_{t}^{(i)}$. Using the definition of the empirical risk gradient with the sampled data point $\nabla R^{(i)}(\omega) $ and the triangle inequality, we get for $k \geq 1$ 
\begin{align*}
     \|\widehat{\omega}_{k} - \widehat{\omega}_{k}^{(i)}\|_2 
     \leq \frac{2 \eta L}{N}  +
    \underbrace{ \|\widehat{\omega}_{k-1} - \widehat{\omega}_{k-1}^{(i)} - \eta \big( \nabla R(\widehat{\omega}_{k-1}) - \nabla R(\widehat{\omega}_{k-1}^{(i)}) \|_2 }_{\textbf{Expansiveness of Gradient Update} },
\end{align*}
where $2\eta L/N$ arises from using the Lipschitz property to upper bound $\eta \big( \nabla \ell(\omega,Z_i^{\prime}) -  \nabla \ell(\omega,Z_i) \big)/N$ for any $\omega \in \mathbb{R}^{p}$. The remaining term is referred to as the expansiveness of the gradient update \cite{Hardt:2016:TFG:3045390.3045520}. To provide context, we now describe some previous approaches for bounding this term.

Following previous work \cite{Hardt:2016:TFG:3045390.3045520}, when the loss is convex we have for any $x,y \in \mathbb{R}^{p}$ that $\|x-y - \eta (\nabla R(x)  - \nabla R(y))\|_2 \leq \|x-y\|_2 $. This allows the deviation to be simply unravelled and bounded $\|\widehat{\omega}_{k} - \widehat{\omega}_{k}^{(i)}\|_2  \leq 2\eta k /N$.  Meanwhile, in the non-convex setting
the expansiveness was upper bounded as $\|x-y - \eta (\nabla R(x)  - \nabla R(y))\|_2 \leq (1 + \eta \beta) \|x-y\|_2$, ultimately yielding a bound on the order of $\|\widehat{\omega}_{k} - \widehat{\omega}_{k}^{(i)}\|_2  \leq \exp(\eta \beta t) \eta t /N$. To control this exponential term we then require $\eta = O(1/t)$. 

The presence of the smoothness coefficient $\beta$ within the exponential in the non-convex setting arises to control the loss curvature through an upper bound on the Hessian's  spectral norm $\|\nabla^2 \ell(\cdot,Z)\|_2$ (see also the data-dependent bound in \cite{kuzborskij2017data} and Lemma 12 in appendix of \cite{yuan2019stagewise}). Although, when comparing to the convex case, we intuitively believe that the exponential term may only depend upon negative Eigenvalues of the Hessian, i.e., the least Eigenvalue, since that can differentiate the convex and non-convex cases.  For clarity, let us then assume that the loss is $\epsilon$-weakly convex so $\nabla^2 R(\omega) \succeq -\epsilon I$ for any $\omega \in \mathbb{R}^{p}$. Then the expansiveness of the gradient updated can be upper bounded $\|x-y - \eta (\nabla R(x)  - \nabla R(y))\|_2 \leq \|x-y\|_2/\sqrt{1 - 2 \eta \epsilon}$ (see proof of Theorem \ref{thm:GenErrorBound:GlobalWeakConvexity} in Appendix \ref{sec:ProofGen:Global}). The multiplicative factor can then be interpreted as $1/(1 - 2 \eta \epsilon) = 1 + 2 \eta \epsilon/(1-2\eta \epsilon) \leq \exp\big(2 \eta \epsilon/(1- 2\eta \epsilon)) $ and therefore directly controlled by the weak convexity parameter $\epsilon$. In contrast, previously the multiplicative factor was $1 + \eta \beta$. Unravelling the deviation with this upper bound yields a generalisation error bound on the order $\exp(\epsilon \eta t) \eta t /N$ (see Theorem \ref{thm:GenErrorBound:GlobalWeakConvexity} in Appendix), with the convex case being recovered when $\epsilon = 0$. Now, the condition on the sample size $N$ within Theorem~\ref{thm:GenErrorBound} arises from the fact we consider a milder  \emph{pointwise} weak  convexity assumption which only evaluates the Hessian at the iterates of gradient descent (note standard weak convexity is a global lower bound). Precisely, the milder \emph{pointwise} assumption results in the expansiveness of the gradient update containing higher order terms.

\begin{remark}[Extension to Stochastic Gradient Descent]
Extending to the stochastic gradient setting, where a subset of randomly chosen data points are evaluated at each iteration, is challenging here as the higher order terms in the gradient expansiveness bound  results in higher order moments of the deviation at previous iterations. We thus leave this direction to future work as we may require stronger high probability bounds, see also \cite{madden2020high}. 
\end{remark}

\section{Test Error Bounds for Gradient Descent with Weakly Convex Losses}
\label{sec:GeneralTestError}
In this section we utilise our generalisation error bounds to achieve guarantees on the population risk for gradient descent with weakly convex losses.
The remainder of this section is then as follows. Section \ref{sec:TestErrorDecomp} presents the error decomposition. Section \ref{sec:TestErrorWeaklyConvex} presents bounds on the population risk for the iterates of gradient descent.

\subsection{Test Error Decomposition}
\label{sec:TestErrorDecomp}
Recall we wish to produce a model that minimised the \emph{population risk} $r(\omega)$. Therefore, denoting a population risk minimiser $\omega^{\star} \in \argmin_{\omega} r(\omega)$, we now set to investigate the  \emph{Test Error} of gradient descent. Specifically, for an independent uniform random variable $I \sim \text{Uniform}(1,\dots,t)$ we consider 
$
    \E_{I}[\E[ r(\widehat{\omega}_{I})]] - r(\omega^{\star}).
$
We note that the iterate is evaluated at a uniform random variable $\widehat{\omega}_I$ as the loss is non-convex, and thus, the loss at the average of iterates $\frac{1}{t} \sum_{s=1}^{t}\widehat{\omega}_s$ is not immediately upper bounded by the average of the losses $E_{I}[r(\widehat{\omega}_I)]$.

To bound the test error we consider the following decomposition
\begin{align}
\label{equ:TestError:Decomp}
    \underbrace{\E_{I}[\E[ r(\widehat{\omega}_{I})]] - r(\omega^{\star})}_{\textbf{Test Error}}
     = 
    \underbrace{ 
    \E_{I}[\E[  r(\widehat{\omega}_{I}) - R(\widehat{\omega}_{I}) ]]
    }_{\textbf{Gen. Error}} 
    + 
    \underbrace{ \E_{I}[ E[ R(\widehat{\omega}_{I}]]  - r(\omega^{\star})}_{\textbf{Opt. \& Approx. Error}}.
\end{align}
The first term intuitively accounts for the discrepancy between the empirical and population risk and aligns with the generalisation error studied within Section \ref{sec:GeneralResults}. The second term, the \textbf{Optimisation and Approximation Error}, will be composed of two components. The first component is an optimisation error that, intuitively, accounts for how well the iterates of gradient descent minimise the empirical risk. The second component is an approximation error arising from the non-convex setting. Specifically, it results from a proof technique of adding regularisation to make the loss convex, and thus, the optimisation error tractable. The approximation error therefore reflects the weak convexity assumption. Specifically, it will be scaled by the weak convexity parameter and more generally (see Proposition~\ref{prop:TestError:General} in Appendix~\ref{app:GenWeakConv}) can depend upon the Hessian structure. Section~\ref{sec:TestErrorWeaklyConvex} then presents the upper bound on the test error utilising this error decomposition.

\subsection{Test Error for Weakly Convex Losses}
\label{sec:TestErrorWeaklyConvex}
In this section we present a test error bound for weakly convex losses. Let us begin with the following assumption.
\begin{assumption}[Weak Convexity]
\label{ass:WeakConvexity:Global}
There exists an $\epsilon > 0$ such that $\nabla^2 R(\omega) \succeq -\epsilon I$ for all $\omega \in \mathbb{R}^{p}$. 
\end{assumption}
Given this assumption, let us define the minimiser of the penalised objective $\widehat{\omega}_{\epsilon}^{\star} = \argmin_{\omega}\big\{ R(\omega) + \epsilon\|\widehat{\omega}_0 - \omega\|_2^2\big\}$, as well as a minimiser of the unpenalised objective $\widehat{\omega}^{\star} \in \argmin_{\omega} R(\omega)$. The test error bound is then presented within the following theorem.
\begin{theorem}
\label{thm:TestError:Simple}
Let Assumptions \ref{ass:LossReg} and \ref{ass:WeakConvexity:Global} hold, and consider constant step size $\eta_{s} = \eta$ for all $s \geq 1$. If $\eta \beta \leq 1/2$, $2 \eta \epsilon < 1$ then the test error of gradient descent is bounded
\begin{align*}
    \E_{I}[\E[ r(\widehat{\omega}_{I})]] \! - \! r(\omega^{\star})
     \! \leq \!
    \underbrace{ 
    \frac{2L^2 \eta t }{N} 
    \exp\Big( 
    \frac{\eta t \epsilon}{1 - 2 \eta \epsilon}
    \Big)
    }_{\textbf{Generalisation Error}} 
    \! + \! 
    \underbrace{ \frac{\E[ \|\widehat{\omega}_0  -  \widehat{\omega}_{\epsilon}\|_2^2] }{ 2 \eta t } }_{\textbf{Optimisation Error}}
     \! + \! 
    \underbrace{  \epsilon \Big(  \eta t \E[R(\widehat{\omega}_0)  - R(\widehat{\omega}^{\star})]  
      +  \|\widehat{\omega}_0 - \omega^{\star}\|_2^2 \Big)}_{\textbf{Approximation Error}}
\end{align*}
\end{theorem}
We firstly note that we have now assumed standard weak convexity (Assumption \ref{ass:WeakConvexity:Global}), and therefore, there is no condition on the sample size $N$ to control the \textbf{Generalisation Error} (recall proof sketch in Section \ref{sec:ProofSketch}). Meanwhile, the \textbf{Optimisation Error} and \textbf{Approximation Error} upper bounds the \textbf{Optimisation and Approximation Error} term given in \eqref{equ:TestError:Decomp}.
Naturally, the \textbf{Optimisation Error} decreases with both the step size and number of iterations $\eta t$. Meanwhile, the \textbf{Approximation Error} arises, in short, from the proof technique of using the empirical risk penalised by $\epsilon \|\cdot\|_2^2$ within the analysis. To achieve guarantees on the test error, the product of the smallest Eigenvalue and the norm of the population risk minimiser $\epsilon \|\omega^{\star}\|_2^2$ must be sufficiently small, which can then be interpreted as a complexity assumption on the learning problem. This interplay is precisely investigated for a two layer neural network within Section \ref{sec:TwoLayerNN:Approx}. Given that the \textbf{Approximation Error} is sufficiently small, we then see that an $O(1/\sqrt{N})$ test error bound can be achieved by choosing $\eta t= \sqrt{N}$. The test error bound in the convex case  \cite{Hardt:2016:TFG:3045390.3045520} can then be recovered by setting $\epsilon = 0$.

\begin{remark}[Iteration Condition from Theorem \ref{thm:GenErrorBound}]
It is natural to consider the test error bound that would arise if we restrict ourselves to the iteration condition in Theorem \ref{thm:GenErrorBound}. In short, if $\eta = (t+1)^{-\alpha}/\log(t)$ we then get an $O( N^{- \frac{1-\alpha}{1/2+2(1-\alpha)}})$ test error bound, with $\alpha = 3/4$ then yielding an $O(N^{-1/4})$ bound in  $t=O(N)$  iterations. Interestingly, this aligns with \cite{davis2018stochastic} who showed a stochastic proximal algorithm converges to a stationary point at the rate of $O(t^{-1/4})$ in the weakly convex case. Although, some care should be taken in making a direct comparison between our work and \cite{davis2018stochastic}, since they consider: a more general non-smooth objective, a different algorithm and convergence to a stationary point. We therefore leave an investigation into a connection between the two methods to future work. 
\end{remark}

\section{Two Layer Neural Networks}
\label{sec:TwoLayerNN}
In this section investigate the case of a two layer neural network. Section \ref{sec:TwoLayerNN:Setup} formally introduces the setting we consider. Section \ref{sec:TwoLayerNN:WeakConvexity} investigates the weak convexity parameter for two layer neural networks. Section \ref{sec:TwoLayerNN:Approx} considers the approximation error for two layer neural networks. Section \ref{sec:Experiments} presents experimental results. 

\subsection{Setup}
\label{sec:TwoLayerNN:Setup}
Consider the standard supervised learning setting where data points decompose as $Z = (x,y)$ with covariates $x \in \mathbb{R}^{d}$ and a response $y \in \mathbb{R}$. Let $g: \mathbb{R} \times \mathbb{R}  \rightarrow \mathbb{R}$ denote a smooth convex function that quantifies the discrepancy $g(\widehat{y},y)$ between predicted $\widehat{y}$ and observed $y$ responses. Consider prediction functions parameterised by $\omega \in \mathbb{R}^{p}$ and denoted $f(\cdot,\omega): \mathbb{R}^{d} \rightarrow \mathbb{R}$. The loss function at the point $Z = (x,y)$ is then the composition between $g(\cdot,y)$ and $f(x,\omega)$ so that $\ell(\omega,Z) : =  g(f(x,\omega),y)$.
We will be considering the gradient with respect to $\omega$, therefore denote the first and second order gradient of the function $g$ with respect to the first argument as $g^{\prime}(\cdot,y):\mathbb{R} \rightarrow \mathbb{R}$ and $g^{\prime\prime}(\cdot,y):\mathbb{R} \rightarrow \mathbb{R}$, respectively. Throughout we will assume that $g$ has uniformly bounded derivative $\max_{\widetilde{y},y}|g^{\prime}(\widetilde{y},y)| \leq L_{g^{\prime}}$, which is satisfied, for instance, if $g$ is the logistic loss.  

The class of prediction functions considered in this section will be two layer neural networks of width $M > 0$ and scaling $1 \geq c \geq 1/2$. The network consists of a first layer of weights represented as a matrix $A \in\mathbb{R}^{M \times d}$ with $j$th row $A_{j} \in \mathbb{R}^{d}$, a second layer of weights denoted as a vector $v = (v_1,\dots,v_{M}) \in \mathbb{R}^{M}$ and a smooth activation $\sigma: \mathbb{R} \rightarrow \mathbb{R}$. For an input $x \in \mathbb{R}^{d}$ the neural network then takes the form
\begin{align}
\label{equ:TwoLayerNN}
    f(x,\omega) :=  \frac{1}{M^c} \sum_{j=1}^{M} v_j \sigma( \langle A_j,x \rangle ).
\end{align}
Typically, both the first and second layer of weights of the network are optimised, in which case, let parameter vector $\omega$ be the concatenation $\omega = (A,v) \in \mathbb{R}^{Md + M}$, where $A$ is vectorised in a row-major manner. As we will see later, studying the case of optimising both layers simultaneously is challenging owing to the interactions between the first and second layers. It will therefore be insightful to also consider fixing the second layer of weights and only optimising the first layer. In this case denote $\omega = (A) \in \mathbb{R}^{Md}$.

\subsection{Weak Convexity of Two Layer Neural Networks}
\label{sec:TwoLayerNN:WeakConvexity}
We begin with the following theorem which considers weak convexity of the empirical risk. With data points $Z_i = (x_i,y_i)$ for $i=1,\dots,N$ let us denote the covariates empirical covariance as $\widehat{\Sigma} := \sum_{i=1}^{N} x_i x^{\top}_i/ N \in \mathbb{R}^{d \times d}$. \begin{theorem}
\label{thm:WeakConvexity:SingleLayerNN:Both}
Consider the loss function in Section \ref{sec:TwoLayerNN:Setup} with Two Layer Neural Network \eqref{equ:TwoLayerNN}. Suppose that the activation $\sigma$ has first and second derivatives bounded by $L_{\sigma^{\prime}}$ and $L_{\sigma^{\prime\prime}}$ respectively. Then with $\omega = (A,v)$,
\begin{align*}
    \nabla^2 \! R( \omega  )
    \! \succeq \!
    - \Big( \! L_{g^{\prime}} L_{\sigma^{\prime\prime}}  \|v\|_\infty \|\widehat{\Sigma}\|_2 \! + \! 2 L_{\sigma^{\prime}} L_{g^{\prime}} \! \sqrt{\|\widehat{\Sigma}\|_2 }\Big) \frac{1 }{M^c} I
\end{align*}
\end{theorem}
Theorem \ref{thm:WeakConvexity:SingleLayerNN:Both} demonstrates that the weak convexity parameter is on the order of the neural network scaling $O(1/M^c)$, suggesting it is smaller for wider networks with a larger scaling $c$. Note the bound consists of two terms: the first arising from the Hessian restricted of the first layer; and the second from the first and second layer of weights interacting.  We now discuss using this bound in conjunction with Theorem \ref{thm:GenErrorBound} and \ref{thm:TestError:Simple}.

Recall the generalisation error bound in Theorem \ref{thm:GenErrorBound} requires pointwise weak convexity Assumption \ref{ass:LocalWeakConvexity}. This considers the Hessian evaluated at the iterates of the gradient descent, and thus, if the initialisation and activation are bounded (to ensure the infinity norm of the second layer is $O(\sum_{s=0}^{t} \eta_s /M^{c})$) as well as the covariates co-ordinates so $\max_{i=1,\dots,N}\|x_i\|_{\infty} \leq L_{x}$,  Theorem \ref{thm:WeakConvexity:SingleLayerNN:Both} then ensures pointwise weak convexity Assumption \ref{ass:LocalWeakConvexity} holds with $\epsilon_s = O( d /M^c)$. As a consequence, it is sufficient to scale the network width $M \geq b (d\sum_{s=0}^{t} \eta_s)^{1/c} $, for some constant $b \geq 0$, to ensure the exponential terms within the generalisation error bound of Theorem \ref{thm:GenErrorBound} remain bounded. 

To bound the test error utilising Theorem \ref{thm:TestError:Simple} recall we require standard weak convexity Assumption \ref{ass:WeakConvexity:Global} to hold. Since the first term in Theorem  \ref{sec:TwoLayerNN:WeakConvexity} depends upon the infinity norm of the second layer $\|v\|_{\infty}$, it is not feasible here to satisfy standard weak convexity as we can take $\|v\|_{\infty} \rightarrow \infty$. This (as well as Remark \ref{remark:LimitationWeakConvexity}) motivates a more refined notation of weak convexity which is discussed in Appendix \ref{app:GenWeakConv}. For this reason, let us now consider fixing the second layer and optimising the first layer only. Applying Theorem \ref{thm:TestError:Simple} then requires controlling the \textbf{Approximation Error} which itself requires an assumption on the squared Euclidean norm of the population risk minimiser $\|\omega^{\star}\|_2^2$. Since the scaling and complexity of the network can interplay, this is discussed within Section \ref{sec:TwoLayerNN:Approx}.

\begin{remark}[Three Layer Neural Networks]
\label{remark:ThreeLayerNN}
In Appendix \ref{sec:WeakConvexityNN:Thm3} we show that the minimum Eigenvalue can be bounded for a three layer neural network when fixing the middle layer. If the width of each layer is $M$ the magnitude of the minimum negative Eigenvalue is $O(1/M^{2c-1/2})$, and thus, can be smaller than a two layer network when $c > 1/2$.  
\end{remark}

\subsection{Approximation Error for Two Layer Neural Networks}
\label{sec:TwoLayerNN:Approx}
In this section we study the \textbf{Approximation Error} term within the test error bound of Theorem \ref{thm:TestError:Simple} in the case of a two layer neural network. Following the discussion at the end of Section \ref{sec:TwoLayerNN:WeakConvexity} (as well as remark \ref{remark:LimitationWeakConvexity}), we  fix the second layer and only optimise the first so $\omega = (A)\in\mathbb{R}^{Md}$, with the case of optimising both layers studied in Appendix \ref{app:GenWeakConv}.

Controlling the \textbf{Approximation Error} requires placing assumptions on the Euclidean norm of the neural network weights, and thus, the complexity of the network. We therefore compare to another notion of complexity to ensure the network is not made too simple. While a number of different notions of complexity have been investigated in previous work, see for instance \cite{neyshabur2015norm,bartlett2017spectrally,arora2019fine}, we consider the network's Total Weight \cite{bartlett1998sample} defined as $\text{TW}(f) :=  \sum_{j=1}^{M} |v_j| \|A_j\|_2/M^c$. The following assumption then introduces the complexity assumption used to control the approximation error in our case. 
\begin{assumption}
\label{ass:StatisticalAssumption}
There exists $1 \geq \mu \geq 0$ and population risk minimiser $A^{\star}$ such that  $\|A^{\star}\|_F \leq M^{1/2-\mu}$.
\end{assumption}
Assumption \ref{ass:StatisticalAssumption} states there exists a set of first layer weights $A^{\star}$ with Frobenius norm bounded by $M^{1/2-\mu}$ that achieves the minimum population risk. Therefore, larger $\mu$ aligns with a stronger assumption, as the minimum risk can be attained by a set of weights with smaller norm. Although, some choices of $\mu$ are more natural than others when considering the choice of scaling $1 \geq c \geq 1/2$. Specifically, consider the following upper bound on the Total Weight $\text{TW}(f)= \frac{\|v\|_2}{M^c} \sum_{i=1}^{M} \frac{|v_i|}{\|v\|_2} \|A_i\|_2  \leq \|v\|_2 \|A^{\star}\|_F/ M^c $. Now, if each of the second layer weights are a constant so $\|v\|_2 = O(\sqrt{M})$ the Total Weight of the population risk minimiser under Assumption \ref{ass:StatisticalAssumption} is $O(M^{1-\mu -c})$. As such if $\mu > 1-c$, the Total Weight of the network goes to zero as the network becomes wider (i.e., larger $M$), yielding effectively constant prediction functions for wide networks. Therefore, a larger scaling $c$ can be seen to encourage networks with large norm weights. 

Let us now consider the approximation error in Theorem \ref{thm:TestError:Simple} when gradient descent is initialised at $\widehat{\omega}_0 = 0$ and Assumption \ref{ass:StatisticalAssumption} holds. The approximation error is then $O(d ( \eta t + M^{1-2\mu})/M^c)$, and therefore, to ensure it is decreasing in the width of the network $M$, we require $\mu > (1-c)/2$. How the network scaling $c$ and the statistical assumption $\mu$ then interplay can be clearly seen in Figure  \ref{fig:Stat:vs:Scale}. Intuitively, increasing the scaling $c$ allows networks with larger norms (larger $\mu$) to be approximated, but at the cost of not being able to consider networks with smaller norms. Whether encouraging larger norm networks through the scaling $c$ leads to improved generalisation will likely depend upon the problem setting. 
\tikzset{mylabel/.style  args={at #1 #2  with #3}{
    postaction={decorate,
    decoration={
      markings,
      mark= at position #1
      with  \node [#2] {#3};
 } } } }
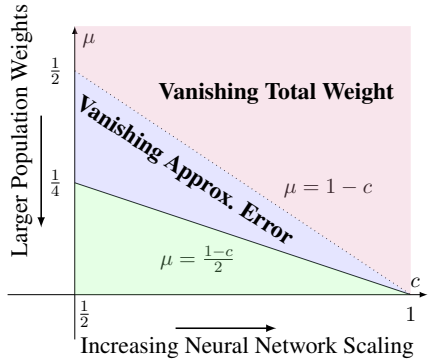
\begin{figure}[!h]
\centering
\resizebox{0.35\textwidth}{!}{
\centering
    \begin{tikzpicture}[scale=8,>=latex,x=1.5cm]
    \begin{scope}[domain=0:1]
            \draw[->,thick] (0.45,0.4) -- (0.45,0.15) node[xshift = -2ex,yshift = 10ex,rotate =90,thick,font=\large]{ Larger Population Weights };
            \draw[->,thick,font=\large] (0.65,-0.075) -- (0.8,-0.075) node[xshift = -3ex,yshift = -2ex,rotate =0,thick]{ Increasing Neural Network Scaling };
            \draw[->] (0.4,0) -- (1.025,0) node[above left]{ $c$};
            \draw[->] (0.5,-0.1) -- (0.5,0.6) node[below right]{ $\mu$};
            \draw[-,dotted,domain=0.5:1,samples=250,mylabel=at 0.6 above right with {$\mu = 1-c $}] plot (\x,{1-\x});
            \draw[-,domain=0.5:1,samples=250,mylabel=at 0.5 below left with {$\mu = \frac{1-c}{2} $}] plot (\x,{(1-\x)/2});
            \fill[blue!40,nearly transparent] (0.5,0.5) -- (0.5,0.25) -- (1,0);
            \fill[purple!40,nearly transparent] (0.5,0.5) -- (0.5,0.6) -- (1,0.6)-- (1,0);
            \fill[green!40,nearly transparent] (0.5,0.25) -- (0.5,0) -- (1,0) ;
            \node at (1,0)  [yshift=-2ex] { $1$};
            \node at (0.5,0.25)  [xshift=-2ex] { $\frac{1}{4}$};
            \node at (0.5,0.5)  [xshift=-2ex] { $\frac{1}{2}$};
            \node at (0.5,0)  [yshift=-2ex,xshift =1ex] { $\frac{1}{2} $};
            \node at (0.8,0.45) [rotate=0,font=\large] {\textbf{Vanishing Total Weight}};
            \node at (0.6625,0.275) [rotate=-33,font=\large] {\textbf{Vanishing Approx. Error}};
            \node at (0.6,0.1) [rotate=-25] {};
        \end{scope}
    \end{tikzpicture}
    }
    \caption{Weight Assumption \ref{ass:StatisticalAssumption} $\mu$ versus network scaling $c$. \textit{Red Region} (\textbf{Vanishing Total Weight}): values of $c$ and $\mu$ with Total Weight decreasing in $M$. \textit{Blue region} (\textbf{Vanishing Approx. Error}): values of $c$ and $\mu$ with approximation error decreasing in $M$. \textit{Solid line and green region}: values of $c$ and $\mu$ where our bounds do not guarantee approximation error decreasing in $M$.
    }
    \label{fig:Stat:vs:Scale}
\end{figure}

\begin{remark}[Limitation of Weak Convexity]
\label{remark:LimitationWeakConvexity}
Consider optimising both the first and second layers for a Two Layer Neural Network and let $\omega^{\star} = (A^{\star},v^{\star})$ denote a population risk minimiser. Then the \textbf{Approx. Error} in Theorem \ref{thm:TestError:Simple} is  $O( (\| A^{\star}\|_F^2 + \|v^{\star}\|_2^2)/M^c)$, while the Total Weight is upper-bounded (from Young's inequality)  as $\text{TW}(f) \leq (\| A^{\star}\|_F^2 + \|v^{\star}\|_2^2)/M^c$. It is therefore not possible in this case for the \textbf{Approx. Error} to decrease without the Total Weight vanishing. To remedy this, we show in Appendix \ref{app:GenWeakConv} that a generalised weak convexity assumption can yield an \textbf{Approx. Error} that can alternatively be upper bounded by the Total Weight.
\end{remark}

\subsection{Experimental Results}
\label{sec:Experiments}
\begin{figure}[!h]
    \centering
    \includegraphics[width=0.2125\textwidth]{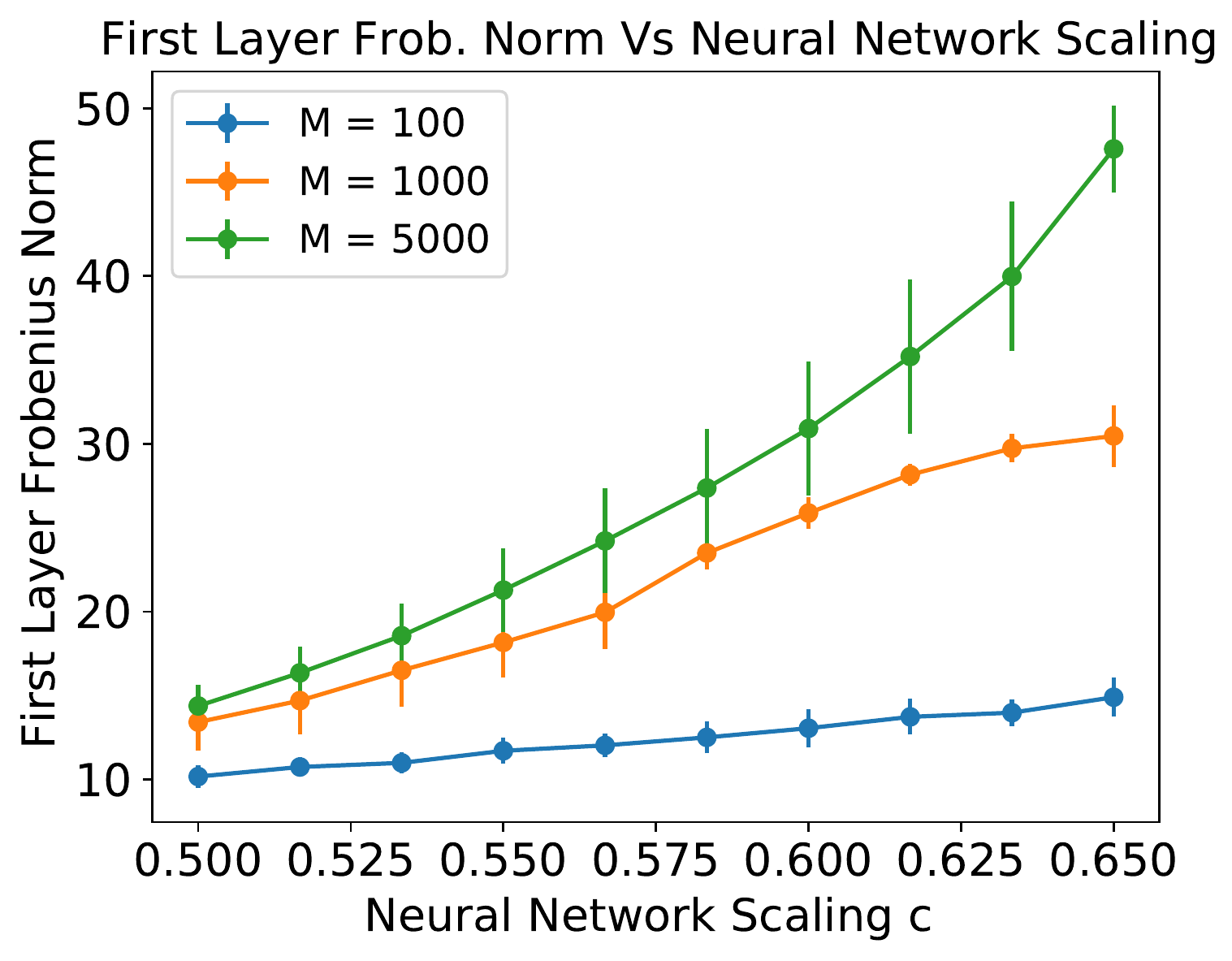}
    \includegraphics[width=0.22\textwidth]{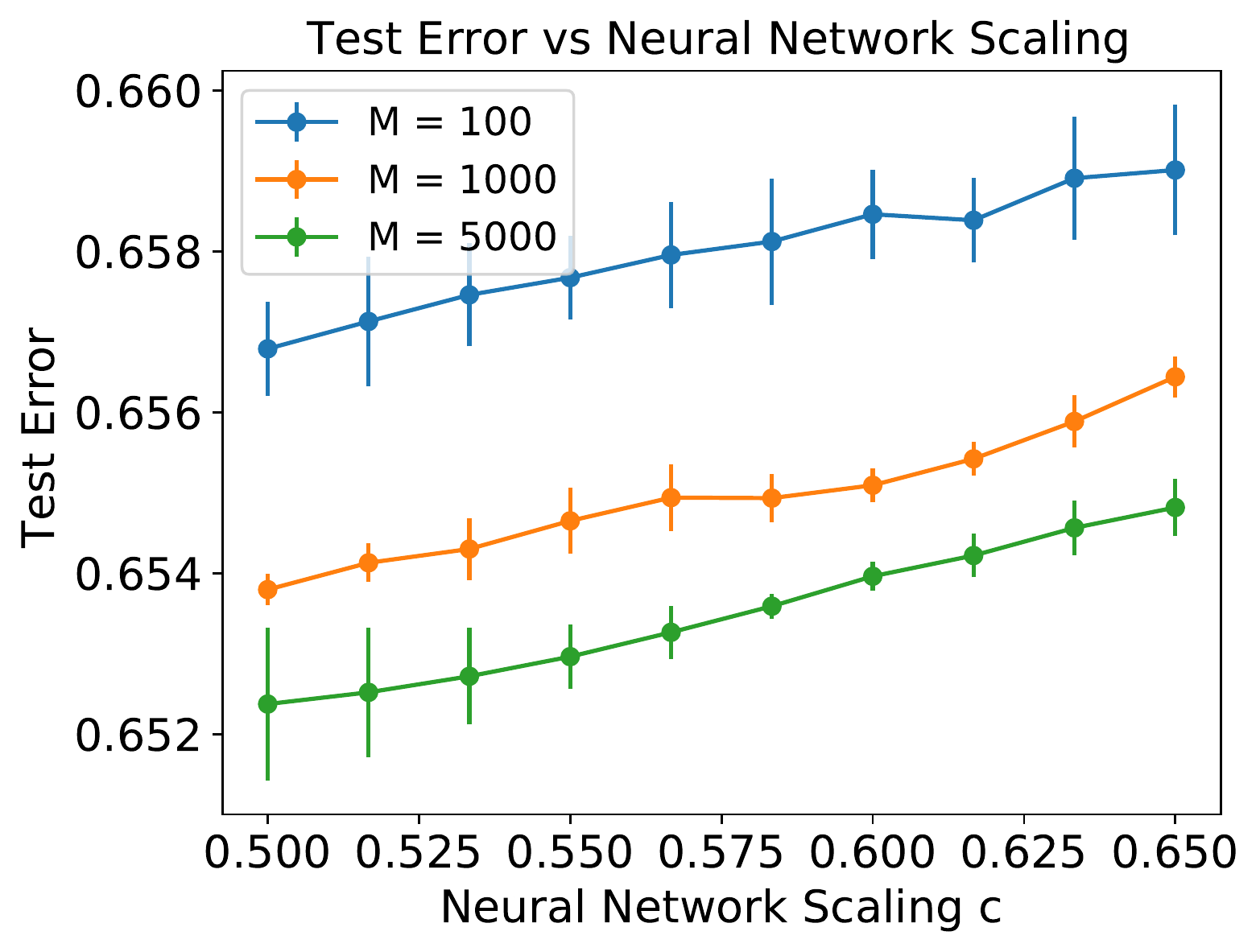}\\
    \includegraphics[width=0.21\textwidth]{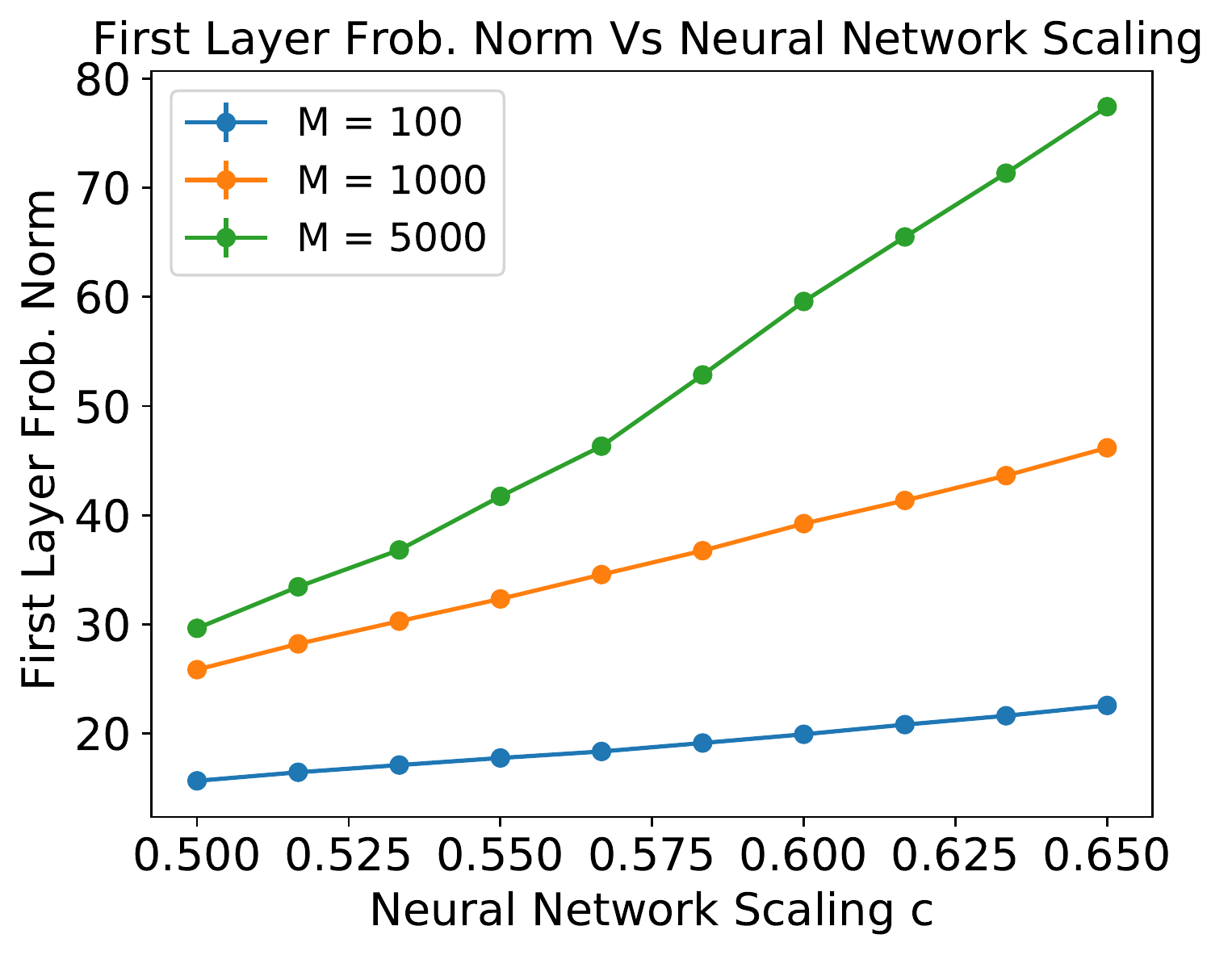}
    \includegraphics[width=0.22\textwidth]{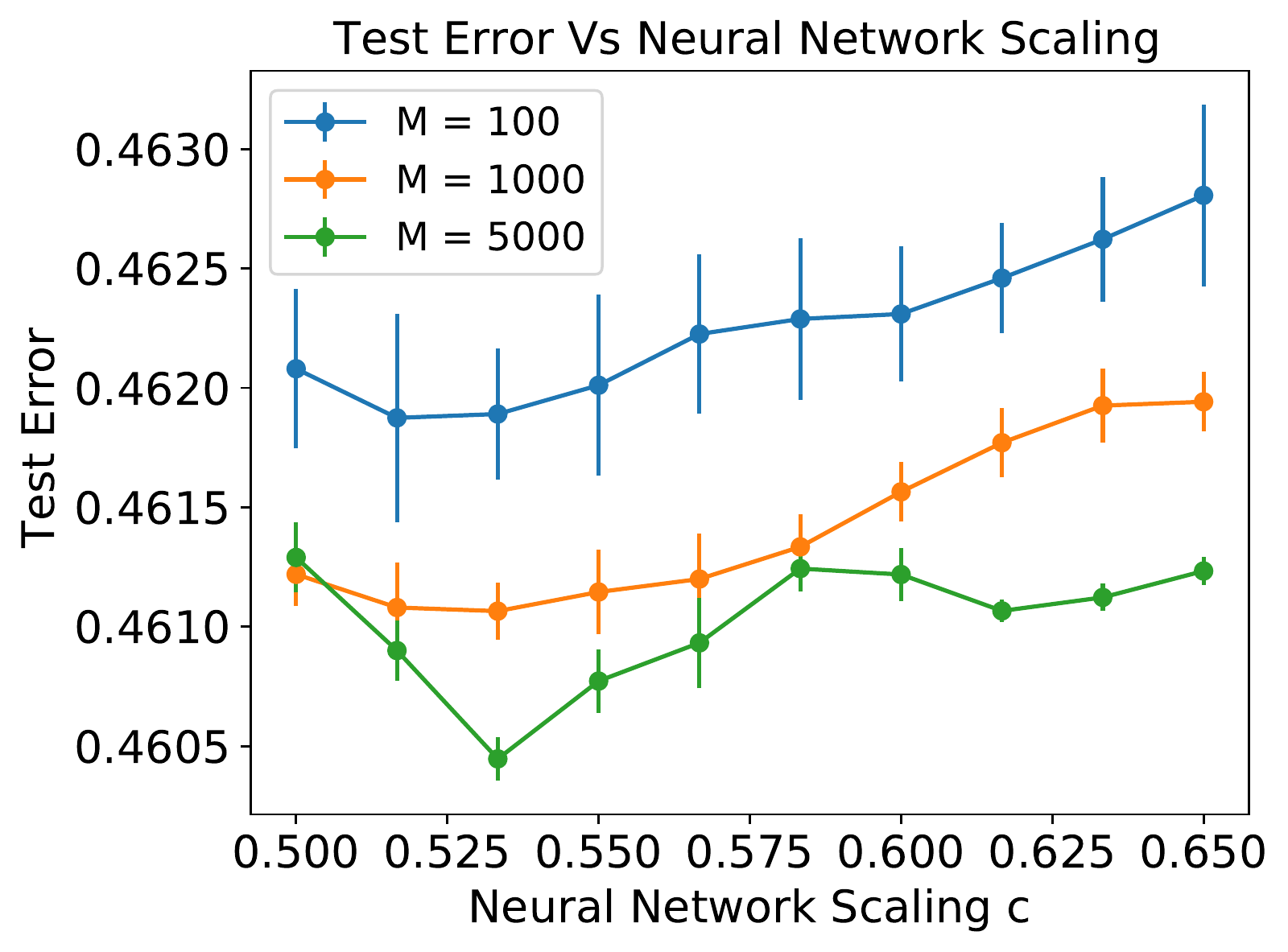}\\
    \includegraphics[width=0.22\textwidth]{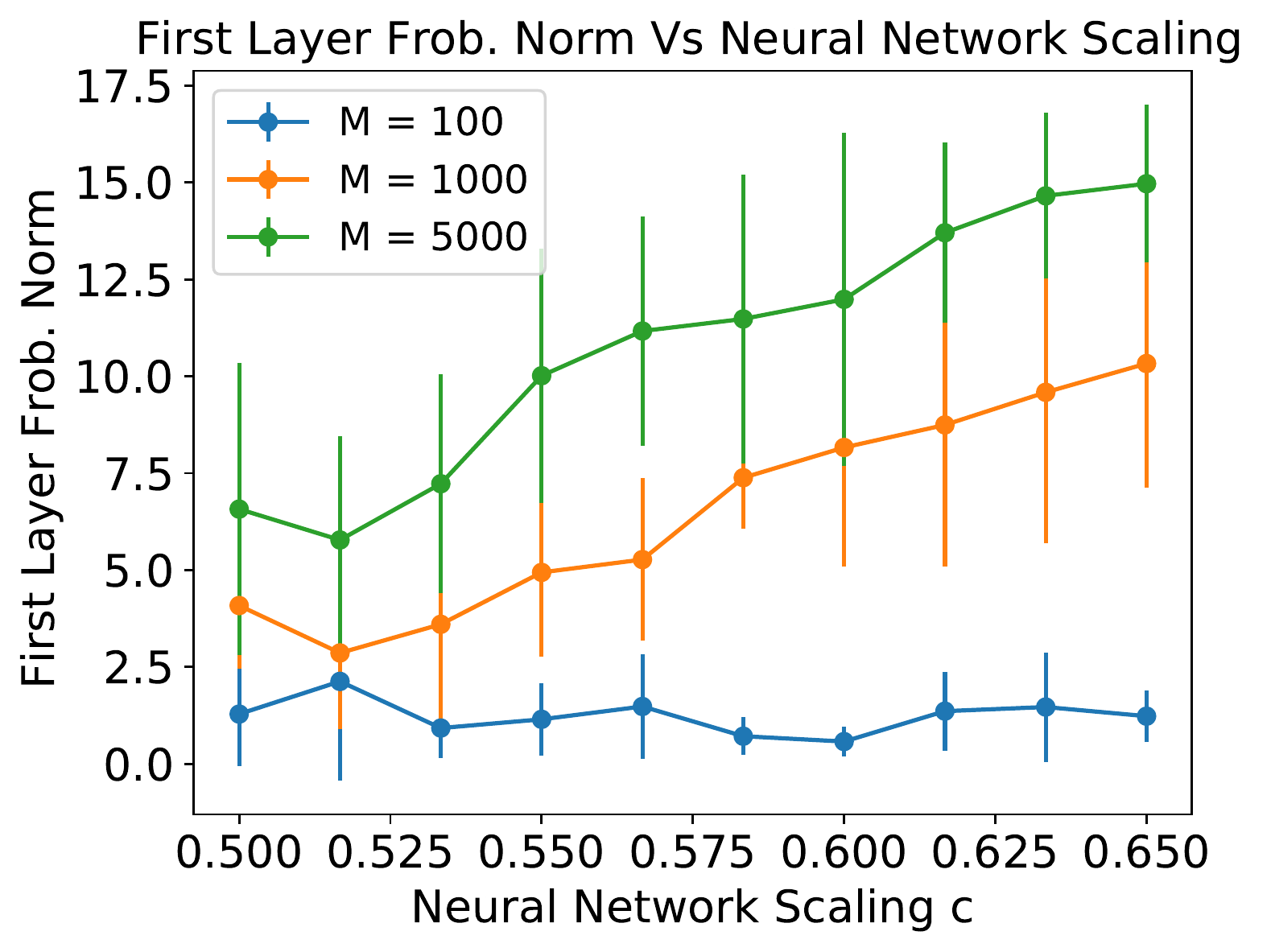}
    \includegraphics[width=0.22\textwidth]{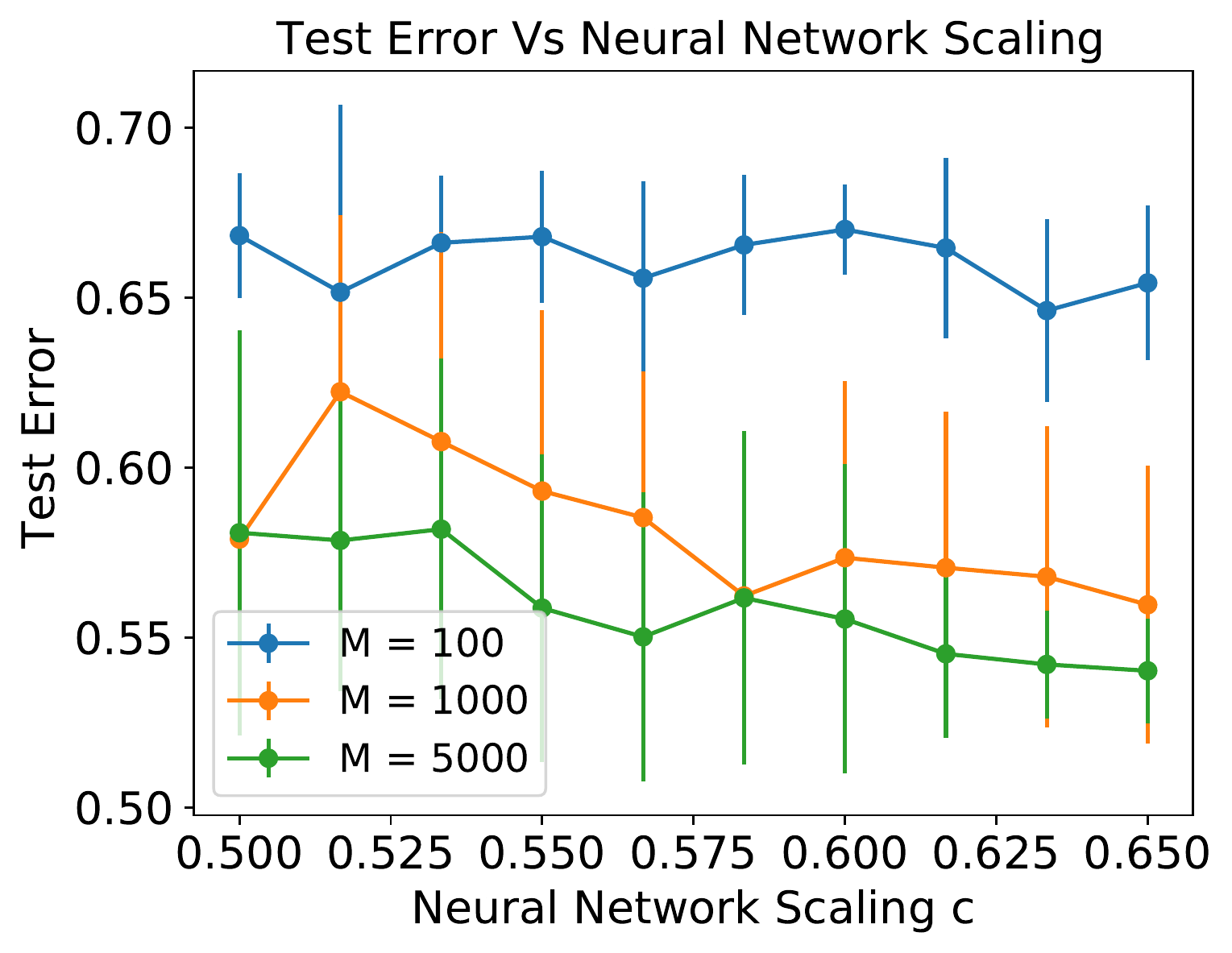}
    \caption{Plot of (\textit{Left}) first layer Frobenius norm $\|A\|_F$ and (\textit{Right}) population risk (labelled test error) versus network scaling $c$, for datasets: HIGGS (\textit{Top}), SUSY (\textit{Middle}) and COVTYPE (\textit{Bottom}). \textit{Top and Middle}: Step size $\eta = 0.1$, train size $N=10^{3}$, test set size $10^{5}$. Error bars from 2 subsets of data, each replicated 10 times. \textit{Bottom}: step size $\eta = 10^{-3}$, train size $N= 10^{4}$ test set size $4 \times 10^{4}$. Error bars from 10 replications with single subset of data.  }
    \label{fig:Experiments}
\end{figure}

In this section we present experiments supporting the discussion in section \ref{sec:TwoLayerNN:Approx}. We consider a classification task on subsets of 3 datasets: HIGGS, SUSY \cite{baldi2014searching} and COVTYPE \cite{blackard1999comparative}, all of which can be found on the UCI Machine Learning Repository \cite{Dua:2019}. The loss function is as in Section \ref{sec:TwoLayerNN:Setup} with $g$ being the logistic loss, and $f$ a two layer neural network with sigmoid activation $\sigma$ and both layers being optimised. Full batch gradient descent is performed on the $N$ samples with a fixed step size, and the population risk is estimated every $500$ iterations with training stopped once it increased consecutively for more than $5$ batches of $500$ iterations. Due to performing full batch gradient descent, we considered $c \in  [0.5,0.65]$ as the iterations required increased with the scaling $c$. The network was initialised with the first layer at $0$ and the second layer from a standard Gaussian. Figure \ref{fig:Experiments} then plots the Frobenius norm of the first layer and population risk (Test Error) against the neural network scaling $c$. 
Observe across the three data sets that the Frobeinus norm of the first layer is positively correlated with the scaling $c$, supporting the theoretical results discussed in Section \ref{sec:TwoLayerNN:Approx}. Moreover, note the estimated population risk decreases with the width of the neural network $M$.

\section{Conclusion}
In this work we have investigated the stability of gradient descent under a notation of weak convexity. Generalisation error bounds were proven that both, hold under milder assumptions on the step size when compared to previous works \cite{Hardt:2016:TFG:3045390.3045520,kuzborskij2017data}, and can be combined with optimisation and approximation error bounds to achieve guarantees on the test error. In the case of a two layer neural networks, we demonstrated that the network scaling can control the weak convexity parameter, and thus, allow test error bounds to be achieved when a complexity assumption is placed on the population risk minimiser. 
Moving forward it would be natural to extend to stochastic and accelerated gradient methods, as well as deeper neural networks.

\section{Acknowledgements}
We would like to thank Levent Sagun, Aaron Defazio, Yann Ollivier, 
Sho Yaida, Patrick Rebeschini, Tomas Va\v{s}kevi\v{c}ius and Tyler Farghly for their feedback and suggestions.

\bibliographystyle{plain}
\bibliography{References}

\newpage
\appendix

\section{Generalised Weak Convexity and Two Layer Neural Networks}
\label{app:GenWeakConv}
In this section we generalise the notation of weak convexity considered within the main body of the manuscript. This section proceeds as follows. Section \ref{sec:GeneralisedWeakConvexity} presents the generalised notation of weak convexity. Section \ref{sec:TwoLayerNN:Approx:FirstLayer} gives bounds on the \textbf{Optimisation \& Approximation Error} for two layer neural networks utilising generalised weak convexity. 

\subsection{Generalised Weak Convexity}
\label{sec:GeneralisedWeakConvexity}
As highlighted in Remark \ref{remark:LimitationWeakConvexity}, a drawback of Theorem \ref{thm:TestError:Simple} when optimising both layers for two layer neural network is the \textbf{Approximation Error}'s dependence on the squared Euclidean norm of the population risk minimiser $\|\omega^{\star}\|_2^2$. This stems from the weak convexity assumption (Assumption \ref{ass:WeakConvexity:Global}) penalising all co-ordinates equally i.e. adding $\|\omega\|_2^2$. To improve upon this we introduce the following assumption which aims to better encode the Hessian's structure. 
\begin{assumption}
\label{ass:WeakConvexity:Local}
For a convex set $\mathcal{X} \subseteq \mathbb{R}^{p}$ there exists non-negative function $G_{\mathcal{X}}: \mathbb{R}^{p} \rightarrow \mathbb{R}$ such that almost surely
\begin{align*}
   u^{\top}\nabla^2 R(\omega) u & \geq - G_{\mathcal{X}}(u) \quad\quad \text{ for any } u \in \mathbb{R}^{p} , \omega \in \mathcal{X}
\end{align*}
\end{assumption}
Assumption \ref{ass:WeakConvexity:Local} is a weakening of the weak convexity Assumption \ref{ass:WeakConvexity:Global} in two respects. The first is that we have restricted ourselves to a convex set $\mathcal{X}$ while Assumption \ref{ass:WeakConvexity:Global} considers case $\mathcal{X} = \mathbb{R}^{p}$. The second difference, is that the quadratic form of the Hessian with vectors $u$ is lower bounded by a function $G_{\mathcal{X}}(u)$. This function can then encode additional structure of the Hessian which can be utilised to obtain tighter control on the \textbf{Approximation Error}. We note by considering the function $G_{\mathcal{X}}(u) = \frac{\epsilon_{\mathcal{X}}}{2} \|u\|_2^2 $ for some $\epsilon_{\mathcal{X}} \geq 0$, we recover Assumption \ref{ass:WeakConvexity:Global} restricted to the convex set $\mathcal{X}$. Meanwhile more generally, $G_{\mathcal{X}}(u)$ may depend upon other norms which penalise subsets of co-ordinates as well as interactions. Given this assumption, we are present the following proposition which bounds the  \textbf{Optimisation \& Approximation Error} that appears within the test error decomposition \eqref{equ:TestError:Decomp}. 
\begin{proposition}[\textbf{Opt. \& Approx. Error}]
\label{prop:TestError:General}
Consider Assumption \ref{ass:LossReg} and \ref{ass:WeakConvexity:Local}, step size $\eta_{s}= \eta$ for all $s \geq 1$ and $\widetilde{\omega} \in \mathcal{X}$. Suppose that $\omega_{s} \in \mathcal{X}$ for all $s \geq 1$. If $\eta \beta \leq 1/2$
then 
\begin{align*}
     \underbrace{ \E_{I}[ E[ R(\widehat{\omega}_{I}]]  - r(\omega^{\star})}_{\textbf{Opt. \& Approx. Error}}
     \leq
    \underbrace{ \frac{\E[ \|\widehat{\omega}_0 - \widetilde{\omega}\|_2^2 ]}{2 \eta t } }_{\textbf{Optimisation Error}}
    + 
    \underbrace{ \frac{1}{2}\frac{1}{t} \sum_{s=1}^{t} \E[G_{\mathcal{X}}(\widetilde{\omega} - \widehat{\omega}_{s})] +  \E[R(\widetilde{\omega})] - r(\omega^{\star}) }_{\textbf{Approximation Error}}
\end{align*}
\end{proposition}
Observe that within Proposition \ref{prop:TestError:General} the bound now depends on the Hessian structure through $G_{\mathcal{X}}(\cdot)$ as well as now a free parameter $\widetilde{\omega} \in \mathcal{X}$. The right most term can then be interpreted as an \textbf{Approximation Error}, as the series involving $G_{\mathcal{X}}(\widetilde{\omega}-\widehat{\omega}_s)$ can be viewed as a regulariser, with $\widetilde{\omega}$ chosen thereafter to minimise a penalised empirical risk. Although, we note that the regulariser depends upon the iterates of gradient descent i.e. the difference $\widetilde{\omega} - \widehat{\omega}_{s}$, as such, to proceed we must decouple $\widetilde{\omega}$ and $\widehat{\omega}_{s}$ by utilising the structure of $G_{\mathcal{X}}(\cdot)$. We therefore now consider the case of a two layer neural network.

\subsection{Optimisation and Approximation Error for Two Layer Neural Networks} 
\label{sec:TwoLayerNN:Approx:FirstLayer}
In this section we demonstrate how the generalised notation of weak convexity just described in Section \ref{sec:GeneralisedWeakConvexity} can be applied to a two layer neural network when optimising both layers. Specifically, we consider the setting described in Section \ref{sec:TwoLayerNN:Setup}, where the loss is a composition of a convex function $g(\cdot,y)$ and a Two Layer Neural network $f$. Let us begin by introducing some additional notation. For a vector $z \in \mathbb{R}^{q}$ let $\text{Diag}(z) \in \mathbb{R}^{q \times q}$  be the square diagonal matrix where $\text{Diag}(z)_{ii} = z_i$ for $i=1,\dots,q$. Moreover, let us denote the vector $u= (A(u),v(u)) \in \mathbb{R}^{Md +M}$ which is composed in a manner matching  $\omega= (A,v)$, so that $A(u) \in \mathbb{R}^{M \times d}$ is a matrix associated to the first layer of weights and $v(u) \in \mathbb{R}^{M}$ is the vector associated to the second layer of weights. Let us denote the maximum between two real numbers $a,b \in \mathbb{R}$ as $a \vee b = \max\{a,b\}$.

To consider the generalised weak convexity Assumption~\ref{ass:WeakConvexity:Local} we first introduce the appropriate set $\mathcal{X}$ and functional $G_{\mathcal{X}}$. For $L_{v} \geq 0$ consider the set $\mathcal{X}_{L_{v}} \subseteq \mathbb{R}^{Md + d}$ as well as  for $C \geq 0$ the function $G_{C}: \mathbb{R}^{p} \rightarrow \mathbb{R}^{+}$. The set is then defined as $\mathcal{X}_{L_{v}}  := \{\omega \in \mathbb{R}^{Md + d}: \omega = (A,v), \|v\|_{\infty} \leq L_{v} \}$, while the functional is defined for $u= (A(u),v(u)) \in \mathbb{R}^{Md +M}$  as
\begin{align*}
    &G_{C}(u)  : = \frac{C}{M^c} \|A(u)\|_F^2  + 
    \frac{C}{M^c} 
    \max_{ \|z\|_{\infty} \leq 1}
    \big\| A(u)^{\top} \text{Diag}(z) v(u) \big\|_2.
\end{align*}
The set $\mathcal{X}_{L_{v}}$ in our case is technical and is included to, in short, control the second layer weights that arises within second derivative of first layer, see also Theorem \ref{thm:WeakConvexity:SingleLayerNN:Both}. Meanwhile, the function $G_{C}$ now incorporates the structure of the empirical risk Hessian. In particular, there is no term depending on the squared norm $\|v\|_2^2$, since the Hessian restricted to the second layer is zero. The second term in $G_{C}(\cdot)$ then arises to control the interaction between the first and second layers, with the maximum over vectors $\|z\|_{\infty} \leq 1$ coming from the activation $\sigma(\cdot)$. 

Given the function $G_{C}$ we must now introduce a related function $H:\mathbb{R}^{Md + M} \rightarrow \mathbb{R}$ which encodes the structure of regularisation. It is similarly defined as follows  
\begin{align*}
    H(u) := 
    \| A(u)\|_F^2  + \sqrt{ \eta t }\|v(u)\|_2 
    + 
    \max_{ \|z\|_{\infty} \leq 1  } \| A(u)^{\top}\text{Diag}(z) v(u)\|_2. 
\end{align*}
Note the structure of $H(\cdot)$ closely aligns with $G_{C}(\cdot)$ although an extra term $\sqrt{ \eta t }\|v(u)\|_2$ is included. This arises due to the coupling between the gradient descent iterates and place holder $\widetilde{\omega}$ i.e. $G_{C}(\widetilde{\omega} - \widehat{\omega}_{s})$. Given this, it will be convenient to denote for $\lambda \geq 0$ a minimiser $\widehat{\omega}_{\lambda} \in \argmin_{\omega \in \mathcal{X}_{L_{v}}}\big\{ R(\omega) + \lambda H(\omega - \widehat{\omega}_0) \big\}$. The following theorem then utilises Proposition \ref{prop:TestError:General} to bound the \textbf{Optimisation \& Approximation Error} within the test error decomposition \eqref{equ:TestError:Decomp}.  
\begin{theorem}[Two Layer Neural Network]
\label{thm:WeakConvexityError:TwoLayer}
Consider loss regularity Assumption \ref{ass:LossReg} alongside the setting of Theorem \ref{thm:WeakConvexity:SingleLayerNN:Both} with bounded activation $|\sigma(\cdot)| \leq L_{\sigma}$. Initialise gradient descent at $\widehat{\omega}_0 = (A^{0},v^{0}) \in \mathbb{R}^{Md + M}$. If the following holds almost surely 
\begin{align*}
    L_{v} & \geq \|v^{\star}\|_{\infty} \vee \Big( \|v^{0}\|_{\infty} + \frac{\eta t L_{g^{\prime}}L_{\sigma}}{M^c} \Big)\\
    C & \geq 2 L_{g^{\prime}}  
\Big[ \Big(  L_{\sigma^{\prime}} \sqrt{\trace\big( \widehat{\Sigma} \big)} \Big) \vee ( L_{\sigma^{\prime\prime}} L_{v} \|\widehat{\Sigma}\|_2 ) \Big],   
\end{align*}
then the optimisation and approximation error is bounded
\begin{align*}
\underbrace{ \E_{I}[ \E[ R(\widehat{\omega}_{I}]] \! - \! r(\omega^{\star})}_{\textbf{Opt. \& Approx. Error}}
  \leq  \underbrace{ \frac{\E[ \|\widehat{\omega}_{\lambda}  - \widehat{\omega}_0\|_2^2 ]}{2\eta t } }_{\textbf{Opt. Error}}  + \underbrace{ \frac{3C \E[ R(\omega_0)  -  R(\widehat{\omega}^{\star})]   \eta t }{M^c}   }_{\textbf{Opt. Component. }} + 
\underbrace{ 
\E[\lambda] H(\omega^{\star} - \widehat{\omega}_0) }_{\textbf{Stat. Approx.}}
\end{align*}
where $\lambda = 3 C (\sqrt{ R(\widehat{\omega}_0) - R(\widehat{\omega}^{\star}) } \vee 1)/M^c$.
\end{theorem}
Observe in Theorem \ref{thm:WeakConvexityError:TwoLayer} that the parameter $L_v$, which controls the constraint set $\mathcal{X}_{L_{v}}$,  is required to 
grow as $O( 1 + \eta t /M^c)$, and therefore, is constant for sufficiently wide neural networks (see also discussion in Section \ref{sec:TestErrorWeaklyConvex}). Meanwhile, the parameter $C$, which controls the functional, must almost surely upper bound spectral quantities related to the covariates covariance, namely, the trace $\trace\big(\widehat{\Sigma}\big)$. This can then be more refined than assuming almost surely bounded co-ordinates i.e. $\|x\|_{\infty}$, as done within the main body of the manuscript. 
The resulting bound on the \textbf{Optimisation and Approximation Error} then consists of two new terms: \textbf{Opt. Component} and \textbf{Stat. Approx.}. The \textbf{Opt. Component.} is order $O(d \eta t / M^c)$ and arises from the dependence on gradient descent iterates $\widehat{\omega}_{s}$. Meanwhile, the \textbf{Stat. Approx.} depends upon the population risk minimiser evaluated at the regulariser i.e. $\E[\lambda] H(\omega^{\star}-\widehat{\omega}_{0})$, and can be interpreted as a statistical bias resulting from the non-convexity. Note it now depends upon $H(\cdot)$ and is scaled by $\lambda$ which is $O\Big(\sqrt{\trace(\widehat{\Sigma})}/M^c\Big)$.

It is natural to investigate (picking $\widehat{\omega}_0 = 0$) the size of the \textbf{Stat. Approx.} for a particular problem instance. Following the limitations of the weakly convex setting highlighted within remark \ref{remark:LimitationWeakConvexity} in the main body of the manuscript, we focus on whether \textbf{Stat. Approx.} can be smaller than the Total Weight of the network $\text{TW}(f)$. Denoting $\omega^{\star} = (A^{\star}, v^{\star})$ a minimiser of the population risk, we focus on upper bounding  $\sqrt{\trace(\widehat{\Sigma})} H(\omega^{\star})/M^c$ by the Total Weight of the network, since  $\sqrt{\trace(\widehat{\Sigma})} H(\omega^{\star})/M^c$ equals the \textbf{Stat. Approx.} up to constants

Begin by noting that $H(\cdot)$ depends upon the interaction between the first and second layer, therefore, introduce the single valued decomposition 
$A^{\star} = \Gamma \Lambda \Xi^{\top} =   \sum_{\ell=1}^{ d } \lambda_{\ell} \gamma_{\ell} \xi_{\ell}^{\top}$ where left-singular vectors $\{\gamma_{\ell}\}_{\ell=1}^{d}$ are the columns of $\Gamma \in \mathbb{R}^{M \times d}$, the right-singular vectors $\{\xi\}_{\ell=1}^{d}$ are the columns of $\Xi \in \mathbb{R}^{d \times d}$ and $\{\lambda_i\}_{i=1}^{d}$ are the singular values.  Let us also denote the element wise multiplication of two vectors $u,v \in \mathbb{R}^{p}$ as $u \odot v = (u_1 v_1,\dots,u_p v_p) \in \mathbb{R}^{p}$. With the single valued decomposition the regulariser can then be written as follows
\begin{align*}
    \frac{1}{M^c} H(\omega^{\star}) 
     = 
    \underbrace{ \frac{1}{M^c} \big( \|A^{\star}\|_{F}^2  + \sqrt{ \eta t} \|v^{\star}\|_2 \Big)}_{\textbf{Norm Condition}}   +  
    \underbrace{ 
    \max_{\|z\|_{\infty} \leq 1} \frac{1}{M^c} 
    \sqrt{ \sum_{j=1}^{d}|\lambda_j|^2 |\langle z, v^{\star} \odot \gamma_j \rangle|^2}}_{\textbf{Interaction}}.
\end{align*}
The first two terms above depend upon the norm of the population risk minimiser $\omega^{\star}$, while the third term now encodes the interaction between the first and second layer. For clarity, let $a_1,a_2 \in \mathbb{R}$ and assume each of the second layers weights are the same magnitude so $|v^{\star}_j| = a_1 $ for $j=1,\dots,M$ as well as the singular values so $|\lambda_j| = a_2$ for $j=1,\dots,d$. The \textbf{Interaction} term can then be upper bounded by taking the maximum $\max_{\|z\|_{\infty}\leq 1}$ inside the series 
\begin{align*}
    \textbf{Interaction}= 
    \frac{1}{M^c} \max_{\|z\|_{\infty} \leq 1} \sqrt{ 
     \sum_{j=1}^{d}|\lambda_j|^2 |\langle z, v^{\star} \odot \gamma_j \rangle|^2 } 
     \leq 
    \frac{1}{M^c}\sqrt{ \sum_{j=1}^{d} |\lambda_j|^2 \|v^{\star} \odot \gamma_j\|_1^2}
    = 
    a_1 a_2 \frac{1}{M^c}\sqrt{  \sum_{j=1}^{d}\|\gamma_j\|_1^2}.
\end{align*} 
Note that this quantity now aligns with the $\ell_{1,2}$ element-wise matrix norm on the left-singular vectors of $A^{\star}$ i.e. $\Gamma$. Therefore, let us assume that the left-singular vectors $\{\gamma_i\}_{i=1}^{d}$ are supported on disjoint sets of size $M/d \geq s \geq 1 $ and are such that $|(\gamma_i)_{j}| = \frac{1}{\sqrt{s}}$ for $i=1,\dots,d$, $j \in \text{Supp}(\gamma_{i})$ (where $\text{Supp}(\cdot)$ denotes the support of a vector) and zero otherwise. Noting that $\|v^{\star}\|_2 = a_1 \sqrt{ds}$ since the second layer of weights will be supported on the non-zero rows of $A^{\star}$ as well as that $\|\gamma_i\|_1 = \sqrt{s}$, yields the upper bound on the \textbf{Stat. Approx.} term
\begin{align*}
    \frac{\sqrt{ \trace\big(\widehat{\Sigma} \big) } H(\omega^{\star})}{M^c} 
    \leq 
    \sqrt{ \trace\big(\widehat{\Sigma} \big) }
    \Big( \underbrace{ \frac{d a_2}{M^c} + \frac{a_1 \sqrt{\eta t} \sqrt{d s}}{M^c} }_{\textbf{Norm Condition}} + 
    \underbrace{ \frac{a_1 a_2 \sqrt{ds}}{M^c} }_{\textbf{Interaction}} \Big).
\end{align*}
Let us now consider the Total Weight with this particular choice of weights $\omega^{\star} = (v^\star,A^{\star})$. Precisely, for this particular choice of second layer weights $v^{\star}$, singular values $\{\lambda_j\}_{j=1}^{d}$ and singular vectors $\{\gamma_i\}_{i=1}^{d}$, the Total Weight aligns with the $\ell_{2,1}$ element-wise matrix norm of $\Gamma^{\top}$
\begin{align*}
    TW(f) = \frac{1}{M^c} \sum_{j=1}^{M} |v_j^{\star} \|A^{\star}_j\|_2 
    =  \frac{a_1 a_2 }{M^c} \sum_{j=1}^{M} \sqrt{\sum_{i=1}^{d} (\gamma_{i})_{j}^2}
    = 
    \frac{a_1 a_2 }{M^c} \sum_{j=1}^{d \times s} \sqrt{ \frac{1}{s}}
    = a_1 a_2 \frac{d \sqrt{s}}{M^c}.
\end{align*}
Where we note that the first equality arises from the single valued decomposition $A^{\star}_j = \sum_{i=1}^{d} \lambda_{i} (\gamma_{i})_j \xi_i $ and thus $\|A_j^{\star}\|_2 = \sqrt{ \sum_{i=1}^{d} \lambda_i^2  (\gamma_{i})_{j}^2 }$. Meanwhile for the second equality, for each $j=1,\dots,d \times s$ we have $j \in \text{Supp}(\gamma_k)$ for at most one $k \in \{1,\dots,d\}$ since $\{\gamma_i\}_{i=1}^{d}$ are supported on disjoint sets. Meanwhile for $j=d\times s + 1,\dots,M$ we have $j \not\in \text{Supp}(\gamma_k)$ since the supports are of size at most $s$. This means for $j=1,\dots,d\times s$ we get $\sum_{i=1}^{d} (\gamma_i)^2_{j} = \sum_{i: j \in \text{Supp}(\gamma_i)}(\gamma_i)^2_{j}  =   \frac{1}{s} $, and then zero otherwise.  Dividing the \textbf{Stat. Approx.} by the Total Weight we have 
\begin{align*}
    \frac{1}{TW(f)} \frac{\sqrt{ \trace\big(\widehat{\Sigma} \big) } H(\omega^{\star})}{M^c}
    & = 
    \sqrt{\trace\big(\widehat{\Sigma} \big)} 
    \Big( \frac{1}{a_1 \sqrt{s}} + \frac{\sqrt{\eta t}}{a_2\sqrt{d}} + \frac{1}{\sqrt{d}} \Big).
\end{align*}
Now, if $s > d \geq 9$, $a_1,a_2 \geq 1$ and $ \eta t \trace(\widehat{\Sigma}) \leq d/9 $ then the \textbf{Stat. Approx.} error is upper bounded by the Total Weight $\sqrt{\trace(\widehat{\Sigma})} H(\omega^{\star})/M^c \leq TW(f)$, as required.

\section{Proof of Generalisation Error Bounds under Weak Convexity}
In this section we present the proofs related to the first half of the manuscript which gives generalisation error bounds for gradient descent under pointwise weak convexity Assumption \ref{ass:LocalWeakConvexity} and standard weak convexity Assumption \ref{ass:WeakConvexity:Global}. We begin by presenting the proof of Theorem \ref{thm:GenErrorBound} in Section \ref{sec:ProofGen:Thm}. Section \ref{sec:ProofGen:Lem} present the proof of Lemma \ref{lem:WeakCocoer} which is technical result used within the proof of Theorem \ref{sec:ProofGen:Thm}.  
Section \ref{sec:ProofGen:Global} then presents and proves a generalisation error bound for gradient descent under standard weak convexity Assumption \ref{ass:WeakConvexity:Global}. 

\subsection{Proof of Theorem \ref{thm:GenErrorBound}}
\label{sec:ProofGen:Thm}
In this section we give the proof of Theorem \ref{thm:GenErrorBound}. Using equation \eqref{equ:GenErrorStability} we can then bound the generalisation error for gradient descent in terms of the difference between gradient descent with and without a resampled datapoint. Specifically, using that the loss is  $L$-Lipschitz as well as Jensen's Inequality to take the absolute value inside the expectation we get 
\begin{align*}
    |\E[R(\widehat{w}_{t}) - r(\widehat{w}_{t})]|
    & \leq 
\frac{1}{N}\sum_{i=1}^{N} |\E[ 
  \ell(\widehat{w}^{(i)}_{t},Z_{i}^{\prime}) - \ell(\widehat{w}_{t},Z_{i}^{\prime}) ]|  \leq 
  \frac{L}{N}\sum_{i=1}^{N} \E[ \|\widehat{w}^{(i)}_{t} - \widehat{w}_{t}\|_2].
\end{align*}
For $i=1,\dots,N$ it then suffices to bound the deviation $\|\widehat{w}_t - \widehat{w}^{(i)}_{t}\|_2$. Using that the gradient of the empirical risk can be denoted $\nabla R^{(i)}(w) = \nabla R(w) + \frac{1}{N} ( \nabla \ell(w,Z_i^{\prime}) - \nabla \ell(w,Z_i))$ alongside the Lipschitz assumption we get for any $k \geq 1$,
\begin{align}
\label{equ:Deviation}
    \|\widehat{w}_k - \widehat{w}^{(i)}_{k}\|_2
    \leq 
    \|\widehat{w}_{k-1} - \widehat{w}^{(i)}_{k-1} - \eta_{k-1} 
    \big( \nabla R(\widehat{w}_{k-1}) - \nabla R(\widehat{w}^{(i)}_{k-1})\big)\|_2
    + \frac{2 \eta_{k-1} L}{N}.
\end{align}
The first term on the right hand side is then referred to as the expansiveness of the gradient update. Note, for $k=1$ it is zero since the iterates with and without the resampled data point are initialised at the same location $\widehat{\omega}_0 = \widehat{\omega}_0^{(i)}$, and thus,
\begin{align}
\label{equ:BaseCase}
    \|\widehat{w}_1 - \widehat{w}^{(i)}_{1}\|_2
    \leq 
    \frac{2 \eta_{0} L}{N}.
\end{align}
Therefore, let us consider the difference $\|\widehat{w}_k - \widehat{w}^{(i)}_{k}\|_2$ for $k \geq 2$. 
Expanding the expansiveness of the gradient update term we get  
\begin{align*}
    & \|\widehat{w}_{k-1} - \widehat{w}^{(i)}_{k-1} - 
    \eta_{k-1} \big( \nabla R(\widehat{w}_{k-1}) - \nabla R(\widehat{w}^{(i)}_{k-1})\big) \|_2^2\\
    & = 
     \|\widehat{w}_{k-1} - \widehat{w}^{(i)}_{k-1}\|_2^2 
     + \eta^2_{k-1} \|
     \nabla R(\widehat{w}_{k-1}) - \nabla R(\widehat{w}^{(i)}_{k-1}
     \|_2^2 
     - 2 \eta_{k-1} \langle \nabla R(\widehat{w}_{k-1}) - \nabla R(\widehat{w}^{(i)}_{k-1}), 
     \widehat{w}_{k-1} - \widehat{w}^{(i)}_{k-1}
     \rangle.
\end{align*}
Now we must lower bound $\langle \nabla R(\widehat{w}_{k-1}) - \nabla R(\widehat{w}^{(i)}_{k-1}), \widehat{w}_{k-1} - \widehat{w}^{(i)}_{k-1} \rangle$ utilising both the loss regularity (Assumption \ref{ass:LossReg}) and the pointwise weak convexity (Assumption \ref{ass:LocalWeakConvexity}). These steps are summarised within the following lemma. 
\begin{lemma}
\label{lem:WeakCocoer}
Consider assumptions \ref{ass:LossReg} and \ref{ass:LocalWeakConvexity}. Then for $s \geq 1$ and $\eta \geq 0$
\begin{align*}
     \langle \nabla R(\widehat{w}_{s}) - \nabla R(\widehat{w}^{(i)}_{s}), \widehat{w}_{s} - \widehat{w}^{(i)}_{s} \rangle
    & \geq 2 \eta \Big( 1 - \frac{\beta \eta}{2} \Big) \|\nabla R(\widehat{w}_{s}) - \nabla R(\widehat{w}^{(i)}_{s})\|_2^2 \\
    & \quad\quad 
    -  \big( \epsilon_{s} + \frac{2 \beta}{N} \big)  
    \|\widehat{w}_{s} - \widehat{w}^{(i)}_{s} - 
    \eta \big( \nabla R(\widehat{w}_{s}) - \nabla R(\widehat{w}^{(i)}_{s})\big) \|_2^2\\
    & \quad\quad -  \frac{\rho}{3} \|\widehat{w}_{s} - \widehat{w}^{(i)}_{s} - 
    \eta \big( \nabla R(\widehat{w}_{s}) - \nabla R(\widehat{w}^{(i)}_{s})\big) \|_2^3
\end{align*}
\end{lemma}
Utilising Lemma \ref{lem:WeakCocoer} with $s = k-1$ and $\eta = \eta_{k-1}$ the expansiveness of the gradient update term can then be upper bounded  
\begin{align*}
     & \|\widehat{w}_{k-1} - \widehat{w}^{(i)}_{k-1} - 
    \eta_{k-1} \big( \nabla R(\widehat{w}_{k-1}) - \nabla R(\widehat{w}^{(i)}_{k-1})\big) \|_2^2 \leq 
    \|\widehat{w}_{k-1} - \widehat{w}^{(i)}_{k-1}\|_2^2 \\
    & \quad\quad 
    + 
    \eta^2_{k-1} \big( 1 - 4 \big( 1 - \frac{\beta \eta_{k-1}}{2} \big) \big) \|\nabla R(\widehat{w}_{k-1}) - \nabla R(\widehat{w}^{(i)}_{k-1})\|_2^2\\
    & \quad\quad 
    + 2 \eta_{k-1} \big( \epsilon_{k-1} + \frac{2 \beta}{N} \big)  
    \|\widehat{w}_{k-1} - \widehat{w}^{(i)}_{k-1} - 
    \eta_{k-1} \big( \nabla R(\widehat{w}_{k-1}) - \nabla R(\widehat{w}^{(i)}_{k-1})\big) \|_2^2\\
    & \quad\quad + 2 \eta_{k-1} \frac{\rho}{3} \|\widehat{w}_{k-1} - \widehat{w}^{(i)}_{k-1} - 
    \eta_{k-1} \big( \nabla R(\widehat{w}_{k-1}) - \nabla R(\widehat{w}^{(i)}_{k-1})\big) \|_2^3.
\end{align*}
Note from assumptions within the theorem that $\eta_{k-1} \leq \frac{3}{2 \beta}$  so the second term on the right hand side is negative. Meanwhile, if we denote 
$\Delta(k) = \widehat{w}_{k-1} - \widehat{w}^{(i)}_{k-1} - \eta_{k-1} \big( \nabla R(\widehat{w}_{k-1}) - \nabla R(\widehat{w}^{(i)}_{k-1})\big)$, the third term can be bounded using Young's inequality $ab \leq \frac{1}{2} a^2 + \frac{1}{2} b^2 $ as
\begin{align*}
    2 \eta_{k-1} \frac{\rho}{3} \|\Delta(k)\|_2^3
    & =
    (\sqrt{2}\eta^{\frac{1}{2\alpha}}_{k-1} \|\Delta(k)\|_2) (  \eta^{1 - \frac{1}{2\alpha}}_{k-1} \frac{\sqrt{2}\rho}{3} \|\Delta(k)\|^2_2)\\
    & \leq \eta^{\frac{1}{\alpha}}_{k-1} \|\Delta(k)\|_2^2 
    + 
    \eta^{2(1 - \frac{1}{2\alpha})}_{k-1} \frac{\rho^2}{9} \|\Delta(k)\|^4_2.
\end{align*}
Collecting the squared terms and taking on the left hand side, the expansiveness of the gradient update term can then be upper bounded  
\begin{align}
\label{equ:ExpansiveBound}
    (1 - 2 \eta_{k-1}(\epsilon_{k-1} + \frac{2 \beta}{N}) - \eta^{\frac{1}{\alpha}}_{k-1} ) \|\Delta(k)\|_2^2 
    & \leq 
    \|\widehat{w}_{k-1} - \widehat{w}^{(i)}_{k-1}\|_2^2 
    + \eta^{2(1 - \frac{1}{2\alpha})} \frac{\rho^2}{9} \|\Delta(k)\|^4_2 \nonumber 
    \\
    & \leq 
    \|\widehat{w}_{k-1} - \widehat{w}^{(i)}_{k-1}\|_2^2 
    + 9 \eta^{2(1 - \frac{1}{2\alpha})}_{k-1}  \rho^2 \|\widehat{w}_{k-1} - \widehat{w}^{(i)}_{k-1}\|^4_2. 
\end{align}
Note that the second inequality above uses the Lipschitz property of the loss's gradient and that $\eta_{k-1} \beta \leq 3/2$ to say 
\begin{align*}
    \|\widehat{w}_{k-1} - \widehat{w}^{(i)}_{k-1} - 
    \eta_{k-1} \big( \nabla R(\widehat{w}_{k-1}) - \nabla R(\widehat{w}^{(i)}_{k-1})\big) \|_2
    & \leq 
    (1 + \eta_{k-1} \beta) \|\widehat{w}_{k-1} - \widehat{w}^{(i)}_{k-1}\|_2\\
    & \leq 
    3 \|\widehat{w}_{k-1} - \widehat{w}^{(i)}_{k-1}\|_2.
\end{align*}
Dividing both sides of \eqref{equ:ExpansiveBound} by $1 - 2 \eta_{k-1}(\epsilon_{k-1} + \frac{2 \beta}{N} ) - \eta^{\frac{1}{\alpha}}_{k-1}$, taking care this quantity is non-negative from an assumption within the theorem, applying the square root and plugging into \eqref{equ:Deviation} then yields the recursion 
\begin{align*}
    \|\widehat{w}_{k}  - \widehat{w}^{(i)}_{k}\|_2 
    & \leq \Big( \frac{1}{ 1 - 2 \eta_{k-1} ( \epsilon_{k-1} + \frac{2\beta}{N}) - \eta^{\frac{1}{\alpha}}_{k-1} } \Big)^{1/2}
    \|\widehat{w}_{k-1}  - \widehat{w}^{(i)}_{k-1}\|_2
    + \frac{2 \eta_{k-1} L}{N}\\
    & \quad\quad 
    +
    3 \eta^{1 - \frac{1}{2 \alpha}}_{k-1} \rho 
    \Big( \frac{ 1 }{1 - 2 \eta_{k-1} ( \epsilon_{k-1} + \frac{2\beta}{N}) - \eta^{\frac{1}{\alpha}}_{k-1} }\Big)^{1/2}
    \|\widehat{w}_{k-1} - \widehat{w}^{(i)}_{k-1}\|_2^{2}.
\end{align*}
Unravelling the iterates with the convention $\prod_{s=k}^{k-1}\big( \frac{1}{ 1 - 2 \eta_{s} ( \epsilon_{s} + \frac{2\beta}{N}) - \eta^{\frac{1}{\alpha}}_{s} }  \big)^{1/2}  = 1$ gives
\begin{align*}
\|\widehat{w}_{k}  - \widehat{w}^{(i)}_{k}\|_2 
    & \leq 
    \frac{2L}{N} \sum_{j=0}^{k-1} \prod_{s=j+1}^{k-1}\big( \frac{1}{ 1 - 2 \eta_{s} ( \epsilon_{s} + \frac{2\beta}{N}) - \eta^{\frac{1}{\alpha}}_{s} }  \big)^{1/2}\eta_{j}  \\
    & \quad\quad 
    +
    3\rho 
    \sum_{j=1}^{k-1} \prod_{s=j}^{k-1}\big( \frac{1}{ 1 - 2 \eta_{s} ( \epsilon_{s} + \frac{2\beta}{N})  - \eta^{\frac{1}{\alpha}}_{s} } \big)^{1/2} \eta^{1 - \frac{1}{2\alpha}}_{j}  \|\widehat{w}_{j} - w^{(i)}_{j} \|_2^{2}
\end{align*}
We must now bound the product of terms. Using the assumption within the theorem that $2 \eta_{s}(\epsilon_{s} + \frac{2\beta}{N}) + \eta^{\frac{1}{\alpha}}_{s} < 1/2$, as well as th inequality $1 + x \leq e^{x}$, we get the upper bound
\begin{align*}
    \sum_{j=0}^{k-1} \prod_{s=j+1}^{k-1}\big( \frac{1}{ 1 - 2 \eta_{s} ( \epsilon_{s} + \frac{2\beta}{N}) - \eta^{\frac{1}{\alpha}}_{s} }  \big)^{1/2}\eta_{j}  
    & = 
\sum_{j=0}^{k-1} \prod_{s=j+1}^{k-1}
\big( 1 + \frac{2 \eta_{s} ( \epsilon_{s} + \frac{2\beta}{N}) + \eta^{\frac{1}{\alpha}}_{s} }{ 1 - 2 \eta_{s} ( \epsilon_{s} + \frac{2\beta}{N}) - \eta^{\frac{1}{\alpha}}_{s} }  \big)^{1/2}\eta_{j}  \\
    & \leq 
    \sum_{j=0}^{k-1}
    \exp\big( 
    2  \sum_{s=j+1}^{k-1} \eta_s \epsilon_{s} + \frac{4 \sum_{s=j+1}^{k-1} \eta_{s} \beta}{N}
    +  \sum_{s=j+1}^{k-1} \eta^{\frac{1}{\alpha}}_{s}
    \big)\eta_j 
\end{align*}
where we have adopted the convention $\sum_{s=k}^{k-1} \eta_{s} = 0$. This then leads to following upper bound for $k \geq 2$
\begin{align}
\label{equ:InductiveArgument}
    \|\widehat{w}_{k}  - \widehat{w}^{(i)}_{k}\|_2 
    & \leq 
    \frac{2L}{N} \sum_{j=0}^{k-1}
    \exp\big( 
    2  \sum_{s=j+1}^{k-1} \eta_s \epsilon_{s} + \frac{4 \sum_{s=j+1}^{k-1} \eta_{s} \beta}{N}
    +  \sum_{s=j+1}^{k-1} \eta^{\frac{1}{\alpha}}_{s}
    \big)\eta_j 
    \\
    \nonumber
    & \quad\quad 
    +
    3\rho 
    \sum_{j=1}^{k-1} \prod_{s=j}^{k-1}\big( \frac{1}{ 1 - 2 \eta_{s} ( \epsilon_{s} + \frac{2\beta}{N})  - \eta^{\frac{1}{\alpha}}_{s} } \big)^{1/2} \eta^{1 - \frac{1}{2\alpha}}_{j}  \|\widehat{w}_{j} - w^{(i)}_{j} \|_2^{2}.
\end{align}
Observe that the above bound depends on higher order terms from previous time steps $\|\widehat{w}_{j} - w^{(i)}_{j} \|_2^{2}$. To control  $\|\widehat{w}_{k}  - \widehat{w}^{(i)}_{k}\|_2 $ for some $k \geq 2$, we now utilise the fact that the difference from earlier iterations $\|\widehat{w}_{j}  - \widehat{w}^{(i)}_{j}\|_2$ for $j=1,\dots,k-1$ can also be small. To this end, we use the upper bounds \eqref{equ:BaseCase} and \eqref{equ:InductiveArgument} to show inductively, under the assumptions of the theorem, that the following holds for $t \geq k \geq 2$ 
\begin{align*}
    \|\widehat{w}_{k}  - \widehat{w}^{(i)}_{k}\|_2 
    \leq 
    \frac{4L}{N}\sum_{j=0}^{k-1}
    \exp\big( 
    2  \sum_{s=j+1}^{k-1} \eta_s \epsilon_{s} + \frac{4 \sum_{s=j+1}^{k-1} \eta_{s} \beta}{N}
    +  \sum_{s=j+1}^{k-1} \eta^{\frac{1}{\alpha}}_{s}
    \big)\eta_j.
\end{align*}
Showing the above would then imply the bound presented within the theorem. 
Let us begin by proving the base case $k=2$. Looking to \eqref{equ:InductiveArgument} and plugging in the upper bound on $\|\widehat{w}_{1}  - \widehat{w}^{(i)}_{1}\|_2 \leq 2 \eta_0 L /N$ from \eqref{equ:BaseCase} yields 
\begin{align*}
    \|\widehat{w}_{2}  - \widehat{w}^{(i)}_{2}\|_2
    & \leq 
    \frac{2L}{N} \sum_{j=0}^{1}
    \exp\big( 
    2  \!\! \sum_{s=j+1}^{1} \eta_s \epsilon_{s} \! + \! \frac{4 \sum_{s=j+1}^{1} \eta_{s} \beta}{N}
    \! + \! \sum_{s=j+1}^{1} \eta^{\frac{1}{\alpha}}_{s}
    \big)\eta_j \\
    &\quad\quad\quad\quad\quad\quad +
    3\rho 
     \big( \frac{1}{ 1 - 2 \eta_{1} ( \epsilon_{1} + \frac{2\beta}{N})  - \eta^{\frac{1}{\alpha}}_{1} } \big)^{1/2} \eta^{1 - \frac{1}{2\alpha}}_{1} \frac{4 \eta_0^2 L^2}{N^2} \\
     & = 
     \frac{2 L}{N}
     \exp\big( 
    2  \eta_1 \epsilon_{1} + \frac{4  \eta_{1} \beta}{N}
    +\eta^{\frac{1}{\alpha}}_{1}
    \big)\eta_0 + 
    \frac{2L\eta_1}{N}
    +3\rho 
     \big( \frac{1}{ 1 - 2 \eta_{1} ( \epsilon_{1} + \frac{2\beta}{N})  - \eta^{\frac{1}{\alpha}}_{1} } \big)^{1/2} \eta^{1 - \frac{1}{2\alpha}}_{1} \frac{4 \eta_0^2 L^2}{N^2}.
\end{align*}
Note from the assumption within the theorem $2 \eta_{1} ( \epsilon_{1} + \frac{2\beta}{N})  - \eta^{\frac{1}{\alpha}}_{1}  \leq 1/2$ the third term can be upper bounded in a similar manner to previously
\begin{align*}
     \big( \frac{1}{ 1 - 2 \eta_{1} ( \epsilon_{1} + \frac{2\beta}{N})  - \eta^{\frac{1}{\alpha}}_{1} } \big)^{1/2} \eta^{1 - \frac{1}{2\alpha}}_{1} \frac{4 \eta_0^2 L^2}{N^2}
     & \leq 
    \exp( 2 \eta_1 ( \epsilon_1 + \frac{2 \beta }{N} ) + \eta_1^{\frac{1}{\alpha}})
     \eta^{1 - \frac{1}{2\alpha}}_{1} \frac{4 \eta_0^2 L^2}{N^2},
\end{align*}
and thus 
\begin{align*}
    \|\widehat{w}_{2}  - \widehat{w}^{(i)}_{2}\|_2
    & \leq
    \frac{2 L}{N}
     \exp\big( 
    2  \eta_1 \epsilon_{1} + \frac{4  \eta_{1} \beta}{N}
    +\eta^{\frac{1}{\alpha}}_{1}
    \big)\eta_0
    \Big( 1 + 
    \underbrace{ 6 \rho \eta^{1 - \frac{1}{2\alpha}}_{1} \frac{ \eta_0 L}{N} }_{\textbf{Remainder Term}}
    \Big) 
    + 
    \frac{2L\eta_1}{N}.
\end{align*}
It is then sufficient to show $\textbf{Remainder Term} \leq 1$ for the base case to hold. Note that this is then implied by the condition on the sample size within the theorem, namely that 
\begin{align*}
    N &  \geq 
     24\rho L \!
    \exp\Big( 
    2  \! \sum_{s=1}^{t} \eta_s\big( \epsilon_{s} \! + \! \frac{4  \beta}{N} \big)
    \! + \! \eta^{\frac{1}{\alpha}}_{s}
    \Big)
    \sum_{j=1}^{t} \eta_{j}^{1-\frac{1}{2\alpha}} \sum_{\ell=0}^{j-1} \eta_\ell \\
    & \geq 
    6 \rho L \eta^{1 - \frac{1}{2\alpha}}_{1}  \eta_0. 
\end{align*}
Let us now assume the inductive hypothesis holds up-to $k$ and consider the case $k+1$. Utilising the inductive hypothesis for $u=1,\dots,k$ as well as multiplying and dividing by $\big( \sum_{j=0}^{u-1} \eta_j \big)^2 $ allows the squared deviation to be bounded
\begin{align*}
    \|\widehat{w}_{u}  - \widehat{w}^{(i)}_{u}\|_2^2 
    & \leq
    \big( \sum_{j=0}^{u-1} \eta_j \big)^2 
    \Big( 
    \frac{4L}{N}\sum_{j=0}^{u-1}
    \exp\big( 
    2  \sum_{s=j+1}^{u-1} \eta_s \epsilon_{s} + \frac{4 \sum_{s=j+1}^{u-1} \eta_{s} \beta}{N}
    +  \sum_{s=j+1}^{u-1} \eta^{\frac{1}{\alpha}}_{s}
    \big)\frac{ \eta_j}{\sum_{j=0}^{u-1}\eta_j }
    \Big)^2
    \\
    & \leq 
    \big(\sum_{j=0}^{u-1} \eta_j \big) 
    \frac{16L^2}{N^2}
    \sum_{j=0}^{u-1}
    \eta_j 
    \exp\big( 
    4  \sum_{s=j+1}^{u-1} \eta_s \epsilon_{s} + \frac{8 \sum_{s=j+1}^{u-1} \eta_{s} \beta}{N}
    +  2 \sum_{s=j+1}^{u-1} \eta^{\frac{1}{\alpha}}_{s}
    \big)\\
    & \leq 
    \big(\sum_{j=0}^{u-1} \eta_j \big) 
    \frac{16L^2}{N^2}
    \exp\big( 
    2  \sum_{s=1}^{u-1} \eta_s \epsilon_{s} + \frac{4 \sum_{s=1}^{u-1} \eta_{s} \beta}{N}
    +   \sum_{s=1}^{u-1} \eta^{\frac{1}{\alpha}}_{s}
    \big) \\
    & \quad\quad 
    \times 
    \Big( 
    \sum_{j=0}^{k}
    \eta_j 
    \exp\big( 
    2 \sum_{s=j+1}^{k} \eta_s \epsilon_{s} + \frac{4 \sum_{s=j+1}^{k} \eta_{s} \beta}{N}
    +  \sum_{s=j+1}^{k} \eta^{\frac{1}{\alpha}}_{s}
    \big)
    \Big),
\end{align*}
where we note the second inequality arises from convexity of the squared function, and the third inequality from adding positive terms within the exponentials. Plugging the above into the squared terms of \eqref{equ:InductiveArgument} for $u=1,\dots,k$, as well as factoring out $\frac{2L}{N} \sum_{j=0}^{k}
    \exp\big( 
     2\sum_{s=j+1}^{k} \eta_s \epsilon_{s} + \frac{4 \sum_{s=j+1}^{k} \eta_{s} \beta}{N}
    +  \sum_{s=j+1}^{k} \eta^{\frac{1}{\alpha}}_{s}
    \big)\eta_j$ allows the deviation at time $k+1$ to be bounded 
\begin{align*}
    & \|\widehat{w}_{k+1}  - \widehat{w}^{(i)}_{k+1}\|_2 
    \leq 
    \Big( \frac{2L}{N} \sum_{j=0}^{k}
    \exp\big( 
    2  \sum_{s=j+1}^{k} \eta_s \epsilon_{s} + \frac{4 \sum_{s=j+1}^{k} \eta_{s} \beta}{N}
    +  \sum_{s=j+1}^{k} \eta^{\frac{1}{\alpha}}_{s}
    \big)\eta_j
    \Big)\\
    & \, \times 
    \Big( \! 1 \! + \! 
\underbrace{ 
\frac{24 \rho L}{N} 
    \sum_{j=1}^{k} \prod_{s=j}^{k}\big( \frac{1}{ 1 - 2 \eta_{s} ( \epsilon_{s} + \frac{2\beta}{N})  - \eta^{\frac{1}{\alpha}}_{s} } \big)^{1/2} \eta^{1 - \frac{1}{2\alpha}}_{j}\big(\sum_{\ell=0}^{j-1} \eta_\ell \big)
    \exp\big( 
    2  \sum_{s=1}^{j-1} \eta_s \epsilon_{s} \! + \! \frac{4 \sum_{s=1}^{j-1} \eta_{s} \beta}{N}
    \! + \!  \sum_{s=1}^{j-1} \eta^{\frac{1}{\alpha}}_{s}
    \big)
    }_{\textbf{Remainder Term}}
    \Big).
\end{align*}
To prove the inductive hypothesis holds for $k+1$ we must show that $\textbf{Remainder Term} \leq 1$. To this end,  follow the previous steps to bound the product of terms for $j=1,\dots,k$ 
\begin{align*}
    \prod_{s=j}^{k}\big( \frac{1}{ 1 - 2 \eta_{s} ( \epsilon_{s} + \frac{2\beta}{N})  - \eta^{\frac{1}{\alpha}}_{s} } \big)^{1/2}
    \leq 
    \exp\big( 
    2  \sum_{s=j}^{k} \eta_s \epsilon_{s} + \frac{4 \sum_{s=j}^{k} \eta_{s} \beta}{N}
    +  \sum_{s=j}^{k} \eta^{\frac{1}{\alpha}}_{s}
    \big)
\end{align*}
to upper bound the \textbf{Remainder Term}
\begin{align*}
    \textbf{Remainder Term}
    \leq 
    \frac{24\rho L}{N} 
    \sum_{j=1}^{k} \eta_{j}^{1-\frac{1}{2\alpha}} \big(\sum_{\ell=0}^{j-1} \eta_\ell \big)
    \exp\big( 
    2  \sum_{s=1}^{k} \eta_s \epsilon_{s} + \frac{4 \sum_{s=1}^{k} \eta_{s} \beta}{N}
    +  \sum_{s=1}^{k} \eta^{\frac{1}{\alpha}}_{s}.
    \big)
\end{align*}
Following the assumption with the theorem, we then have  when $k \leq t$
\begin{align*}
    N & \geq 
    24\rho L 
    \sum_{j=1}^{t} \eta_{j}^{1-\frac{1}{2\alpha}} \big(\sum_{\ell=0}^{j-1} \eta_\ell \big)
    \exp\big( 
    2  \sum_{s=1}^{t} \eta_s \epsilon_{s} + \frac{4 \sum_{s=1}^{t} \eta_{s} \beta}{N}
    +  \sum_{s=1}^{t} \eta^{\frac{1}{\alpha}}_{s}
    \big)\\
    & \geq 
    24\rho L 
    \sum_{j=1}^{k} \eta_{j}^{1-\frac{1}{2\alpha}} \big(\sum_{\ell=0}^{j-1} \eta_\ell \big)
    \exp\big( 
    2  \sum_{s=1}^{k} \eta_s \epsilon_{s} + \frac{4 \sum_{s=1}^{k} \eta_{s} \beta}{N}
    +  \sum_{s=1}^{k} \eta^{\frac{1}{\alpha}}_{s}
    \big)
\end{align*}
and therefore $\textbf{Remainder Term} \leq 1$ as required and the inductive hypothesis holds. This completes the proof.

\subsection{Proof of Lemma \ref{lem:WeakCocoer}}
\label{sec:ProofGen:Lem}
In this section we give the proof of Lemma \ref{lem:WeakCocoer}. We begin recalling a function $f$ is $\beta$-smooth if for any $x,y$ we have
\begin{align*}
    |f(y) - f(x) - \langle \nabla f(x),y-x \rangle| \leq \frac{\beta}{2} \|x - y\|_2^2
\end{align*}
Moreover, to save on notational burden let us denote $x = \widehat{w}_{s}$ and $y = \widehat{w}^{(i)}_{s}$. 

We begin by following the standard proof for showing co-coercivity in the smooth convex setting, see for instance \cite{nesterov2013introductory}. Let us define the functions $\phi_x(\omega) = R(\omega) - \langle R(x),\omega \rangle$ and $\phi_y(\omega) = R(\omega) - \langle R(y),\omega\rangle$. It is then clear that $\phi_x,\phi_y$ are both $\beta$-smooth. As such using smoothness we get 
\begin{align*}
    \phi_x(y - \eta \nabla \phi_x(y)) 
    & \leq \phi_x(y) + \langle \nabla \phi_x(y),y - \eta \nabla \phi(y) - y \rangle + \frac{\beta \eta^2 }{2} \|\nabla \phi_x(y)\|_2^2\\
    & = \phi_x(y)   + \eta \big( \frac{\eta \beta}{2}  - 1 \big) \|\nabla \phi_x(y)\|_2^2
\end{align*}
Plugging in the definition of $\phi_x$, and repeating the steps for with $x,y$ swapped we get the two inequalities 
\begin{align*}
    R(y - \eta \big( \nabla R(y) - \nabla R(x) \big) ) - \langle \nabla R(x),y - \eta \big( \nabla R(y) - \nabla R(x) \big) \rangle 
    & \leq R(y) - \langle R(x),y \rangle \\
    &\quad\quad 
    + \eta \big( \frac{\eta \beta}{2}  - 1 \big) \|\nabla \phi_x(y)\|_2^2\\
    R(x - \eta \big( \nabla R(x) - \nabla R(y) \big) ) - \langle \nabla R(y),x - \eta \big( \nabla R(x) - \nabla R(y) \big) \rangle 
    & \leq R(x) - \langle R(y),x \rangle \\
    &\quad\quad + \eta \big( \frac{\eta \beta}{2}  - 1 \big) \|\nabla \phi_y(x)\|_2^2
\end{align*}
We would now like to lower bound the left side of each of these inequalities. In the convex setting this is immediate from the definition of convexity. In our case, we wish to use the local convexity of the iterates $\widehat{w}_{s}$ on the objective $\nabla R(\cdot)$. We begin with lower bounding the left side of the first inequality.

\paragraph{Lower Bounding First Inequality} Let us denote 
$\widetilde{y} = y - \eta \big( \nabla R(y) - \nabla R(x) \big)$. We then must lower bound 
\begin{align*}
    R(\widetilde{y}) - \langle \nabla R(x), \widetilde{y} \rangle. 
\end{align*}
Recall from Assumption \ref{ass:LocalWeakConvexity} that the Hessian $\nabla^2 R(\cdot)$ has minimum Eigenvalue lower bounded by  $-\epsilon_{s}$ at the point $x = \widehat{\omega}_{s}$. To utilise this let us define the function for $\alpha \in [0,1]$
\begin{align*}
    g(\alpha) = R(x + \alpha ( \widetilde{y} - x)) + \frac{\epsilon_{s}}{2} \|x + \alpha ( \widetilde{y} - x)\|_2^2 + \frac{\alpha^3}{6} \rho \|x - \widetilde{y}\|_2^3.
\end{align*}
Taking derivatives of $g$ with respect to $\alpha$ we observe that
\begin{align*}
    g^{\prime\prime}(\alpha) 
    & = (\widetilde{y} - x)^{\top}\nabla^2 R(x + \alpha (\widetilde{y} - x) ) (\widetilde{y} - x)
    + \epsilon_{s}  \|\widetilde{y} - x\|_2^2 + \alpha \rho \|x - \widetilde{y}\|_2^3\\
    & = 
    (\widetilde{y} - x)^{\top} ( \nabla^2 R(x) + \epsilon_{s} I ) (\widetilde{y} - x) \\
    &\quad\quad\quad\quad 
    + 
    (\widetilde{y} - x)^{\top} ( \nabla^2 R(x + \alpha (\widetilde{y} - x) ) - \nabla^2 R(x) ) (\widetilde{y} - x)
    + \alpha \rho \|x - \widetilde{y}\|_2^3\\
    & \geq 
    0 - \|\widetilde{y} - x \|_2^2 \| \nabla^2 R(x + \alpha (\widetilde{y} - x) ) - \nabla^2 R(x)\|_2
    + \alpha \rho \|x - \widetilde{y}\|_2^3\\
    & \geq 0
\end{align*}
where we have added and subtracted $(\widetilde{y} - x)^{\top} \nabla^2 R(x) (\widetilde{y} - x)$ on the second equality, and note that the first term is lower bounded from the pointwise weak convexity $\nabla^2 R(x) \succeq - \epsilon_s I$.

Therefore $g$ is convex in $\alpha \in [0,1]$, and thus for $\alpha^{\prime} \in [0,1]$  we get $0 \leq g(\alpha) - g(\alpha^{\prime}) - g^{\prime}(\alpha^{\prime})(\alpha - \alpha^{\prime})  $. Picking $\alpha = 1$ and $\alpha^{\prime} =0$ and plugging in the definition of $g$ we have  
\begin{align*}
    0 & \leq R(\widetilde{y}) + \frac{\epsilon_{s}}{2}\|\widetilde{y}\|_2^2 + \frac{\rho}{6} \|x - \widetilde{y}\|_2^3 
    - R(x) - \frac{\epsilon_{s}}{2} \|x\|_2^2 - (\widetilde{y} - x)^{\top} ( \nabla R(x) + \epsilon_{s} x )\\
    & = R(\widetilde{y}) - R(x) - \langle \nabla R(x), \widetilde{y} - x \rangle + 
    \frac{\epsilon_{s}}{2} \|x - \widetilde{y}\|_2^2 + \frac{\rho}{6} \|x - \widetilde{y}\|_2^3.
\end{align*}
Rearranging the above then results in the lower bound 
\begin{align*}
    R(\widetilde{y}) - \langle \nabla R(x),\widetilde{y} \rangle 
    \geq 
    R(x)  - \langle \nabla R(x), x \rangle
    - \frac{\epsilon_{s}}{2} \|x - \widetilde{y}\|_2^2 - \frac{\rho}{6} \|x - \widetilde{y}\|_2^3,
\end{align*}
which will be the lower bound that we will use. 
\\
\paragraph{Lower Bounding Second Inequality} With $\widetilde{x} = x - \eta (\nabla R(x) - \nabla R(y))$ we now lower bound 
\begin{align*}
    R(\widetilde{x}) - \langle \nabla R(y), \widetilde{x} \rangle
\end{align*}
This lower bound will be slightly more technical as the minimum Eigenvalue of $\nabla^2 R(y) = \nabla^2 R(\widehat{w}^{(i)}_{s})$ is not immediately lower bounded from our assumptions. Although, note that we have for any vector  $v \in \mathbb{R}^{p}$ that 
\begin{align*}
    v^{\top} \nabla^2 R(y)v 
    & = 
    v^{\top} \nabla^2 R^{(i)}(y)v + v^{\top} \nabla^2 R^{(i)}(y)  - \nabla^2 R(y) v\\
    & = 
    v^{\top} \nabla^2 R^{(i)}(y)v + 
    \frac{1}{N} v^{\top} \big( \nabla^2 \ell(y,Z_{i}^{\prime}) - \nabla^2 \ell(y,Z_{i}) \big) v\\
    & \geq 
    - \Big( \epsilon_{s}  + \frac{2 \beta}{N} \Big) \|v\|_2^2 
\end{align*}
Therefore $\nabla^2 R(y)$ has minimum Eigenvalue lower bounded by $- \big( \epsilon_{s} + \frac{2\beta}{N}\big)$. Following an identical set of arguments to the lower bound for the first inequality with $\epsilon_s$ swapped with $\epsilon_s + \frac{2 \beta}{N}$ we then get  
\begin{align*}
    R(\widetilde{x}) - \langle \nabla R(y),\widetilde{x}\rangle
    \geq 
    R(y)  - \langle \nabla R(y), y \rangle
    - \frac{\epsilon_{s} + 2\beta /N}{2} \|y - \widetilde{x}\|_2^2 - \frac{\rho}{6} \|\widetilde{x} - y\|_2^3
\end{align*}

\paragraph{Using Lower Bounds} Given the two lower bounds we arrive at after rearranging 
\begin{align*}
    \langle \nabla R(x),x - y \rangle
    & \geq 
    R(x) - R(y)
    +
    \eta \big( 1 - \frac{\eta \beta}{2} \big) \| \nabla \phi_{x}(y)\|_2^2
    - \frac{\epsilon_{s}}{2} \|x - \widetilde{y}\|_2^2 - \frac{\rho}{6} \|x - \widetilde{y}\|_2^3\\
    \langle \nabla R(y), y - x\rangle 
    & \geq R(y) - R(x)
    + \eta \big( 1 - \frac{\eta \beta}{2} \big) \|\nabla \phi_{y}(x) \|_2^2 
    - \frac{\epsilon_{s} + 2\beta /N}{2} \|y - \widetilde{x}\|_2^2 - \frac{\rho}{6} \|\widetilde{x} - y\|_2^3
\end{align*}
The result is then arrived at by adding together the two above bounds and substituting in the definitions of $\phi_x,\phi_y,\widetilde{x},\widetilde{y},x,y$. 

\subsection{Generalisation Error Bound for Gradient Descent under Standard Weak Convexity}
\label{sec:ProofGen:Global}
In this section we present and prove a generalisation error bound of gradient descent under standard weak convexity Assumption \ref{ass:WeakConvexity:Global}. The following theorem presents the generalisation error bound. 
\begin{theorem}[Generalisation Error Bound Standard Weak Convexity]
\label{thm:GenErrorBound:GlobalWeakConvexity}
Consider Assumptions \ref{ass:LossReg} and \ref{ass:WeakConvexity:Global}. If $\eta \beta \leq 3/2$ and $2 \eta \epsilon < 1$, then the generalisation error of gradient descent satisfies
\begin{align*}
     \E[R(\widehat{w}_t) - r(\widehat{w}_t)]
     \leq 
    \frac{2 \eta L^2 }{N} \sum_{k=0}^{t-1} \exp\big( \frac{\eta \epsilon k}{1 - 2 \eta \epsilon} \big)
\end{align*}
\end{theorem}
The proof of Theorem \ref{thm:GenErrorBound:GlobalWeakConvexity} closely follows the steps in proving Theorem \ref{thm:GenErrorBound}. Therefore in the proof we focus on the key differences. We begin with the following lemma which lower bounds the co-coercivity of weakly convex losses. 
\begin{lemma}
\label{lem:WeakCocoer:Global}
Consider Assumptions \ref{ass:LossReg} and \ref{ass:WeakConvexity:Global}. Then for $\eta \geq 0$ and $x,y \in \mathbb{R}^{d}$
\begin{align*}
     \langle \nabla R(x) - \nabla R(y), x - y \rangle
    \geq 2 \eta \Big( 1 - \frac{\beta \eta}{2} \Big) \|\nabla R(x) - \nabla R(y)\|_2^2
    -  \epsilon 
    \|x - y - 
    \eta \big( \nabla R(x) - \nabla R(y)\big) \|_2^2
\end{align*}
\end{lemma}
We now provide the proof of this Lemma. 
\begin{proof}[Lemma \ref{lem:WeakCocoer:Global}]
Follow the proof of Lemma \ref{lem:WeakCocoer} to the point of lower bounding  $R(\widetilde{y}) - \langle \nabla R(x), \widetilde{y} \rangle$ where $\widetilde{y} = y - \eta \big( \nabla R(y) - \nabla R(x) \big)$. Now, let us alternatively choose
\begin{align*}
    g(\alpha) = R(x + \alpha( \widetilde{y} - x) ) + \frac{\epsilon}{2} \|x + \alpha(\widetilde{y} - x) \|_2^2 
\end{align*}
We then immediate see that $g^{\prime\prime}(\alpha) = (\widetilde{y} - x)^{\top} \nabla^2 R(x + \alpha( \widetilde{y} - x) ) (\widetilde{y} - x) + \epsilon \|\widetilde{y} - x \|_2^2 \geq 0 $  since Assumption \ref{ass:WeakConvexity:Global} states that $\nabla^2 R(\omega) \succeq - \epsilon I $ for every $\omega \in \mathbb{R}^{d}$. Therefore $g(\alpha)$ is convex on $\alpha \in [0,1]$. Using that $0 \leq g(1) - g(0) - g^{\prime}(0) $ and rearranging we get the lower bound  
\begin{align*}
    R(\widetilde{y}) - \langle R(x ), \widetilde{y} \rangle \geq R(x) - \langle \nabla R(x),x\rangle - \frac{\epsilon}{2} \|x - \widetilde{y}\|_2^2.
\end{align*}
Performing an identical set of steps for $R(\widetilde{x}) - \langle R(y),\widetilde{x} \rangle $ where $\widetilde{x} = x - \eta\big( \nabla R(x)  - \nabla R(y)\big)$ yields the lower bound 
\begin{align*}
    R(\widetilde{x}) - \langle R(y ), \widetilde{x} \rangle \geq R(y) - \langle \nabla R(y),y\rangle - \frac{\epsilon}{2} \|\widetilde{x} - y\|_2^2.
\end{align*}
Following the steps in Lemma \ref{lem:WeakCocoer} then yields the result. 
\end{proof}
Given this proof, we now provide the proof of Theorem \ref{thm:GenErrorBound:GlobalWeakConvexity}. 
\begin{proof}[Theorem \ref{thm:GenErrorBound:GlobalWeakConvexity}]
Let us begin by bounding the expansiveness of the gradient update. Note for $ 3/(2\beta) \geq \eta \geq 0$ we have when using Lemma \ref{lem:WeakCocoer:Global}
\begin{align*}
    \|x - y - \eta \big( \nabla R(x) - \nabla R(y)\big)\|_2^2 
    & =
    \|x - y\|_2^2 + \eta^2 \|\nabla R(x) - \nabla R(y)\|_2^2 
    - 
    2 \eta \langle x - y, \nabla R(x) - \nabla R(y) \rangle \\
    & \leq 
    \|x - y\|_2^2 
    + \eta^2 \big(1 - 4 \big(1 - \frac{\eta \beta}{2} \big) \big) \|\nabla R(x) - \nabla R(y)\|_2^2  
    \\
    &\quad\quad 
    +
    2 \eta \epsilon 
    \|x - y - 
    \eta \big( \nabla R(x) - \nabla R(y)\big) \|_2^2\\
    & \leq 
    \|x - y\|_2^2 
    +
    2 \eta \epsilon 
    \|x - y - 
    \eta \big( \nabla R(x) - \nabla R(y)\big) \|_2^2
\end{align*}
It is then clear that the expansiveness of the gradient update can be upper bounded
\begin{align*}
    \|x - y - \eta \big( \nabla R(x) - \nabla R(y)\big)\|_2
    \leq 
    \frac{1}{\sqrt{1 - 2 \eta \epsilon }} \|x - y \|_2
\end{align*}
Following the steps in the proof of Theorem \ref{thm:GenErrorBound} (similarly \cite{Hardt:2016:TFG:3045390.3045520}) we immediately get 
\begin{align*}
    \|\widehat{\omega}_{t} - \widehat{\omega}_{t}^{(i)}\|_2
    & \leq 
    \frac{1}{\sqrt{1 - 2 \eta \epsilon}} \|\widehat{\omega}_{t-1} - \widehat{\omega}_{t-1}^{(i)}\|_2
    + 
    \frac{2 \eta L}{N}\\
    & \leq 
    \frac{2 \eta L}{N} \sum_{k=0}^{t-1} \Big( \frac{1}{\sqrt{1 - 2\eta \epsilon}}\Big)^{k}\\
    & \leq 
    \frac{2 \eta L}{N} \sum_{k=0}^{t-1} \exp\big( \frac{\eta \epsilon k}{1 - 2 \eta \epsilon} \big) 
\end{align*}
where we have used that $1/(1-u) = 1 + \frac{u}{1-u} \leq e^{\frac{u}{1-u}}$. This yields the result.

\end{proof}

\section{Proofs of Test Error Bounds for General Loss Functions}
In this section we present proofs related to test error bounds for general loss functions. This section proceeds as follows. Section \ref{sec:testerror:simple} presents the proof of Theorem \ref{thm:TestError:Simple}, which gives a test error bound for weakly convex losses. Section \ref{sec:testerror:general} presents the proof of Proposition \ref{prop:TestError:General} which is presented within Appendix \ref{app:GenWeakConv}. Section \ref{sec:testerror:OptLem} gives the proofs of additional lemmas used within this section.

\subsection{Proof of Test Error Bounds for Weakly Convex Losses (Theorem \ref{thm:TestError:Simple})}
\label{sec:testerror:simple}
In this section we present the proof of Theorem \ref{thm:TestError:Simple}. We begin by introducing the penalised population objective $r_{\epsilon}(\omega):= r(\omega) + \frac{\epsilon}{2} \|\widehat{\omega}_0- \omega\|_2^2$ as well as one of its minimiser $\omega^{\star}_{\epsilon} \in \argmin_{\omega} r_{\epsilon}(\omega)$. The test error is then decomposed as follows 
\begin{align*}
    & \E_{I}[\E[r(\widehat{\omega}_I)]] 
    - \min_{\omega}r(\omega) \\
    & = 
    \E_{I}[\E[r(\widehat{\omega}_I) - R(\widehat{\omega}_I) ]]
    + 
    \E[\E_{I}[R(\widehat{\omega}_I)] - R(\widehat{\omega}^{\star}_{\epsilon})]
    + 
    \E[R(\widehat{\omega}^{\star}_{\epsilon})] - r_{\epsilon}(\omega^{\star}_{\epsilon}) 
    + 
    r_{\epsilon}(\omega^{\star}_{\epsilon}) - \min_{\omega} r(\omega)\\
    & \leq 
    \underbrace{ 
    \max_{k=1,\dots,t} 
    \big\{\E[r(\widehat{\omega}_k) - R(\widehat{\omega}_k) ]\}
    }_{\textbf{Generalisation Error}}
    + 
    \underbrace{ \E[\E_{I}[R(\widehat{\omega}_I)] - R(\widehat{\omega}^{\star}_{\epsilon})]}_{\textbf{Term 1}}
    + 
    \underbrace{ 
    \E[R(\widehat{\omega}^{\star}_{\epsilon})] - r_{\epsilon}(\omega^{\star}_{\epsilon})
    + 
    r_{\epsilon}(\omega^{\star}_{\epsilon}) - \min_{\omega } r(\omega)
    }_{\textbf{Term 2}}
\end{align*}
which has three terms. The \textbf{Generalisation Error} is bounded by Theorem \ref{thm:GenErrorBound:GlobalWeakConvexity} within Appendix \ref{sec:ProofGen:Global}, since we assume standard weak convexity. Note the \textbf{Generalisation Error} term above depends upon the maximum from $k=1,\dots,t$ therefore, apply the bound within Theorem \ref{thm:GenErrorBound} for each instance $k=1,\dots,t$ and take the largest at $k=t$. We now set out to bound \textbf{Term 1} and \textbf{Term 2}, beginning with \textbf{Term 1}. To do so, we introduce the following lemma which is proved in Section \ref{sec:testerror:OptLem}. 
\begin{lemma}
\label{lem:OptError:Simple}
Suppose assumption \ref{ass:LossReg} and \ref{ass:WeakConvexity:Global} hold, $\eta_s = \eta$  for all $s \geq 0$ and  $\eta \beta \leq 1/2$. Then 
\begin{align*}
    \E_{I}[ R(\widehat{\omega}_{I }) ]  
    = \frac{1}{t} \sum_{s=0}^{t-1} 
    R(\widehat{\omega}_{s+1})
    \leq 
    R(\widehat{\omega}^{\star}_{\epsilon}) 
    + 
    \frac{\|\widehat{\omega}_0 - \widehat{\omega}^{\star}_{\epsilon}\|_2^2}{2 \eta t  }
    + \frac{\epsilon}{2} \frac{1}{t} \sum_{s=0}^{t-1} \|\widehat{\omega}_{s+1} - \widehat{\omega}^{\star}_{\epsilon}\|_2^2
\end{align*}
and  $\|\widehat{\omega}_{s+1} - \widehat{\omega}_0\|_2 \leq \sqrt{2 \eta s (R(\widehat{\omega}_0) - R(\widehat{\omega}^{\star}))}$ for $s \geq 0$.
\end{lemma}
Note for $t-1 \geq s \geq 0$ that the deviation $\|\widehat{\omega}_{s+1} - \widehat{\omega}^{\star}_{\epsilon}\|_2^2$ can be bounded by adding and subtracting the initialisation  $\widehat{\omega}_0$ and applying the upper bound on $\|\widehat{\omega}_{s+1} - \widehat{\omega}_0\|_2$ from Lemma \ref{lem:OptError:Simple}. Specifically, using triangle inequality alongside that $(a+b)^2 \leq 2a^2 + 2b^2$ for $a,b \in \mathbb{R}$  gives
\begin{align*}
    \|\widehat{\omega}_{s+1} - \omega^{\star}_{\epsilon}\|_2^2 
    & \leq 2 \|\widehat{\omega}_{s+1} - \widehat{\omega}_{0}\|_2^2 + 2\|\widehat{\omega}_0 - \widehat{\omega}^{\star}_{\epsilon}\|_2^2\\
    & \leq 2 \big( \eta t (R(\widehat{\omega}_0) - R(\widehat{\omega}^{\star})) 
    +2 \|\widehat{\omega}_0 - \widehat{\omega}^{\star}_{\epsilon}\|_2^2.
\end{align*}
Combining this with the first part of Lemma \ref{lem:OptError:Simple} and taking expectation then gives the following upper bound for \textbf{Term 1}
\begin{align*}
    \textbf{Term 1}
    \leq 
    \frac{\E[\|\omega_0 - \widehat{\omega}^{\star}_{\epsilon}\|_2^2]}{2  \eta t  }
    + 
    \epsilon \Big( \eta t \E[R(\widehat{\omega}_0) - R(\widehat{\omega}^{\star})] + 
    \E[\|\widehat{\omega}_0 - \widehat{\omega}^{\star}_{\epsilon}\|_2^2]\Big).
\end{align*}
To bound \textbf{Term 2} begin by noting that we have the lower bound 
\begin{align*}
    r_{\epsilon}(\omega^{\star}_{\epsilon}) 
    & = 
    \E[\ell(\omega^{\star}_{\epsilon},Z) + \epsilon \|\widehat{\omega}_{0} - \omega^{\star}_{\epsilon}\|_2^2 ] \\
    & = 
    \E\Big[ \frac{1}{N} \sum_{i=1}^{N}
    \ell(\omega^{\star}_{\epsilon},Z_i) + \epsilon \|\widehat{\omega}_{0} - \omega^{\star}_{\epsilon}\|_2^2\Big]\\
    &  = \E[ R(\omega^{\star}_{\epsilon}) + \epsilon \|\widehat{\omega}_{0} - \omega^{\star}_{\epsilon}\|_2^2]\\
    & \geq 
    \E[ R(\widehat{\omega}^{\star}_{\epsilon}) + \epsilon \|\widehat{\omega}_{0} - \widehat{\omega}^{\star}_{\epsilon}\|_2^2],
\end{align*}
where we recall that $\widehat{\omega}^{\star}_{\epsilon} = \argmin_{\omega}\big( R(\omega) + \epsilon \|\widehat{\omega}_{0} - \omega \|_2^2 \big)$. Moreover, note that we also have $r_{\epsilon}(\omega^{\star}_{\epsilon}) \leq r_{\epsilon}(\omega^{\star})$ where $\omega^{\star} \in \argmin_{\omega} r(\omega)$. By adding and subtracting $\epsilon\E[ \|\widehat{\omega}_0 - \widehat{\omega}^{\star}_{\epsilon}\|_2^2 ] $ and using these two facts then yields the following upper bound on \textbf{Term 2}
\begin{align*}
    \textbf{Term 2}
    & =  
    \underbrace{ \E[R(\widehat{\omega}^{\star}_{\epsilon}) + \epsilon \|\widehat{\omega}_0 - \widehat{\omega}^{\star}_{\epsilon}\|_2^2 ] - r_{\epsilon}(\omega^{\star}_{\epsilon})}_{\leq 0 }
    + 
    r_{\epsilon}(\omega^{\star}_{\epsilon}) - r(\omega^{\star})
    - 
    \epsilon \E[ \|\widehat{\omega}_{0} - \widehat{\omega}^{\star}_{\epsilon}\|_2^2 ] \\
    & \leq 
    r_{\epsilon}(\omega^{\star}_{\epsilon}) - r(\omega^{\star})
    - 
    \epsilon \E[ \|\widehat{\omega}_0 - \widehat{\omega}^{\star}_{\epsilon}\|_2^2 ] \\
    & \leq 
    \epsilon \|\widehat{\omega}_0 - \omega^{\star}\|_2^2 
    - 
    \epsilon \E[ \|\widehat{\omega}_0 - \widehat{\omega}^{\star}_{\epsilon}\|_2^2 ].
\end{align*}
Bringing together the bounds for the \textbf{Generalisation Error}, \textbf{Term 1} and \textbf{Term 2} and noting that $\epsilon \E[ \|\widehat{\omega}_0 - \widehat{\omega}^{\star}_{\epsilon}\|_2^2 ]$ in \textbf{Term 1} and \textbf{Term 2} cancel, yields the result.  

\subsection{Proof of Optimisation and Approximation Error Bounds under Generalised Weak Convexity (Proposition \ref{prop:TestError:General})}
\label{sec:testerror:general}
In this section we present the proof of Proposition \ref{prop:TestError:General}. Following the proof of Theorem \ref{thm:TestError:Simple} in the previous section, we begin with the following decomposition of the \textbf{Optimisation \& Approximation Error}. Specifically, for $\widetilde{\omega} \in \mathcal{X}$
\begin{align*}
     \underbrace{ \E_{I}[ E[ R(\widehat{\omega}_{I}]]  - r(\omega^{\star})}_{\textbf{Opt. \& Approx. Error}}
    =
    \underbrace{ 
    \E[\E_{I}[R(\widehat{\omega}_I)] - R(\widetilde{\omega})]}_{\textbf{Term 1}}
    + 
    \underbrace{ \E[R(\widetilde{\omega})] - r(\omega^{\star})}_{\textbf{Term 2}}.
\end{align*}
Where we have labelled the \textbf{Term 1} and \textbf{Term 2}. We now proceed to bound \textbf{Term 1}. To do so we will use the following lemma, which is a generalisation of Lemma \ref{lem:OptError:Simple} given previously. 
\begin{lemma}
\label{lem:OptError:General}
Suppose assumption \ref{ass:LossReg} and \ref{ass:WeakConvexity:Local} hold, $\eta_s = \eta$ for $s \geq 0$ and  $\eta \beta \leq 1/2$. Furthermore suppose that $\widehat{\omega}_{s} \in \mathcal{X}$ for $s \geq 0$. Then for $\widetilde{\omega} \in \mathcal{X}$ we have 
\begin{align*}
    \E_{I}[ R(\widehat{\omega}_{I }) ]  
    = \frac{1}{t} \sum_{s=0}^{t-1} 
    R(\widehat{\omega}_{s+1})
    \leq 
    R(\widetilde{\omega}) 
    + 
    \frac{\|\widehat{\omega}_0 - \widetilde{\omega}\|_2^2}{2 \eta t }
    + \frac{1}{2}\frac{1}{t} \sum_{s=0}^{t-1} G_{\mathcal{X}}(\widetilde{\omega}-\widehat{\omega}_{s+1}) 
\end{align*}
and $\|\widehat{\omega}_{s+1} - \widehat{\omega}_0\|_2 \leq \sqrt{2 \eta s (R(\widehat{\omega}_0) - R(\widehat{\omega}^{\star}))}$ for $s \geq 0$.
\end{lemma}
Plugging the bound from the first part of Lemma \ref{lem:OptError:General} and taking expectation gives the result. 

\subsection{Proof of Lemma \ref{lem:OptError:Simple} and \ref{lem:OptError:General} }
\label{sec:testerror:OptLem}
In this section we provide the proof of both Lemma \ref{lem:OptError:Simple} and  \ref{lem:OptError:General}. We note that Lemma \ref{lem:OptError:General} holds under a weaker assumption than Lemma \ref{lem:OptError:Simple}. We therefore begin with the proof of Lemma \ref{lem:OptError:General}, with the proof of Lemma \ref{lem:OptError:Simple} given after.

\begin{proof}[Lemma  \ref{lem:OptError:General}]
Recall that $R(\cdot)$ is $\beta$-smooth, and therefore, using smoothness and the first iteration of gradient descent for $s \geq 0$ i.e. $\widehat{\omega}_{s+1} = \widehat{\omega}_{s} - \eta \nabla R(\widehat{\omega}_{s})$ we get
\begin{align}
\label{equ:OptGeneral:Smooth}
    R(\widehat{\omega}_{s+1}) 
    & \leq R(\widehat{\omega}_{s}) +
    \langle \nabla R(\widehat{\omega}_{s}), \widehat{\omega}_{s+1} - \widehat{\omega}_{s} \rangle 
    + \frac{\beta}{2} \|\widehat{\omega}_{s+1} - \widehat{\omega}_{s}\|_2^2 
    \nonumber \\
    & = 
    R(\widehat{\omega}_{s}) 
    - \eta \|\nabla R(\widehat{\omega}_{s})\|_2^2
    + 
    \frac{\eta^2 \beta }{2} \|\nabla R(\widehat{\omega}_{s})\|_2^2 
    \nonumber 
    \\
    & \leq 
    R(\widehat{\omega}_s) - \frac{\eta}{2} \|\nabla R(\widehat{\omega}_s)\|_2^2
\end{align}
where at the end we used that $\eta \beta \leq 1$. We now wish to upper bound $R(\widehat{\omega}_s)$, for which we will use our Assumption \ref{ass:WeakConvexity:Local}. In particular, let us define the function for $\alpha \in [0,1]$ as 
\begin{align*}
    g(\alpha) = R(\widehat{\omega}_s + \alpha( \widetilde{\omega} - \widehat{\omega}_s) ) 
    +  \frac{\alpha^2}{2} G_{\mathcal{X}}(\widetilde{\omega} - \widehat{\omega}_s).
\end{align*}
Differentiating with respect to $\alpha$ twice we get 
\begin{align*}
    g^{\prime}(\alpha) & = (\widetilde{\omega} - \widehat{\omega}_s)^{\top}
    \nabla R(\widehat{\omega}_s + \alpha( \widetilde{\omega} - \widehat{\omega}_s) ) 
    + \alpha G_{\mathcal{X}}(\widetilde{\omega} - \widehat{\omega}_s) \\
    g^{\prime\prime}(\alpha)
    & = (\widetilde{\omega} - \widehat{\omega}_t)^{\top}
    \nabla^2 R(\widehat{\omega}_s + \alpha( \widetilde{\omega} - \widehat{\omega}_s) )
    (\widetilde{\omega} - \widehat{\omega}_s)
    +  G_{\mathcal{X}}(\widetilde{\omega} - \widehat{\omega}_s).
\end{align*}
Observe that $g^{\prime\prime}(\alpha) \geq 0$ for $\alpha \in [0,1]$ from Assumption \ref{ass:WeakConvexity:Local}. That is $\widehat{\omega}_s + \alpha(\widetilde{\omega} - \widehat{\omega}_s) \in \mathcal{X}$, since  both $\widehat{\omega}_{s},\widetilde{\omega} \in \mathcal{X} $ and therefore so is  the linear combination $(1-\alpha)\widehat{\omega}_s + \alpha \widetilde{\omega}$ provided $\alpha \in [0,1]$ by the convexity of the set $\mathcal{X}$. Since $g(\alpha)$ is convex on $\alpha \in [0,1]$ we then have the inequality $g(0) \leq g(1)  - g^{\prime}(0)$. Plugging in the definition of $g(\cdot)$ into this then gives
\begin{align*}
    R(\widehat{\omega}_s) 
    \leq 
    R(\widetilde{\omega}) + \frac{1}{2} G_{\mathcal{X}}(\widetilde{\omega} - \widehat{\omega}_s) 
    -  
    (\widetilde{\omega} - \widehat{\omega}_s)^{\top}
    \nabla R(\widehat{\omega}_s ).
\end{align*}
Using this upper bound with \eqref{equ:OptGeneral:Smooth} and following the standard steps for proving the convergence of gradient descent yields
\begin{align*}
    R(\widehat{\omega}_{s+1}) 
    & \leq 
    R(\widetilde{\omega}) 
    + \nabla R(\widehat{\omega}_s)^{\top}( \widehat{\omega}_s - \widetilde{\omega})
    - 
    \frac{\eta}{2} \|\nabla R(\widehat{\omega}_{s})\|_2^2 
    + \frac{1}{2} G_{\mathcal{X}}(\widetilde{\omega} - \widehat{\omega}_s) \\
    & = 
    R(\widetilde{\omega}) 
    +
    \frac{1}{\eta} \big( \widehat{\omega}_s - \widehat{\omega}_{s+1} \big)^{\top}
    ( \widehat{\omega}_s - \widetilde{\omega})
    - \frac{1}{2 \eta}\|\widehat{\omega}_{s+1} - \widehat{\omega}_s \|_2^2
    + \frac{1}{2} G_{\mathcal{X}}(\widetilde{\omega} - \widehat{\omega}_s) \\
    & = 
    R(\widetilde{\omega}) 
    + \frac{1}{2\eta}\big( \|\widehat{\omega}_s - \widetilde{\omega}\|_2^2 - \|\widehat{\omega}_{s+1} - \widetilde{\omega}\|_2^2
    \big)
    + \frac{1}{2} G_{\mathcal{X}}(\widetilde{\omega} - \widehat{\omega}_s).
\end{align*}
Summing up this bound for $s=0,\dots,t-1$ and diving by $t$ gives 
\begin{align*}
    \E_{I}[ R(\omega_{I }) ]  
    = \frac{1}{t} \sum_{s=0}^{t-1} 
    R(\widehat{\omega}_{s+1})
    \leq 
    R(\widetilde{\omega}) 
    + 
    \frac{\|\widehat{\omega}_0 - \widetilde{\omega}\|_2^2}{2 \eta t }
    + \frac{1}{2} \frac{1}{t} \sum_{s=0}^{t-1} G_{\mathcal{X}}(\widetilde{\omega}- \widehat{\omega}_{s+1}),
\end{align*}
which proves the first part of the lemma. To show the second part of the lemma, upper bounding $\|\widehat{\omega}_s - \widehat{\omega}_0\|_2$, we look back to \eqref{equ:OptGeneral:Smooth}. Recalling that $\eta \nabla R(\widehat{\omega}_s) = \widehat{\omega}_s - \widehat{\omega}_{s+1}$, and rearranging we get the upper bound
\begin{align*}
    \| \widehat{\omega}_{s+1} - \widehat{\omega}_s\|_2^2 
    \leq 
    2 \eta \big( R(\widehat{\omega}_s) - R(\widehat{\omega}_{s+1})\big).
\end{align*}
Summing up both sides yields 
\begin{align*}
    \sum_{s=0}^{t-1}
    \| \widehat{\omega}_{s+1} - \widehat{\omega}_s\|_2^2 
    \leq 
    \eta 
    \big( R(\widehat{\omega}_0) - R(\widehat{\omega}_{t})\big) 
    \leq 
    \eta 
    \big( R(\widehat{\omega}_0) - R(\widehat{\omega}^{\star})\big),
\end{align*}
where we have plugged in a minimiser $\widehat{\omega}^{\star} \in \argmin_{\omega} R(\omega)$. 
Note that convexity of the squared norm $\|\cdot\|_2^2$ gives us the lower bound 
$\frac{1}{t} \sum_{s=0}^{t} \| \widehat{\omega}_{s+1} - \widehat{\omega}_s\|_2^2 \geq 
\|\frac{1}{t} \sum_{s=0}^{t} \widehat{\omega}_{s+1} -\widehat{\omega}_s \|_2^2 
= \frac{1}{t^2}\|\widehat{\omega}_{t+1} - \widehat{\omega}_0 \|_2^2 $. Using this to lower bound the right hand side of the above when multiplying then proves the second part of the lemma. 
\end{proof}
We now proceed to the present the proof of Lemma \ref{lem:OptError:Simple}, which come as an application of Lemma \ref{lem:OptError:General}.  
\begin{proof}[Lemma \ref{lem:OptError:Simple}]
Note Assumption \ref{ass:WeakConvexity:Global} implies Assumption \ref{ass:WeakConvexity:Local} with $\mathcal{X} = \mathbb{R}^{p}$ and $G_{\mathcal{X}}(u) = \epsilon \|u\|_2^2$. Plugging in the bound from Lemma \ref{lem:OptError:General} immediately yields the result. 
\end{proof}

\section{Proof of Results for Two and Three Layer Neural Networks}
In this section we present proofs of the result within Section \ref{sec:TwoLayerNN} as well as Theorem \ref{thm:WeakConvexityError:TwoLayer} from Appendix \ref{app:GenWeakConv}. Recall that we consider the supervised learning setting described within Section \ref{sec:TwoLayerNN:Setup} where the loss function is a composition of a function $g:\mathbb{R}\times\mathbb{R} \rightarrow \mathbb{R}$, which is convex and smooth in its first argument $g(\cdot,y):\mathbb{R}\rightarrow \mathbb{R}$ for any $y \in \mathbb{R}$, as well as a prediction function parameterised by $\omega \in \mathbb{R}^{p}$ such that $f(\cdot,\omega):\mathbb{R}^{d} \rightarrow \mathbb{R}$. For an observation $Z =(x,y)$ the loss function is then $\ell(\omega,Z) = g(f(x,\omega),y)$. 

Recall the minimum Eigenvalue of a matrix can be defined as the minimum quadratic form with vectors on a unit ball i.e. for $A \in \mathbb{R}^{p \times p}$ the quantity 
$\min_{ u \in \mathbb{R}^{p} \, \|u\|_2 = 1 } u^{\top}A u $ (see for instance Section 6.1.1 in \cite{wainwright2019high}). Therefore, we are focused on lower bounding quadratic forms of the empirical risk Hessian i.e. $u^{\top}\nabla^2 R(\omega)u$ with $u \in \mathbb{R}^{p}$. It will therefore be convenient to write out the gradient and Hessian of the empirical risk in this case 
\begin{align*}
    \nabla R(\omega) & = \frac{1}{N}\sum_{i=1}^{N}g^{\prime}(f(x_i,\omega),y_i) \nabla f(x_i,\omega) \\
    \nabla^2 R(\omega) & = \frac{1}{N}\sum_{i=1}^{N}g^{\prime\prime}(f(x_i,\omega),y_i) \nabla f(x_i,\omega) f(x_i,\omega)^{\top} + 
    \frac{1}{N} \sum_{i=1}^{N}g^{\prime}(f(x_i,\omega),y_i) \nabla^2 f(x_i,\omega).
\end{align*}
Moreover, note for $u \in \mathbb{R}^{p}$ the quadratic form of the Hessian then  decomposes as 
\begin{align}
\label{equ:HessianLowerBound}
    u^{\top} \nabla^2 R(\omega) u 
    = 
    \underbrace{ 
    \frac{1}{N} \sum_{i=1}^{N}
    g^{\prime\prime}(f(x_i,\omega),y_i) \langle u, \nabla f(x_i,\omega)\rangle^2 
    }_{\geq 0 }
    + 
    \frac{1}{N} \sum_{i=1}^{N}
    g^{\prime}(f(x_i,\omega),y_i) u^{\top} \nabla^2 f(x_i,\omega)u.
\end{align}
Since the first quantity above is non-negative, to lower bound the quadratic form $u^{\top} \nabla^2 R(\omega) u$ it is sufficient to lower bound the second term only. 

The remainder of this section is then as follows. Section \ref{sec:WeakConvexityNN:Thm2} presents the proof of Theorem \ref{thm:WeakConvexity:SingleLayerNN:Both} which considers the minimum Eigenvalue of the empirical risk in the case of a two layer neural network. Section \ref{sec:WeakConvexityNN:Thm3} presents the proof of Theorem \ref{thm:WeakConvexity:TwoLayerNN:Both} which considers a three layer neural network when both the first and third layers are optimised.

\subsection{Bounding Empirical Risk Hessian Minimum Eigenvalue for a Two Layer Neural Network (Theorem \ref{thm:WeakConvexity:SingleLayerNN:Both})}
\label{sec:WeakConvexityNN:Thm2}
Recall Section \ref{sec:TwoLayerNN}, where the prediction function is a two layer neural network. This is defined for width $M \geq 1$ and scaling $ 1/2 \leq c \leq 1$ as $f(x,\omega) := \frac{1}{M^c} \sum_{j=1}^{M} v_j \sigma(\langle A_j,x\rangle)$. The parameter is then the concatenation $\omega = (A,v) \in \mathbb{R}^{dM + M} $ where the first layer $A \in \mathbb{R}^{M \times d}$ has been vectorised in a row-major manner. For a parameter $\omega$ and input $x \in \mathbb{R}^{d}$ the gradient and Hessian of the prediction function are, respectively, a vector $\nabla f(x,\omega) \in \mathbb{R}^{Md + M}$ and matrix $\nabla^2 f(x,\omega) \in \mathbb{R}^{(Md+M) \times (Md+M)}$. We now go on to calculate these quantities, noting for $k=1,\dots,d$ that the $k$th co-ordinate of the input $x$ will be denoted $(x)_k$.

Let us decompose co-ordinates of the gradient $(\nabla f(x))_{i}$ for $i=1,\dots,Md +M$ into two parts associated to the first and second layer. Specifically, for the first layer consider $i = (j-1)d + k$ with $j=1,\dots,M$ and $k=1,\dots,d$. With this indexing, the $i$th co-ordinate aligns with the gradient of the $j$th neuron  and $k$th input, which is 
\begin{align*}
    (\nabla f(x,\omega) )_{i} = \frac{\partial f(x,\omega)}{\partial A_{jk}} = 
    \frac{1}{M^c} v_j \sigma^{\prime}(\langle A_j , x \rangle ) (x)_k.
\end{align*}
Meanwhile, for the second layer consider $i=Md + k$ with $k=1,\dots,M$. The $i$th co-ordinate then aligns with the gradient of the $k$th weight in the second layer, which is 
\begin{align*}
    (\nabla f(x,\omega))_i = \frac{\partial f(x,\omega)}{\partial v_k} = 
    \frac{1}{M^c} \sigma(\langle A_k,x\rangle).
\end{align*}

Let us now decompose the entries of the Hessian $(\nabla^2 f(x,\omega))_{i\ell}$ for $i,\ell = 1,\dots,Md + M$ into three parts associated to the second derivative of the first layer only, second derivative of the second layer only, and the derivative with respect to both the first and second layers. Specifically, for the second derivative with respect to the first layer consider $\ell = (\widetilde{j} - 1)d + \widetilde{k}$ and $i=(j-1)d + k$ with $j,\widetilde{j} = 1,\dots,M$ and $k,\widetilde{k} = 1,\dots,d$. For these indices the Hessian is  
\begin{align*}
    (\nabla^2 f(x,\omega))_{i\ell} = 
    \frac{\partial^2 f(x,\omega)}{\partial A_{jk} \partial A_{\widetilde{j}\widetilde{k}}} 
    = \begin{cases}
    \frac{1}{M^c} v_j \sigma^{\prime\prime}(\langle A_j , x \rangle ) (x)_k (x)_{\widetilde{k}} & \text{ if } j=\widetilde{j} \\ 
    0 & \text{otherwise}
    \end{cases}.
\end{align*}
For the second derivative with respect to the second layer, consider the indices $i=Md + j$ and $\ell = Md + \widetilde{j}$ with $j,\widetilde{j}=1,\dots,M$. For these indices Hessian is  
\begin{align*}
    (\nabla^2 f(x,\omega))_{i \ell} 
    = 
    \frac{f(x,\omega)}{\partial v_j \partial v_{\widetilde{j}}} = 
    0.
\end{align*}
Finally, for the derivative with respect to the first and second layers consider the indicies $i=(j-1)d + k$ and $\ell = Md + \widetilde{j}$ with $j,\widetilde{j}=1,\dots,M$ and $k=1,\dots,d$. For these indices the Hessian is
\begin{align*}
    (\nabla^2 f(x,\omega))_{i \ell} = 
    \frac{\partial^2 f(x,\omega)}{\partial A_{jk} \partial v_{\widetilde{j}}}
    = 
    \begin{cases}
    \frac{1}{M^c} \sigma^{\prime}(\langle A_j,x \rangle) (x)_k & \text{ if } j=\widetilde{j} \\
    0 & \text{Otherwise}
    \end{cases}.
\end{align*}
Now let the vector $u \in \mathbb{R}^{Md + M}$ have unit norm $\|u\|_2 = 1$ and be composed in a manner matching the parameter $\omega = (A,v)$ so that $u = (A(u),v(u))$ where $A(u) \in \mathbb{R}^{M \times d}$ has been vectorised in a row-major manner and $u(v) \in \mathbb{R}^{M}$. Following the previous definitions, the quadratic form of the prediction function Hessian is 
\begin{align*}
    u^{\top} \nabla^2 f(x,\omega) u 
     & = 
     \sum_{\widetilde{j},j=1}^{M} \sum_{k,\widetilde{k} =1}^{d} A(u)_{jk} A(u)_{\widetilde{j}\widetilde{k}} 
     \frac{\partial^2 f(x,\omega)}{\partial A_{jk} \partial A_{\widetilde{j}\widetilde{k}}} 
     + 
     2  \sum_{j,\widetilde{j}=1}^{M} \sum_{k=1}^{d} A(u)_{jk} v(u)_{\widetilde{j}} \frac{\partial^2 f(x,\omega)}{\partial A_{jk} \partial v_{\widetilde{j}}} \\
    & \quad\quad 
    +  \sum_{j,\widetilde{j} =1}^{M}
    v(u)_j v(u)_{\widetilde{j}}
    \frac{\partial^2 f(x,\omega)}{\partial v_{j} \partial v_{\widetilde{j}}}\\
    & = 
    \frac{1}{M^c} \sum_{j=1}^{M} v_j \sigma^{\prime\prime}(\langle A_j , x \rangle ) \langle x,A(u)_j\rangle^2 + 
     2  \frac{1}{M^c} \sum_{j=1}^{M} v(u)_{j} \sigma^{\prime}(\langle A_j,x\rangle) \langle x,A(u)_j\rangle.
\end{align*}
Let us now lower bound the quadratic form of the empirical risk's Hessian $u^\top \nabla^2 R(\omega) u$, and thus, the minimum Eigenvalue. Recall the discussion around equation \eqref{equ:HessianLowerBound}, in that the first term in the decomposition of the quadratic form $u^\top \nabla^2 R(\omega) u$ is non-negative. Utilising this and plugging in the above gives the following lower bound
\begin{align}
    u^{\top}\nabla^2 R(\omega) u 
    & \geq 
    \frac{1}{N} \sum_{i=1}^{N}
    g^{\prime}(f(x_i,\omega),y_i) u^{\top} \nabla^2 f(x_i,\omega)u
    \nonumber
    \\
    & = 
    \frac{1}{M^c} \frac{1}{N} 
    \sum_{i=1}^{N} g^{\prime}(f(x_i,\omega),y_i) 
    \sum_{j=1}^{M} v_j \sigma^{\prime\prime}(\langle A_j , x_i \rangle ) \langle x_i,A(u)_j\rangle^2
    \nonumber \\
    &\quad\quad 
    + 
     2  \frac{1}{M^c} \frac{1}{N} \sum_{i=1}^{N} g^{\prime}(f(x_i,\omega),y_i)  
     \sum_{j=1}^{M} v(u)_{j} \sigma^{\prime}(\langle A_j,x_i \rangle) \langle x_i,A(u)_j\rangle.
     \label{equ:ThmCheckpoint:1}
\end{align}
We now set to lower bound each of the terms in \eqref{equ:ThmCheckpoint:1}. For the first term note we have, using upper bounds on the derivative of $g$ as well as the activation $\sigma$, the lower bound 
\begin{align*}
    & \frac{1}{M^c} \frac{1}{N} \sum_{i=1}^{N} g^{\prime}(f(x_i,\omega),y_i) 
    \sum_{j=1}^{M} v_j \sigma^{\prime\prime}(\langle A_j , x_i \rangle ) \langle x_i,A(u)_j\rangle^2 \\
    & \geq 
    - \frac{1}{M^c} \frac{1}{N} \sum_{i=1}^{N} \sum_{j=1}^{M} 
    | g^{\prime}(f(x_i,A),y_i) 
     v_j \sigma^{\prime\prime}(\langle A_j , x_i \rangle )|  \langle x_i,A(u)_j\rangle^2 \\
     & \geq 
     - \frac{L_{g^{\prime}} L_{\sigma^{\prime\prime}} \|v\|_{\infty} }{M^c} \frac{1}{N} \sum_{i=1}^{N} \sum_{j=1}^{M} 
     \langle x_i,A(u)_j\rangle^2 \\
     & =  
     - \frac{L_{g^{\prime}} L_{\sigma^{\prime\prime}} \|v\|_{\infty} }{M^c}  \sum_{j=1}^{M} 
     A(u)_j^{\top} \Big( \frac{1}{N} \sum_{i=1}^{N}x_i x_i^{\top} \Big) A(u)_j \\
     & \geq 
     - \frac{L_{g^{\prime}} L_{\sigma^{\prime\prime}} \|v\|_{\infty} }{M^c}
     \|\widehat{\Sigma}\|_2 
     \sum_{j=1}^{M}
     \|A(u)_j\|_2^2
\end{align*}
Let us now consider the second term in \eqref{equ:ThmCheckpoint:1}. Bringing out the summation over $j=1\dots,M$ and using Cauchy-Schwarz we get 
\begin{align*}
    & \frac{1}{N} \sum_{i=1}^{N} g^{\prime}(f(x_i,\omega),y_i)  
     \sum_{j=1}^{M} v(u)_{j} \sigma^{\prime}(\langle A_j,x_i \rangle) \langle x_i,A(u)_j\rangle\\
     & = 
      \sum_{j=1}^{M} v(u)_j \frac{1}{N} \sum_{i=1}^{N}g^{\prime}(f(x_i,\omega),y_i) \sigma^{\prime}(\langle A_j,x_i \rangle) \langle x_i,A(u)_j\rangle \\
      & \geq 
      - \Big| 
      \sum_{j=1}^{M} v(u)_j \frac{1}{N} \sum_{i=1}^{N}g^{\prime}(f(x_i,\omega),y_i) \sigma^{\prime}(\langle A_j,x_i \rangle) \langle x_i,A(u)_j\rangle \Big|\\
     & \geq
      - \sqrt{ \sum_{j=1}^{M }v(u)_j^2}
     \sqrt{ \sum_{j=1}^{M} \Big( \frac{1}{N} \sum_{i=1}^{N} g^{\prime}(f(x_i,\omega),y_i)   \sigma^{\prime}(\langle A_j,x_i \rangle) \langle x_i,A(u)_j\rangle 
     \Big)^2 } \\
     & \geq 
      - \sqrt{ \sum_{j=1}^{M }v(u)_j^2} 
     \sqrt{ \sum_{j=1}^{M} \frac{1}{N}
     \sum_{i=1}^{N} \big( g^{\prime}(f(x_i,\omega),y_i)  \sigma^{\prime}(\langle A_j,x_i \rangle) \big)^2 \langle x_i ,A(u)_j\rangle^2}
      \\
      & \geq 
       - L_{\sigma^{\prime}} L_{g^{\prime}} 
       \sqrt{ \sum_{j=1}^{M }v(u)_j^2} 
       \sqrt{\sum_{j=1}^{M} \frac{1}{N}\sum_{i=1}^{N} \langle x_i ,A(u)_j\rangle^2} \\
       & \geq 
       - L_{\sigma^{\prime}} L_{g^{\prime}} 
       \sqrt{ \sum_{j=1}^{M }v(u)_j^2} \sqrt{\|\widehat{\Sigma}\|_2}
       \sqrt{\sum_{j=1}^{M} \|A(u)_j\|_2^2}
\end{align*}
where we note some key steps within the above calculation. The third inequality arises from convexity of the squared function. Meanwhile the fourth inequality follows from the steps to bound the first term. Plugging each of these bounds in \eqref{equ:ThmCheckpoint:1} then immediately yields the lower bound
\begin{align*}
    u^{\top} R(\omega) u
    & \geq - 
    \frac{L_{g^{\prime}} L_{\sigma^{\prime\prime}} \|\widehat{\Sigma}\|_2 \|v\|_{\infty} }{M^c}
      \sum_{j=1}^{M} \|A(u)_j\|_2^2 
    - 2 \frac{L_{\sigma^{\prime}} L_{g^{\prime}} \sqrt{\|\widehat{\Sigma}\|_2}}{M^c} \|v(u)\|_2 \sqrt{ \sum_{j=1}^{M} \|A(u)_j\|_{2}^2}\\
    & \geq 
    - \frac{L_{g^{\prime}} L_{\sigma^{\prime\prime}}\|\widehat{\Sigma}\|_2  \|v\|_{\infty} }{M^c} 
    - 
    2 \frac{L_{\sigma^{\prime}} L_{g^{\prime}} \sqrt{\|\widehat{\Sigma}\|_2}}{M^c},
\end{align*}
where at the end we used that $\|u\|_2^2 =\|v(u)\|_2^2 +  \sum_{j=1}^{M} \|A(u)_j\|_2^2  = 1$. This is then a lower bound on the minimum Eigenvalue, as required.

\subsection{Bounding Empirical Risk Hessian Minimum Eigenvalue for a Three Layer Neural Network }
\label{sec:WeakConvexityNN:Thm3}
In this section we provide a lower bound on the minimum Eigenvalue of the empirical risk Hessian when the prediction function is a three layer neural network. Specifically, for $M_1,M_2 \geq 1$ and $1/2 \leq c \leq 1$ the prediction function takes the form
\begin{align}
\label{equ:ThreeLayerNN}
    f(x,\omega) = \frac{1}{M_2^c} \sum_{i=1}^{M_2}v_i 
    \sigma\Big( \frac{1}{M_1^c} \sum_{s=1}^{M_1}A^{(2)}_{is} \langle A^{(1)}_{s},x\rangle \Big),
\end{align}
where we have denoted the first layer of weights $A^{(1)} \in \mathbb{R}^{M_1 \times d}$, the second layer of weights $A^{(2)} \in \mathbb{R}^{M_2 \times M_1}$ and third layer of weights $v \in \mathbb{R}^{M_2}$. The first activation is linear, while the second activation is more generally $\sigma : \mathbb{R} \rightarrow \mathbb{R}$. We consider the setting in which we only optimise the first and third set of weights, and thus, the parameter will be the concatenation $\omega = (A^{(1)},v) \in \mathbb{R}^{M_1 d + M_2}$ where the first layer of weights $A^{(1)}$ have been vectorised in a row-major manner. It will also be convenient to denote the empirical covariance of the second layer $\widehat{\Sigma}_{A^{(2)}}:= \frac{1}{M_2} (A^{(2)})^{\top} A^{(2)}$. The following theorem then presents the lower bound on the minimum Eigenvalue in this case. 
\begin{theorem} 
\label{thm:WeakConvexity:TwoLayerNN:Both}
Consider the loss function as in Section \ref{sec:TwoLayerNN:Setup} with three layer neural network prediction function \eqref{equ:ThreeLayerNN}. Suppose the activation $\sigma$ has first and second derivative bounded by $L_{\sigma^{\prime}}$ and $L_{\sigma^{\prime\prime}}$ respectively. Moreover, suppose the function $g$ has bounded derivative $|g^{\prime}(\widetilde{y},y)| < L_{g^{\prime}}$. Then with $\omega = (A^{(1)},v)$
\begin{align*}
    \nabla^2 R(\omega) \succeq
    - 
    \Big( L_{\sigma^{\prime\prime}}L_{g^{\prime}} \frac{ \|\widehat{\Sigma}\|_2 \|\widehat{\Sigma}_{A^{(2)}}\|_2 }{M_1^{2c} M_2^{c-1} } \|v\|_{\infty} 
    + 
    2 L_{g^{\prime}} L_{\sigma^{\prime}} \frac{\sqrt{\|\widehat{\Sigma}\|_2 \|\widehat{\Sigma}_{A^{(2)}}\|_2} }{M_2^{c-1/2} M_1^{c}} 
    \Big)I.
\end{align*}
\end{theorem}
We now briefly discuss the above theorem, with the proof presented thereafter. We note that the lower bound in Theorem \ref{thm:WeakConvexity:TwoLayerNN:Both} now scales with the product of layer widths i.e. $M_1^{2c}M_2^{c-1}$ and $M^{c-1/2}_2 M_1^c$. Let us then suppose that each layer is the same width so $M_1 = M_2 = M$, the spectral norm of the second layer's empirical covariance $\|\widehat{\Sigma}_{A^{(2)}}\|_{2}$ is constant and the third layer remains bounded $\|v\|_{\infty}$. In this case the bound is then $O(1/(M^{2c-1/2}))$. In contrast, the bound for a two layer neural network presented in Theorem \ref{thm:WeakConvexity:SingleLayerNN:Both} is $O(1/M^c)$. The lower bounds are then the same order for $c=1/2$, while for $c > 1/2$, the three layer neural network case is smaller. We now give the proof of Theorem \ref{thm:WeakConvexity:TwoLayerNN:Both}.
\begin{proof}[Theorem \ref{thm:WeakConvexity:TwoLayerNN:Both}]

In a similar manner to the proof of Theorem \ref{thm:WeakConvexity:SingleLayerNN:Both}, let us begin by defining the gradient and Hessian of the prediction function with respect to the parameter $\omega$. In this case they are, respectively, a vector $\nabla f(x,\omega) \in \mathbb{R}^{M_1 d  + M_2} $ and matrix $\nabla^2 f(x,\omega) \in \mathbb{R}^{(M_1 d + M_2) \times (M_1 d + M_2)}$.

Once again split the co-ordinates of the gradient $(\nabla f(x,\omega))_{i}$ for $i=1,\dots,M_1d + M_2$ into two parts associated to the first and third layers. For the first layer, consider $i=(j-1)d + k$ with $j=1,\dots,M_1$ and $k=1,\dots,d$. This aligns with the gradient of the $j$th neuron in the first layer and $k$th input, which is  
\begin{align*}
    (\nabla f(x,\omega))_{i} =  \frac{\partial f(x,\omega)}{\partial A^{(1)}_{jk}}
    = 
    \frac{1}{M_2^c}\sum_{i=1}^{M_2}v_i \sigma^{\prime}\Big( \frac{1}{M_1^c}\sum_{s=1}^{M_1}A_{is}^{(2)}\langle A_s^{(1)},x\rangle \Big) \frac{A^{(2)}_{i j}}{M_1^c} (x)_{k}.
\end{align*}
For the third layer consider $i=M_1 d + k$ with $k=1,\dots,M_2$. This aligns with the gradient of the $k$th weight in the third layer, which is  
\begin{align*}
    (\nabla f(x,\omega))_{i} = \frac{\partial f(x,\omega)}{\partial v_k} =  \frac{1}{M_2^c} \sigma\Big( \frac{1}{M_1^c} \sum_{s=1}^{M_1}A^{(2)}_{ks} \langle A^{(1)}_{s},x\rangle \Big).
\end{align*}

Let us now decompose the entries of the Hessian $(\nabla^2 f(x,\omega))_{i \ell}$ for $i,\ell = 1,\dots, M_1 d + M_2$ into three parts associated to the second derivative of the first layer only, second derivative of the third layer only, and the derivative with respect to the first and third layer. For the second deriative of the first layer consider the indices $i=(j-1)d +k$ and $\ell = (\widetilde{j} - 1) d + \widetilde{k}$ with $j,\widetilde{j}=1,\dots,M_1$ and $k,\widetilde{k} =1,\dots,d$. For these indices the Hessian is   
\begin{align*}
    (\nabla^2 f(x,\omega))_{i \ell} 
    := \frac{\partial^2 f(x,\omega)}{\partial A_{jk}^{(1)} \partial A^{(1)}_{\widetilde{j}\widetilde{k}}} = \frac{1}{M_2^c}\sum_{i=1}^{M_2}v_i \sigma^{\prime\prime}\Big( \frac{1}{M_1^c}\sum_{s=1}^{M_1}A_{is}^{(2)}\langle A_s,x\rangle \Big) \frac{A^{(2)}_{i j}}{M_1^c} \frac{A^{(2)}_{i \widetilde{j}}}{M_1^c} (x)_{k}(x)_{\widetilde{k}}.
\end{align*}
Meanwhile for the Hessian with respect to the third layer consider the indices $i=M_1 d +k$ and $\ell=M_1 d + \widetilde{k}$ with $k,\widetilde{k} = 1,\dots,M_2$. The Hessian in this case is then 
\begin{align*}
    (\nabla^2 f(x,\omega))_{i \ell} 
    := \frac{\partial^2 f(x,\omega)}{\partial v_k \partial v_{\widetilde{k}} } 
    = 
    0.
\end{align*}
Finally, for the derivative with respect to the first and third layers consider the indices $i=(j-1)d + k$ and $\ell = M_1 d + \widetilde{k}$ with $j=1,\dots,M_1$ and $k = 1,\dots,d$ and $\widetilde{k}=1,\dots,M_2$. The Hessian in this case is then 
\begin{align*}
    (\nabla^2 f(x,\omega))_{i \ell} 
    := 
    \frac{\partial^2 f(x,\omega)}{ \partial A^{(1)}_{jk} \partial v_{\widetilde{k}} } = 
    \frac{1}{M_2^c} 
    \sigma^{\prime}\Big( \frac{1}{M_1^c}\sum_{s=1}^{M_1}A_{\widetilde{k}s}^{(2)}\langle A_s^{(1)},x\rangle \Big) \frac{A^{(2)}_{\widetilde{k} j}}{M_1^c} (x)_{k}.
\end{align*}
Let the vector $u\in \mathbb{R}^{M_1 d + M_2}$ have unit norm $\|u\|_2 = 1$ and be composed in a manner matching the parameter $\omega = (A^{(1)},v)$, so that $u = (A^{(1)}(u),v(u))$ where $A^{(1)}(u) \in \mathbb{R}^{M_1\times d}$ has been vectorised in a row-major manner and $v(u)\in \mathbb{R}^{M_2}$. Following the previous definitions, the quadratic form of the prediction function Hessian then takes the form 
\begin{align*}
    u^{\top}\nabla^2 f(x,\omega) u 
    & = 
    \sum_{j,\widetilde{j}=1}^{M_1} \sum_{k,\widetilde{k}=1}^{d} A^{(1)}(u)_{jk}A^{(1)}(u)_{\widetilde{j}\widetilde{k}} 
    \frac{\partial^2 f(x,\omega)}{\partial A_{jk}^{(1)} \partial A^{(1)}_{\widetilde{j}\widetilde{k}}} 
    + 
     2 \sum_{j=1}^{M_1}\sum_{k=1}^{d}\sum_{\widetilde{k}=1}^{M_2} 
     A^{(1)}(u)_{jk} v(u)_{\widetilde{k}} \frac{\partial^2 f(x,\omega)}{ \partial A^{(1)}_{jk} \partial v_{\widetilde{k}} } \\
     & \quad\quad 
     + 
     \sum_{k,\widetilde{k} = 1}^{M_2} v(u)_{k}v(u)_{\widetilde{k}} \frac{\partial^2 f(x,\omega)}{\partial v_{k} \partial v_{\widetilde{k}}}\\
     & = 
     \sum_{j,\widetilde{j}=1}^{M_1} 
     \frac{1}{M_2^c}\sum_{i=1}^{M_2}v_i \sigma^{\prime\prime}\Big( \frac{1}{M_1^c}\sum_{s=1}^{M_1}A_{is}^{(2)}\langle A_s^{(1)},x\rangle \Big) \frac{A^{(2)}_{i j}}{M_1^c} \frac{A^{(2)}_{i \widetilde{j}}}{M_1^c} \langle A^{(1)}(u)_j,x\rangle\langle A^{(1)}(u)_{\widetilde{j}},x\rangle \\
     & \quad\quad 
     + 2 \sum_{j=1}^{M_1}\sum_{\widetilde{k}=1}^{M_2} 
     v(u)_{\widetilde{k}}
     \frac{1}{M_2^c} 
    \sigma^{\prime}\Big( \frac{1}{M_1^c}\sum_{s=1}^{M_1}A_{\widetilde{k}s}^{(2)}\langle A_s^{(1)},x\rangle \Big) \frac{A^{(2)}_{\widetilde{k} j}}{M_1^c} \langle x, A^{(1)}(u)_j\rangle \\
    & = 
     \frac{1}{M_2^c}\sum_{i=1}^{M_2}v_i \sigma^{\prime\prime}\Big( \frac{1}{M_1^c}\sum_{s=1}^{M_1}A_{is}^{(2)}\langle A_s^{(1)},x\rangle \Big) 
     \frac{1}{M_1^{2c}} 
     \Big\langle \sum_{j=1}^{M_1} A^{(2)}_{ij} A^{(1)}(u)_{j},x\Big\rangle^2 \\
     & \quad\quad 
     + 
     2 \sum_{\widetilde{k}=1}^{M_2} 
     v(u)_{\widetilde{k}}
     \frac{1}{M_2^c} 
    \sigma^{\prime}\Big( \frac{1}{M_1^c}\sum_{s=1}^{M_1}A_{\widetilde{k}s}^{(2)}\langle A_s^{(1)},x\rangle \Big) \frac{1}{M_1^c} \Big\langle x, \sum_{j=1}^{M} A^{(2)}_{\widetilde{k} j} A^{(1)}(u)_j \Big\rangle 
\end{align*}
where on the second equality we have taken the summation over $j,\widetilde{j}=1,\dots,M$ inside the inner products. We now set to lower bound the quadratic form involving the empirical risk Hessian $u^{\top} \nabla^2 R(\omega) u$, and thus, the minimum Eigenvalue. Recalling the discussion around equation \eqref{equ:HessianLowerBound} and plugging in the above we get
\begin{align*}
    & u^{\top} \nabla^2  R(\omega) u \\
    & \geq 
    \frac{1}{N} \sum_{\ell=1}^{N}
    g^{\prime}(f(x_\ell,\omega),y_\ell) u^{\top} \nabla^2 f(x_\ell,\omega)u \\
    & = 
    \underbrace{ \frac{1}{N}\sum_{\ell=1}^{N}
    g^{\prime}(f(x_\ell,\omega),y_\ell)
    \frac{1}{M_2^c}\sum_{i=1}^{M_2}v_i \sigma^{\prime\prime}\Big( \frac{1}{M_1^c}\sum_{s=1}^{M_1}A_{is}^{(2)}\langle A_s^{(1)},x_{\ell}\rangle \Big) 
     \frac{1}{M_1^{2c}} 
     \Big\langle \sum_{j=1}^{M_1} A^{(2)}_{ij} A^{(1)}(u)_{j},x_{\ell}\Big\rangle^2}_{=:\textbf{Term 1}} \\
     &\quad\quad 
     + 
     \underbrace{ 
     2 \frac{1}{N}\sum_{\ell=1}^{N} 
     g^{\prime}(f(x_\ell,\omega),y_\ell)
    \sum_{\widetilde{k}=1}^{M_2} 
     v(u)_{\widetilde{k}}
     \frac{1}{M_2^c} 
    \sigma^{\prime}\Big( \frac{1}{M_1^c}\sum_{s=1}^{M_1}A_{\widetilde{k}s}^{(2)}\langle A_s^{(1)},x_{\ell}\rangle \Big) \frac{1}{M_1^c} \Big\langle x_{\ell}, \sum_{j=1}^{M} A^{(2)}_{\widetilde{k} j} A^{(1)}(u)_j \Big\rangle
    }_{=:\textbf{Term 2}}.
\end{align*}
Which has then been decomposed into two components labelled \textbf{Term 1} and \textbf{Term 2}, each of which will now be lower bounded separately. For \textbf{Term 1}, aligning with the Hessian of the first layer, we get using the upper bound on the gradient of $g$ i.e $|g^{\prime}(\cdot,y)| \leq L_{g^{\prime}}$ and the activation $|\sigma^{\prime\prime}(\cdot)|\leq L_{\sigma^{\prime\prime}}$
\begin{align*}
    \textbf{Term 1} \geq
    -\frac{L_{g^{\prime}}L_{\sigma^{\prime\prime}} \|v\|_{\infty} }{M_2^c} \frac{1}{N} \frac{1}{M_1^{2c}}  \sum_{\ell=1}^{N}\sum_{i=1}^{M_2} 
    \Big\langle \sum_{j=1}^{M_1} A^{(2)}_{ij} A^{(1)}(u)_{j},x_{\ell}\Big\rangle^2
\end{align*}
Meanwhile, for \textbf{Term 2} bring out the sum over $\widetilde{k}=1,\dots,{M_2}$ and apply Cauchy-Schwarz to get
\begin{align*}
    & \textbf{Term 2} \\
    & = 
    \sum_{\widetilde{k}=1}^{M_2} 
     v(u)_{\widetilde{k}} 
     \frac{1}{N}\sum_{\ell=1}^{N} 
     g^{\prime}(f(x_\ell,\omega),y_\ell)
     \frac{1}{M_2^c} 
    \sigma^{\prime}\Big( \frac{1}{M_1^c}\sum_{s=1}^{M_1}A_{\widetilde{k}s}^{(2)}\langle A_s^{(1)},x_{\ell}\rangle \Big) \frac{1}{M_1^c} \Big\langle x_{\ell}, \sum_{j=1}^{M_1} A^{(2)}_{\widetilde{k} j} A^{(1)}(u)_j \Big\rangle \\
    & \geq 
    \!- \!\! \sqrt{\sum_{\widetilde{k}=1}^{M_2} 
     v(u)_{\widetilde{k}}^2 
     }
     \sqrt{ \sum_{\widetilde{k}=1}^{M_2} 
     \Big( 
     \frac{1}{N}
     \sum_{\ell=1}^{N} 
     g^{\prime}(f(x_\ell,\omega),y_\ell)
     \frac{1}{M_2^c} 
    \sigma^{\prime}\Big( \frac{1}{M_1^c}\sum_{s=1}^{M_1}A_{\widetilde{k}s}^{(2)}\langle A_s^{(1)},x_{\ell}\rangle \! \Big) \frac{1}{M_1^c} \Big\langle x_{\ell},
    \! \sum_{j=1}^{M} A^{(2)}_{\widetilde{k} j} A^{(1)}(u)_j \Big\rangle
     \Big)^2}\\
     & \geq 
     - \frac{L_{g^{\prime}}L_{\sigma^{\prime}}}{M_2^{c} M_1^c}
     \sqrt{\sum_{\widetilde{k}=1}^{M_2} 
     v(u)_{\widetilde{k}}^2 
     }
     \sqrt{\sum_{\widetilde{k}=1}^{M_2}
     \frac{1}{N}\sum_{\ell=1}^{N}
     \Big\langle x_{\ell}, \sum_{j=1}^{M_1} A^{(2)}_{\widetilde{k} j} A^{(1)}(u)_j \Big\rangle^2
     }.
\end{align*}
Where for the second inequality we have followed a similar set of steps to those within the proof of Theorem \ref{thm:WeakConvexity:SingleLayerNN:Both}. In particular, we applied convexity of $x\rightarrow x^2$ as well as the upper bounds on the gradient of $g$ and the activation $\sigma$. We are now left to upper bound the quantity  $\frac{1}{N} \sum_{\ell=1}^{N}\sum_{i=1}^{M_2} \Big\langle \sum_{j=1}^{M_1} A^{(2)}_{ij} A^{(1)}(u)_{j},x_{\ell}\Big\rangle^2$ which appears within the bounds for both \textbf{Term 1} and \textbf{Term 2}. To bound this quantity, let us rewrite the summation  as a vector matrix multiplication so $\sum_{j=1}^{M_1} A^{(2)}_{ij} A^{(1)}(u)_{j} = A^{(2)}_i A^{(1)}(u)$ where $A^{(2)}_i$ is the $i$th row of $A^{(2)}$. Normalising by $M_2$, we note this quantity can be bounded in terms of the spectral norm of the empirical covariance $\|\widehat{\Sigma}\|_2$ and covariance of the second layer $\|\widehat{\Sigma}_{A^{(2)}}\|_2$. Specifically, 
\begin{align*}
    \frac{1}{M_2} \frac{1}{N} \sum_{\ell=1}^{N}\sum_{i=1}^{M_2} \Big\langle \sum_{j=1}^{M_1} A^{(2)}_{ij} A^{(1)}(u)_{j},x_{\ell}\Big\rangle^2
    & = 
    \frac{1}{M_2} \sum_{i=1}^{M_2}
    \Big( A^{(2)}_i A^{(1)}(u) \Big)^{\top} 
    \underbrace{ \Big( \frac{1}{N} \sum_{\ell=1}^{N} x_{\ell} x_{\ell}^{\top} \Big) }_{\widehat{\Sigma}}
    \Big( A^{(2)}_i A^{(1)}(u) \Big) \\
    & \leq 
    \|\widehat{\Sigma}\|_2
    \frac{1}{M_2} \sum_{i=1}^{M_2}
    \|A^{(2)}_i A^{(1)}(u)\|_2^2 \\
    & = 
    \|\widehat{\Sigma}\|_2
    \frac{1}{M_2} \sum_{i=1}^{M_2}
    \trace\Big( A^{(2)}_i A^{(1)}(u) \big( A^{(1)}(u)\big)^{\top}  \big(A^{(2)}_i \big)^{\top} \Big)\\
    & = 
    \|\widehat{\Sigma}\|_2
    \trace \Big( A^{(1)}(u) \big( A^{(1)}(u)\big)^{\top}
    \underbrace{ \Big( \frac{1}{M_2} \sum_{i=1}^{M_2} \big(A^{(2)}_i \big)^{\top}A^{(2)}_i \Big) }_{\widehat{\Sigma}_{A^{(2)}}} \Big)  \\
    & \leq 
    \|\widehat{\Sigma}\|_2
    \|\widehat{\Sigma}_{A^{(2)}}\|_2 
    \|A^{(1)}(u)\|_F^2
\end{align*}
The final lower bound is then arrived at by using the above to lower bound \textbf{Term 1} and \textbf{Term 2} and recalling that $\|u\|_2^2 = \|v(u)\|_2^2 + \|A^{(1)}(u)\|_{F}^{2} = 1$. 
\end{proof}

\subsection{Bounding Optimisation and Generalisation Error for a Two Layer Neural Network (Theorem \ref{thm:WeakConvexityError:TwoLayer})}
In this section we present the proof of Theorem \ref{thm:WeakConvexityError:TwoLayer} from Appendix \ref{sec:TwoLayerNN:Approx:FirstLayer}. The proof proceeds in two steps. The first step is to apply Proposition \ref{prop:TestError:General} from Appendix \ref{sec:GeneralisedWeakConvexity} to bound the \textbf{Opt. \& Approx. Error}. The second step involves utilising the structure of $G_{C}(\cdot)$ to decouple the place holder $\widetilde{\omega}$ and the iterates of gradient descent $\widehat{\omega}_{s}$. Each of these steps will, respectively, be included in the following paragraphs \textbf{Applying Proposition \ref{prop:TestError:General}} and \textbf{Decoupling Gradient Descent Iterates}.

\textbf{Applying Proposition \ref{prop:TestError:General}} To applying Proposition \ref{prop:TestError:General} we require showing two conditions: Assumption \ref{ass:WeakConvexity:Local} holds for the  set $\mathcal{X}_{L_{v}}$ and function $G_{C}(\cdot)$ described in Section \ref{sec:TwoLayerNN:Approx:FirstLayer}; and the iterates of gradient descent to remain within the set $\widehat{\omega}_{s} \in \mathcal{X}_{L_{v}}$ for $s \geq 0$. Let us begin with the first of these conditions.

To ensure Assumption \ref{ass:WeakConvexity:Local} holds we must lower bound the quadratic form $u^{\top} \nabla^2 R(\omega) u \geq -G_{C}(u)$ for $\omega \in \mathcal{X}_{L_{v}}$ and $ u \in \mathbb{R}^{Md + M}$. To this end, let us recall the steps in the proof of Theorem \ref{thm:WeakConvexity:SingleLayerNN:Both} in Appendix \ref{sec:WeakConvexityNN:Thm2}. Specifically, looking to equation \eqref{equ:ThmCheckpoint:1} where we have for $u = (A(u),v(u)) \in \mathbb{R}^{Md + M}$ a lower bound on the quadratic form $u^{\top} \nabla^2 R(\omega)u$ involving the sum of two terms. For the first term simply consider the lower bound presented within Theorem \ref{thm:WeakConvexity:SingleLayerNN:Both} which is  $-L_{g^{\prime}} L_{\sigma^{\prime\prime}} \|\widehat{\Sigma}\|_2 \|v\|_{\infty} \|A(u)\|_{F}^2 /M^c$. For the second term note it can be lower bounded    
\begin{align*}
     & 2  \frac{1}{M^c} \frac{1}{N} \sum_{i=1}^{N} g^{\prime}(f(x_i,\omega),y_i)  
     \sum_{j=1}^{M} v(u)_{j} \sigma^{\prime}(\langle A_j,x_i \rangle) \langle x_i,A(u)_j\rangle\\
     & \geq 
     - 2 \frac{L_{g^{\prime}}}{M^c} \frac{1}{N} \sum_{i=1}^{N} 
     \Big| 
      \Big\langle \sum_{j=1}^{M} v(u)_{j} \sigma^{\prime}(\langle A_j,x_i \rangle) A(u)_j, x_i \Big\rangle
     \Big| \\
     & =  
     - 2 \frac{L_{g^{\prime}}}{M^c} \frac{1}{N} \sum_{i=1}^{N} 
     \Big| 
      \Big\langle 
      \sum_{j=1}^{M} v(u)_{j} \sigma^{\prime}(\langle A_j,x_i \rangle) A(u)_j, x_i \Big\rangle
     \Big| \\
     & \geq 
    - 2\frac{L_{g^{\prime}}}{M^c} \frac{1}{N} \sum_{i=1}^{N} 
     \Big\|  \sum_{j=1}^{M} v(u)_{j} \sigma^{\prime}(\langle A_j,x_i \rangle) A(u)_j\Big\|_{2}
     \| x_i \|_2
\end{align*}
where at the end we applied H\"olders inequality. Furthermore, if we denote the vector $\alpha \in \mathbb{R}^{M}$ such that $\alpha_\ell = \sigma^{\prime}(\langle A_l,x_i \rangle)$ for $\ell =1,\dots,M$, note we can rewrite $\sum_{j=1}^{M} v(u)_{j} \sigma^{\prime}(\langle A_j,x_i \rangle) A(u)_j = A(u)^{\top}\text{Diag}(\alpha)v(u)$. Since the activation $\sigma$ has derivative bounded by $L_{\sigma^{\prime}}$ we then have $\alpha_i/L_{\sigma^{\prime}} \in [-1,1]$ and the upper bound $\big\|A(u)^{\top} \text{Diag}(\alpha) v(u) \big\|_{2} \leq L_{\sigma^{\prime}} \max_{\|z\|_{\infty} \leq 1 }\big\| A(u)^{\top} \text{Diag}(z) v(u) \big\|_{2}$. Combined with the upper bound on the covariates $\frac{1}{N} \sum_{i=1}^{N} \|x_i\|_2 \leq \frac{1}{N} \sqrt{N} \sqrt{ \sum_{i=1}^{N} \|x_i\|_2^2} = \sqrt{\trace\big( \widehat{\Sigma})} $,  we get the lower bound  
\begin{align*}
    & 2  \frac{1}{M^c} \frac{1}{N} \sum_{i=1}^{N} g^{\prime}(f(x_i,\omega),y_i)  
     \sum_{j=1}^{M} v(u)_{j} \sigma^{\prime}(\langle A_j,x_i \rangle) \langle x_i,A(u)_j\rangle\\
     & \geq 
      - 2\frac{L_{g^{\prime}} L_{\sigma^{\prime}} \sqrt{\trace\big( \widehat{\Sigma} \big)} }{M^c}  
      \max_{ \|z\|_{\infty} \leq 1  }\big\|A(u)^{\top} \text{Diag}(z) v(u) \big\|_{2}.
\end{align*}
Bringing together the bounds on the two terms yields 
\begin{align*}
    u^{\top} \nabla^2 R(\omega) u 
    \geq - \frac{ L_{g^{\prime}} L_{\sigma^{\prime\prime}} \|\widehat{\Sigma}\|_2 }{M^c} \|v\|_{\infty} \|A(u)\|_{F}^2 
    - 2\frac{ L_{g^{\prime}} L_{\sigma^{\prime}}\sqrt{\trace\big( \widehat{\Sigma} \big)} }{M^c}  \max_{ \|z\|_{\infty} \leq 1}\Big\| A(u)^{\top} \text{Diag}(z) v(u) \Big\|_{2}.
\end{align*}
We then see that Assumption \ref{ass:WeakConvexity:Local} is satisfied with $\mathcal{X}_{L_{v}}$ and $G_{C}(\cdot)$ provided $C \geq 2 L_{g^{\prime}}  
\Big[ \Big(  L_{\sigma^{\prime}} \sqrt{\trace\big( \widehat{\Sigma} \big)} \Big) \vee ( L_{\sigma^{\prime\prime}} L_{v} \|\widehat{\Sigma}\|_2 ) \Big]$. 

Let us now ensure the iterates of gradient descent remain within the set $\mathcal{X}_{L_{v}}$. To do so we must consider the infinity norm of the second layer throughout training. Let the gradient descent iterates be denoted as $\widehat{\omega}_s = (A^{s},v^{s})$ for $s \geq 0$ where $A^{s} \in \mathbb{R}^{M \times d}$ has been vectorised in a row-major manner. Plugging in the definition of the empirical risk gradient, the second layer is updated for $s \geq 0$ as 
\begin{align*}
    v^{s+1}_{k} = v^{s}_k - \eta \frac{1}{N}\sum_{i=1}^{N}g^{\prime}(f(x_i,\omega),y_i) 
    \frac{1}{M^c} \sigma(\langle A_k,x_i\rangle). 
\end{align*}
Using the upper bound on the derivative $|g^{\prime}(\cdot)|\leq L_{g^{\prime}}$ as well as the activation $|\sigma(\cdot)| \leq L_{\sigma}$ we get $\|v^{s+1}\|_{\infty} \leq \|v^{s}\|_{\infty} + \frac{\eta L_{g^{\prime}} L_{\sigma}}{M^c} \leq \|v^{0}\|_{\infty} + \frac{\eta t L_{g^{\prime}} L_{\sigma}}{M^c} $. Therefore, we can ensure $\widehat{\omega}_{s} \in \mathcal{X}_{L_{v}}$ for $t \geq s \geq 0$ if $L_v \geq \|v^{0}\|_{\infty} + \frac{\eta t L_{g^{\prime}} L_{\sigma}}{M^c} $. We can now apply Proposition \ref{prop:TestError:General}, to upper bound the \textbf{Opt \& Approx. Error}. 

\textbf{Decoupling Gradient Descent Iterates}:  Given the upper bound presented within Proposition \ref{prop:TestError:General}, we must now decoupling the iterates of gradient descent $\widehat{\omega}_{s}$ and the placeholder $\widetilde{\omega}$ within the \textbf{Approximation Error}, specifically $G_{C}(\widetilde{\omega} - \widehat{\omega}_{s})$.  We will, in short, add and subtract the initialisation of gradient descent $\widehat{\omega}_{0}$ and apply triangle inequality a number of times. Let us denote the placeholder $\widetilde{\omega} = (\widetilde{A},\widetilde{v})$ where $\widetilde{A} \in \mathbb{R}^{M \times d}$ has been vectorised in a row-major manner and the second layer of weights is $\widetilde{v} \in \mathbb{R}^{M}$. For a general $\omega = (A,v)$ and $\widehat{\omega}_0 = (A^{0},v^{0})$, we then get when adding and subtracting $A^{0}$ inside the Frobeinus norm to get 
\begin{align*}
    G_{C}(\widetilde{\omega} - \omega)
    & \leq 
    \frac{2 C}{M^c} \|\widetilde{A} - A^{0}\|_F^2
    + 
    \frac{2 C}{M^c} \| A^{0} - A\|_F^2 \\
    & \quad\quad 
    + 
   \frac{C}{M^c}  \max_{ \|z\|_{\infty} \leq 1}
    \big\| (\widetilde{A} - A)^{\top} \text{Diag}(z) (\widetilde{v} - v) \big\|_2.
\end{align*}
We now consider the second term. Adding and subtracting $A^{0}$ and $v^{0}$ we get the following four terms 
\begin{align*}
    & \big\| (\widetilde{A} - A)^{\top} \text{Diag}(z) (\widetilde{v} - v) \big\|_2\\
     & \leq 
    \big\|  (\widetilde{A} - A^{0} )^{\top} \text{Diag}(z) (\widetilde{v} - v) \big\|_2
    + 
    \big\|  ( A^{0} - A )^{\top} \text{Diag}(z) (\widetilde{v} - v) \big\|_2\\
    & \leq
    \big\|  (\widetilde{A} - A^{0} )^{\top} \text{Diag}(z) (\widetilde{v} - v^{0} ) \big\|_2
    + 
    \big\|  (\widetilde{A} - A^{0} )^{\top} \text{Diag}(z) (v^{0}- v) \big\|_2\\
    &\quad\quad
     + \big\|  ( A^{0} - A )^{\top} \text{Diag}(z) (\widetilde{v} - v^{0} ) \big\|_2
     +
     \big\|  ( A^{0} - A )^{\top} \text{Diag}(z) (v^{0} - v ) \big\|_2.
\end{align*}
The second, third and fourth terms can then be bounded as follows.
The second term can be bounded 
\begin{align*}
    \big\|(\widetilde{A} - A^{0} )^{\top} \text{Diag}(z) (v^{0}- v) \big\|_2 
    & \leq 
    \big\|(\widetilde{A} - A^{0} )^{\top} \big\|_2 \| \text{Diag}(z)\|_2 \|v^{0}- v \|_2 \\
    & 
    \leq 
    \frac{1}{2} \| \widetilde{A} - A^{0} \|_F^2 
    + \frac{1}{2}  \|v^{0}- v \|_2^2\\
    & \leq 
    \frac{1}{2} \| \widetilde{A} - A^{0} \|_F^2 
    + \frac{1}{2} \|\widehat{\omega}_0 - \omega \|_2^2
\end{align*}
where we have used the following set of steps. The first inequality arises from using the consistency of matrix norms to be break the product of matrices into operator norms. The second inequality follows from: noting that $\|\text{Diag}(z)\|_2 \leq 1$ since $\|z\|_{\infty} \leq 1$;  upper bounding the matrix $\ell_2$ operator norm by the Frobenius norm; and using young's inequality to break apart the product of norms. The final inequality comes from simply adding the co-ordinates associated to the first layer to the second term. Following a similar set of steps the fourth term can be bounded 
\begin{align*}
    \big\| ( A^{0} - A )^{\top} \text{Diag}(z) (v^{0} - v ) \big\|_2
    & \leq \frac{1}{2}  \| A^{0} - A\|_F^2  + \frac{1}{2} \|v^{0}- v\|_2^2    \\
    & = \frac{1}{2}  \|\widehat{\omega}_0 - \omega \|_2^2.
\end{align*}
Meanwhile, the third term is bounded without applying Young's inequality so 
\begin{align*}
 \Big\| ( A^{0} - A )^{\top} \text{Diag}(z) (\widetilde{v} - v^{0}) \Big\|_2
& \leq  \| A^{0} - A \|_F \|\widetilde{v} - v^{0} \|_2\\
& \leq  \|\widehat{\omega}_0 - \omega\|_2 \|\widetilde{v} - v^{0}\|_2.
 \end{align*}
Bringing everything together we get the following upper bound on the function $G_{C}(\widetilde{\omega} - u)$
\begin{align*}
    G_{C}(\widetilde{\omega} - \omega)
    & \leq
    \frac{3 C }{M^c} \| \widehat{\omega}_0 - \omega\|_2^2
    \\
    &\quad\quad 
    + \frac{3 C}{M^c} \|\widetilde{A} -A^{0}\|_F^2
    +
    \frac{C}{M^c}  \| \widehat{\omega}_0 - \omega \|_2\|\widetilde{v} - v^{0} \|_2\\
    &\quad\quad 
    +
    \frac{C}{M^c} \max_{ \|z\|_{\infty} \leq 1} 
    \big\|(\widetilde{A} - A^{0} )^{\top} \text{Diag}(z) (\widetilde{v} - v^{0} ) \big\|_2.
\end{align*}
We now utilise the above within the \textbf{Approximation Error} in Proposition \ref{prop:TestError:General}. 
In particular, for $s=1,\dots,t$ pick $\omega = \widehat{\omega}_s$ and sum up the above for $s=1,\dots,t$. Utilising the bound $\|\widehat{\omega}_{s} - \widehat{\omega}_{0}\|_2 \leq \sqrt{2 \eta t  (R(\widehat{\omega}_0) - R(\widehat{\omega}^{\star}))}$ (See Lemma \ref{lem:OptError:General}) then yields 
\begin{align*}
     \frac{1}{t} \sum_{s=1}^{t} G_{C}(\widetilde{\omega} - \widehat{\omega}_s) 
    & \leq 
    \frac{6 C  \eta t  (R(\widehat{\omega}_0) - R(\widehat{\omega}^{\star}))}{M^c} 
    + \frac{3 C}{M^c} \|\widetilde{A} - A^{0} \|_F^2
    +
    \frac{C \sqrt{ 2 \eta t ((R(\widehat{\omega}_0) - R(\widehat{\omega}^{\star}))} }{M^c} 
    \|\widetilde{v} - v^{0} \|_2\\
    &\quad\quad 
    +
    \frac{C}{M^c} \max_{ \|z\|_{\infty} \leq 1} 
    \big\| (\widetilde{A} - A^{0} )^{\top} \text{Diag}(z) (\widetilde{v} - v^{0} ) \big\|_2 \\
    & \leq  
    \frac{6C  \eta t  (R(\widehat{\omega}_0) - R(\widehat{\omega}^{\star}))}{M^c} 
    + 
    \frac{3 C (\sqrt{ (R(\widehat{\omega}_0) - R(\widehat{\omega}^{\star})) } \vee 1)}{M^c}\\
    &\quad\quad\quad\quad
    \times 
    \underbrace{ 
    \Big( 
    \|\widetilde{A} - A^{0} \|_F^2 
    + 
    \sqrt{\eta t } \|\widetilde{v} - v^{0} \|_2
    + 
    \max_{ \|z\|_{\infty}\leq 1 } 
    \big\| (\widetilde{A} - A^{0} )^{\top} \text{Diag}(z) (\widetilde{v} - v^{0} ) \big\|_2
    \Big)
    }_{H(\widetilde{\omega} - \widehat{\omega}_0)}
\end{align*}
where on the second inequality we simply pulled out the constant on the second term. Plugging into the  \textbf{Approximation Error} within Proposition \ref{prop:TestError:General}, then yields 
\begin{align*}
    \frac{1}{2}\frac{1}{t} \sum_{s=1}^{t} G_{C}(\widetilde{\omega} - \widehat{\omega}_{s}) +  R(\widetilde{\omega}) - r(\omega^{\star})
    & \leq 
    \frac{3C  \eta t  (R(\widehat{\omega}_0) - R(\widehat{\omega}^{\star}))}{M^c} \\
    & \quad\quad + 
    \frac{3 C (\sqrt{ (R(\widehat{\omega}_0) - R(\widehat{\omega}^{\star})) } \vee 1)}{M^c}
    H(\widetilde{\omega} - \widehat{\omega}_0)
    +
    R(\widetilde{\omega}) - r(\omega^{\star}) \\
    & = 
    \frac{3C  \eta t  (R(\widehat{\omega}_0) - R(\widehat{\omega}^{\star}))}{M^c} 
    + \lambda H(\widetilde{\omega} - \widehat{\omega}_0) 
    +
    R(\widetilde{\omega}) - r(\omega^{\star})
\end{align*}
where we have defined $\lambda = \frac{3 C (\sqrt{(R(\widehat{\omega}_0) - R(\widehat{\omega}^{\star})) } \vee 1)}{M^c} $. With a population risk minimiser $\omega^{\star} = (A^{\star},v^{\star})$, we recall that $L_{v} \geq \|v^{\star}\|_{\infty}$ and thus, $\omega^{\star} \in \mathcal{X}_{L_{v}}$. Therefore, if we pick $\widetilde{\omega} = \widehat{\omega}_{\lambda} \in \argmin_{\omega \in \mathcal{X}_{L_{v}} }\big\{ R(\omega) + \lambda H(\omega -\widehat{\omega}_0) \big\} $, the right most quantity above can be upper bounded 
\begin{align*}
    \lambda H(\widetilde{\omega} - \widehat{\omega}_0) 
    +
    R(\widetilde{\omega}) - r(\omega^{\star}) 
    & \leq 
    \lambda H(\omega^{\star} - \widehat{\omega}_0) 
    +
    R(\omega^{\star}) - r(\omega^{\star}).
\end{align*}
Taking expectation and noting that $\E[R(\omega^{\star})] = r(\omega^{\star})$ yields the following bound on the \textbf{Approximation Error}
\begin{align*}
    \frac{1}{2}\frac{1}{t} \sum_{s=1}^{t} \E[G_{\mathcal{C}}(\widetilde{\omega} - \widehat{\omega}_{s})] +  \E[R(\widetilde{\omega})] - r(\omega^{\star})
    & \leq 
    \frac{3C  \eta t  \E[R(\widehat{\omega}_0) - R(\widehat{\omega}^{\star})]}{M^c}
    + \E[\lambda] H(\omega^{\star} - \widehat{\omega}_0).
\end{align*}
The result is then arrived at by plugging the above into the bound presented within Proposition \ref{prop:TestError:General}, and recalling that $\widetilde{\omega}= \widehat{\omega}_{\lambda}$.

\end{document}